\pdfoutput=1

\documentclass[10pt,twocolumn,letterpaper]{article}

\usepackage[pagenumbers]{wacv} %

\newif\ifarxiv

\usepackage{amsmath}
\usepackage{soul}
\usepackage{bm}
\usepackage{calc}
\usepackage{graphicx}
\usepackage{wrapfig}
\usepackage{enumitem}
\usepackage{graphicx}
\usepackage{array}
\usepackage{multirow}
\usepackage{graphicx}
\usepackage{subcaption}
\usepackage{placeins}
\usepackage{capt-of}

\arxivtrue

\definecolor{wacvblue}{rgb}{0.21,0.49,0.74}
\usepackage[pagebackref,breaklinks,colorlinks,allcolors=wacvblue]{hyperref}

\def\wacvPaperID{323} %
\def\confName{WACV}
\def\confYear{2027}

\title{Every Camera Effect, Every Time, All at Once: 4D Gaussian Ray Tracing for Physics-based Camera Effect Data Generation}

\author{
  Yi-Ruei Liu$^{1,3}$\thanks{Equal contribution. Work done at Academia Sinica as intern.},\space
  You-Zhe Xie$^{1,3}$\footnotemark[1],\ 
  Yu-Hsiang Hsu$^{2,3}$\footnotemark[1],\ 
  Jing-En Huang$^{1}$, \\
  I-Sheng Fang$^{3}$\thanks{Internship mentor.}, \
  Yu-Lun Liu$^{1}$,\ 
  Jun-Cheng Chen$^{3}$ \\
  $^{1}$Department of Computer Science, National Yang Ming Chiao Tung University \\
  $^{2}$National Taiwan University \\
  $^{3}$Research Center for Information Technology Innovation, Academia Sinica \\
}

\begin{document}

\twocolumn[{%
\renewcommand\twocolumn[1][]{#1}%
\maketitle

\vspace{-3em}
\begin{center}
    \centering
    \includegraphics[width=0.9\textwidth]{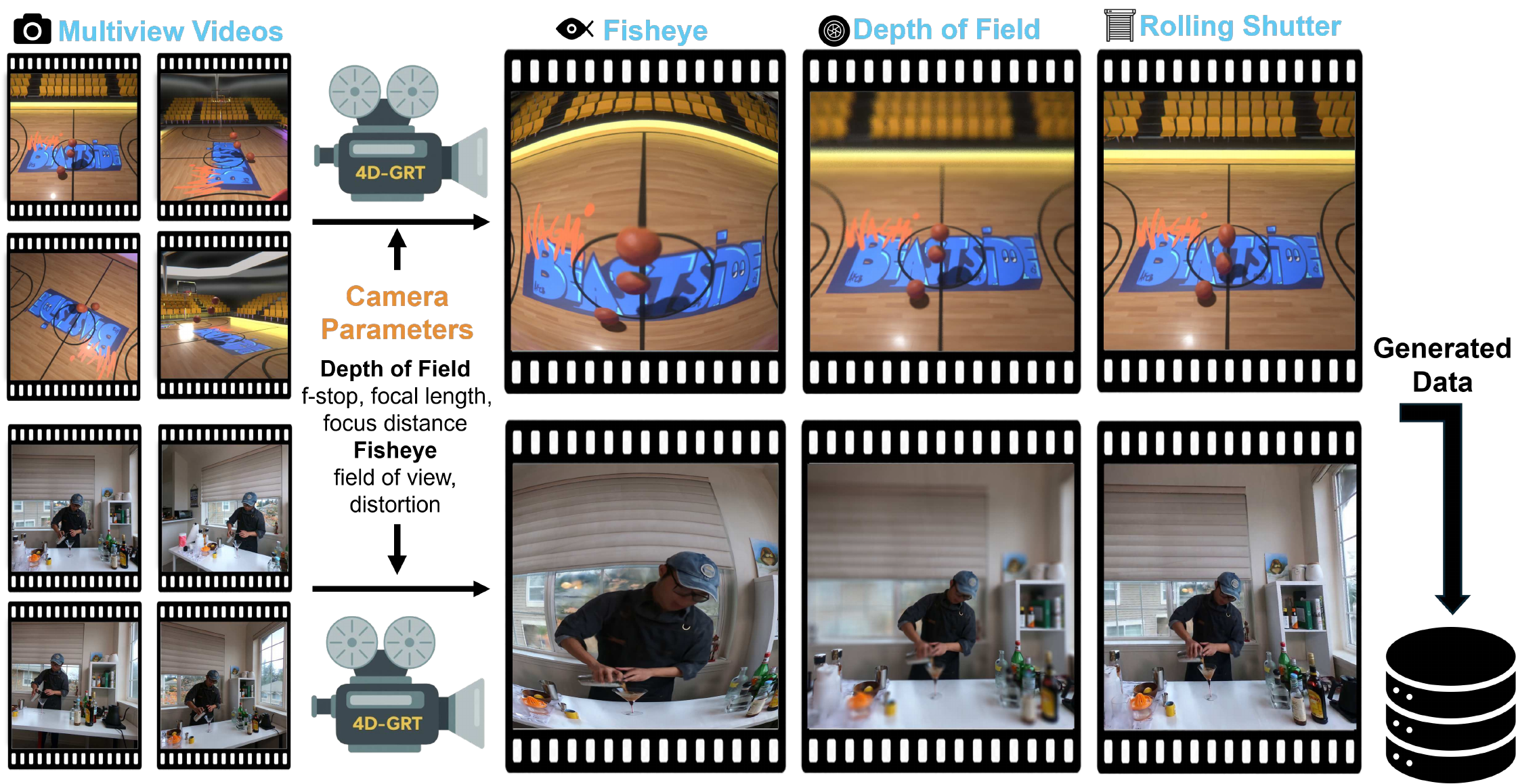}
    \captionof{figure}{
        We propose 4D Gaussian Ray Tracing (4D-GRT), a novel framework for generating physically grounded, controllable camera effects in dynamic scenes. (1) Given multi-view video input, our method reconstructs a dynamic scene using 4D Gaussian Splatting (4D-GS) and differentiable ray tracing. (2) We simulate various camera effects with controllable parameters using ray tracing, generating high-quality videos with controllable camera effects. \ifarxiv See the project page for the videos.\fi
    }
    \label{fig:teaser}
\end{center}%
}]
{\renewcommand{\thefootnote}{\fnsymbol{footnote}}%
\footnotetext[1]{Equal contribution. Work done at Academia Sinica as intern.}%
\footnotetext[2]{Internship mentor.}}

\begin{abstract}
    Current video diffusion models can generate recognizable camera-effect appearances but do not reliably follow specified physical parameters such as field of view, aperture, and readout time, partly because real dynamic videos with diverse and accurate camera-effect annotations are scarce. Physics-based simulation provides exact labels but suffers from a sim-to-real gap. Reconstructing real dynamic scenes offers an alternative, but Dynamic NeRF methods render too slowly, whereas rasterization-based Dynamic Gaussian Splatting cannot model the light transport required by these effects. We propose \textbf{4D Gaussian Ray Tracing (4D-GRT)}, a two-stage data-generation pipeline that reconstructs dynamic scenes as 4D Gaussians from multi-view pinhole videos and re-renders them with physically based camera models under specified parameters. On \textbf{CamVFX-4D}, our controlled benchmark of eight dynamic indoor scenes under four camera effects, 4D-GRT achieves up to 3.4 dB higher PSNR and 6.0$\times$ faster rendering than MSTH. Finally, we demonstrate two downstream applications: fine-tuning a video generation model on our data to improve camera-effect control and extending our pipeline to monocular input. Project page: \href{https://shigon255.github.io/4DGRT-project-page/}{https://shigon255.github.io/4DGRT-project-page}. 
\end{abstract}

\vspace{-2em}
\section{Introduction}
\label{sec:intro}
\vspace{-0.5em}
\begin{figure*}[htbp]
\centering
\setlength{\tabcolsep}{2pt}
\renewcommand{\arraystretch}{1.1}

\newcommand{\imgw}{0.24\linewidth}

\begin{minipage}{\linewidth}
\footnotesize\textit{``A video of a person skateboarding in a city square, captured using a \textbf{fisheye lens with a 180-degree field of view}.''}
\end{minipage}
\\[2pt]
\includegraphics[width=\imgw]{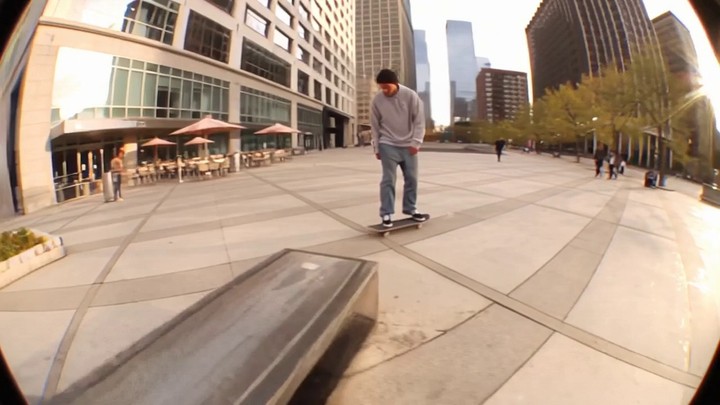}\hfill
\includegraphics[width=\imgw]{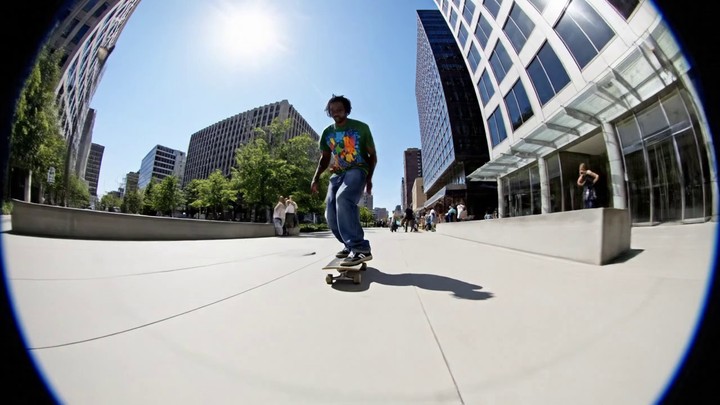}\hfill
\includegraphics[width=\imgw]{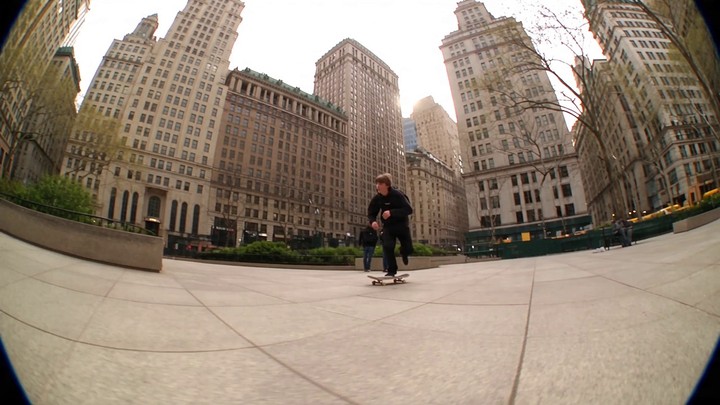}\hfill
\includegraphics[width=\imgw]{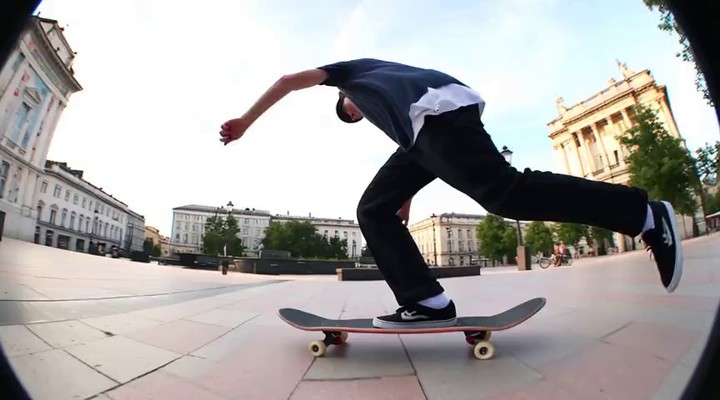}
\\[6pt]

\begin{minipage}{\linewidth}
\footnotesize\textit{``A video of a foreground chess piece on a chessboard, with several chess pieces receding into the background, shot with a \textbf{35 mm lens at aperture f/16.0}, with the focus distance set to the foreground chess piece.''}
\end{minipage}
\\[2pt]
\includegraphics[width=\imgw]{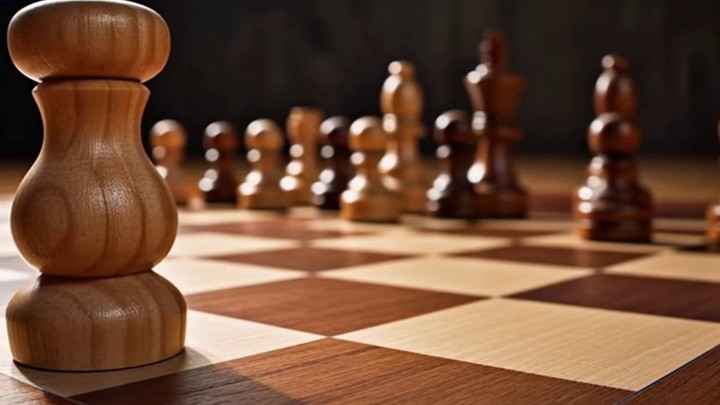}\hfill
\includegraphics[width=\imgw]{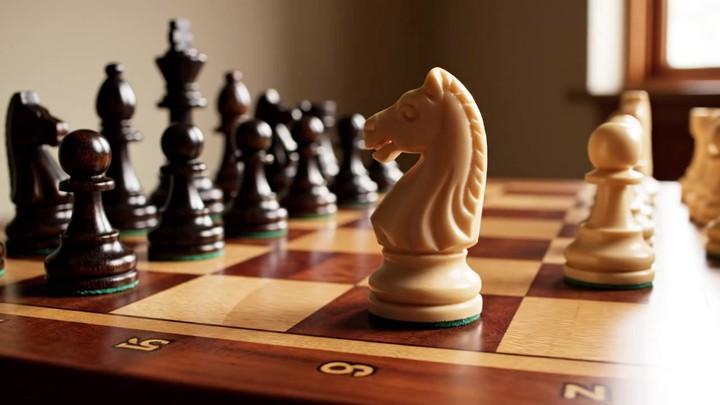}\hfill
\includegraphics[width=\imgw]{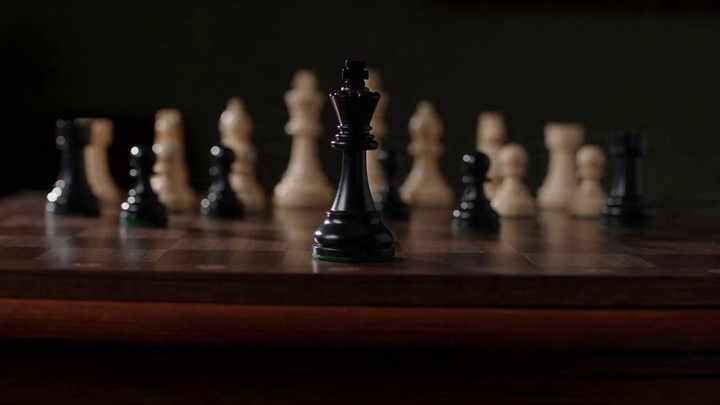}\hfill
\includegraphics[width=\imgw]{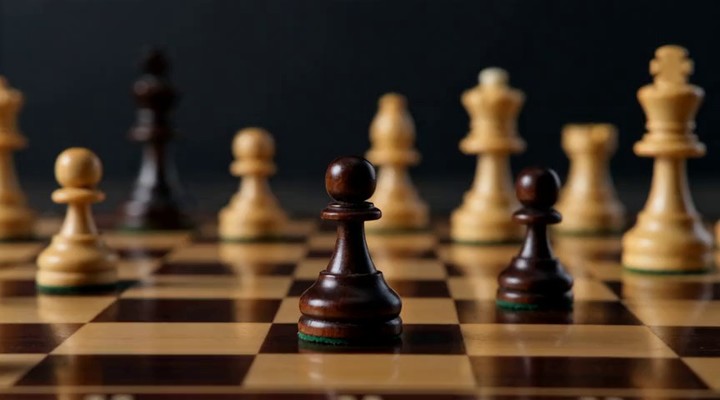}
\\[6pt]

\begin{minipage}{\linewidth}
\footnotesize\textit{``A close-up video of a centered electric fan with clearly visible blades spinning at high speed, captured with a \textbf{rolling-shutter camera using a readout time of 40 ms}. The fan fills most of the frame, the background is simple, and the exposure is short with minimal motion blur.''}
\end{minipage}
\\[2pt]
\includegraphics[width=\imgw]{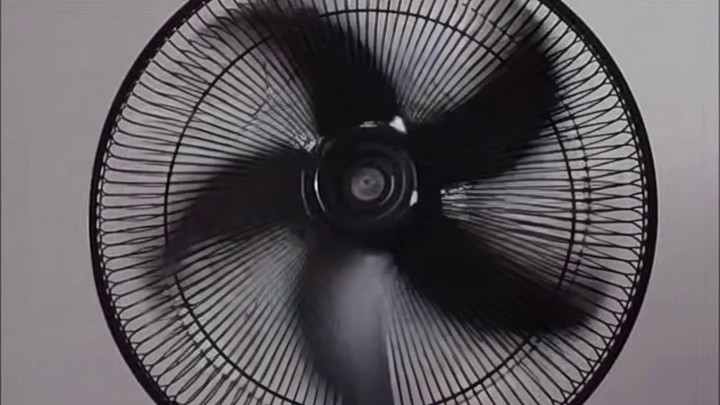}\hfill
\includegraphics[width=\imgw]{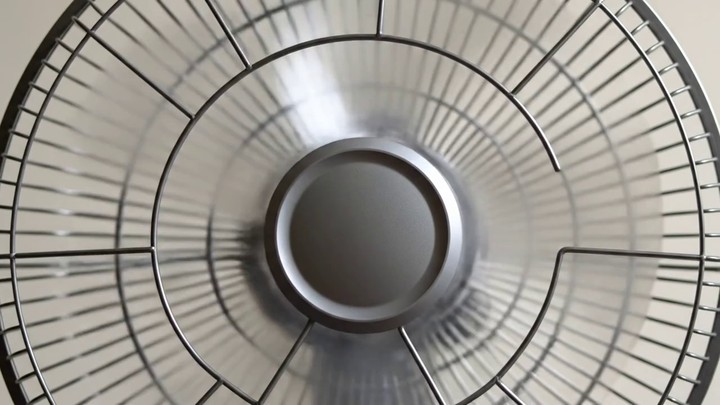}\hfill
\includegraphics[width=\imgw]{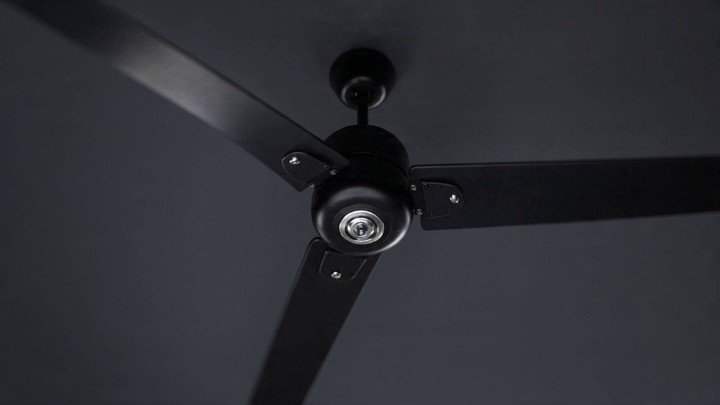}\hfill
\includegraphics[width=\imgw]{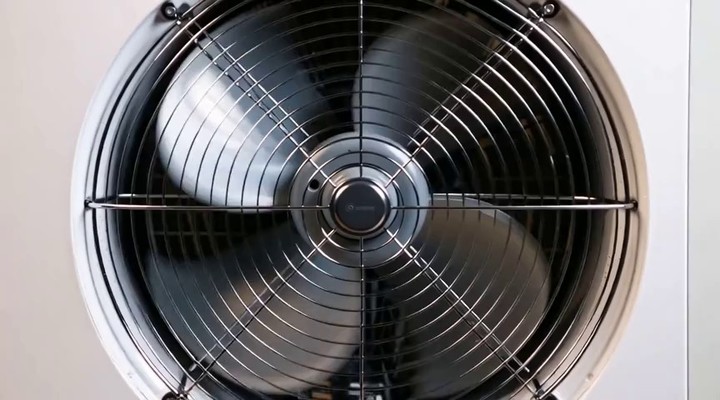}
\\[4pt]

\makebox[\imgw]{\small\textbf{Sora 2 Pro~\cite{OpenAISora2}}}\hfill
\makebox[\imgw]{\small\textbf{Veo 3.1~\cite{GoogleVeo3}}}\hfill
\makebox[\imgw]{\small\textbf{Seedance 2.5~\cite{seedance25}}}\hfill
\makebox[\imgw]{\small\textbf{Wan 2.2 T2V A14B~\cite{wan2025}}}

\caption{We evaluate several state-of-the-art video generation models by specifying camera-effect parameters in input text prompts. The results show that these models often fail to generate videos whose camera effects correspond to the specified physical parameters: fisheye outputs do not reflect the requested field of view, depth-of-field results remain blurry even at small aperture size, and rolling-shutter videos lack the expected skew under large readout times. \ifarxiv Please refer to Appendix~\ref{supp:video_gen} for details.\fi}
\label{fig:pretrained_video_generation}

\vspace{-1em}
\end{figure*}

Camera effects are an integral part of image formation, and their appearance is governed by physical camera parameters. In videos, these effects also interact with motion, such as rolling-shutter distortion~\cite{zhong2021towards, saurer2013rolling}. Video generation models have recently emerged as a promising source of realistic and controllable visual data~\cite{genie3, he2025pretrainedvideogenerativemodels, hu2023gaia1generativeworldmodel, hu2024drivingworldconstructingworldmodel}. However, as shown in Fig.~\ref{fig:pretrained_video_generation}, current models do not consistently respond to requested camera settings. They may reproduce a plausible fisheye, depth-of-field, or rolling-shutter look, but adjusting the field of view, aperture, or readout time does not reliably yield the corresponding physical change. This limits their use as parameter-controlled data generators.

\vspace{-1mm}

One of the reasons for this limitation is that real dynamic videos with dependable camera-effect annotations are scarce, particularly across varied parameter values~\cite{gong2024beyond, fang2024camera, fang2025camerabench, wang2023survey}. Physics-based renderers such as Blender~\cite{blender_software} can provide exact metadata, but constructing realistic dynamic scenes is labor-intensive, and the resulting imagery can exhibit a sim-to-real gap~\cite{nikolenko2021synthetic, tremblay2018training}. Reconstruction offers another route: recover a dynamic scene from captured videos and then render new camera effects from its representation. This retains the captured appearance and motion while allowing camera parameters to be changed explicitly. Existing methods, however, are not well suited to this role. Dynamic NeRF approaches~\cite{pumarola2021d, park2021nerfies, gao2021dynamic, park2021hypernerf, fridovich2023k, fang2022fast, cao2023hexplane} support ray-based light transport but render slowly, whereas rasterization-based Dynamic Gaussian Splatting methods~\cite{luiten2024dynamic, yang2024deformable, wu20244d} are efficient but cannot model the light transport required by these effects.

To address these limitations, we propose 4D Gaussian Ray Tracing (4D-GRT), a two-stage data-generation pipeline that combines 4D Gaussian scene representations with physically based ray tracing. 4D-GRT first fits a 4D Gaussian representation to synchronized pinhole videos using differentiable ray tracing and then renders parameterized camera effects through the same framework. This design provides explicit camera control while retaining the efficiency and reconstruction quality of Gaussian scene representations. We evaluate 4D-GRT on CamVFX-4D, a controlled benchmark under four camera effects. Compared with MSTH, 4D-GRT improves PSNR by 2.1--3.4 dB and achieves a 1.4--6.0$\times$ rendering speedup across the evaluated settings. We further show that fine-tuning a video diffusion model with our generated videos improves its control over physical camera-effect parameters. We also demonstrate an extension to monocular input by using MoVieS~\cite{lin2026movies} for 4D-GS reconstruction, thereby expanding the video sources available for scalable camera-effect data generation.

Our contributions are summarized as follows:
\setlist[itemize]{itemsep=0pt, topsep=2pt}
\begin{itemize}
    \item \textbf{Problem Identification:} We identify that current video diffusion models can mimic camera-effect appearance without reliably responding to the associated physical parameters.
    \item \textbf{Data Generation Engine:} We introduce a reconstruction-based pipeline that couples 4D Gaussian representations with differentiable ray tracing to generate dynamic videos with precise camera-effect annotations, and show that fine-tuning a video diffusion model on these videos improves its camera-effect control.
    \item \textbf{CamVFX-4D Benchmark:} We build a controlled benchmark under four camera effects for evaluating parameter-conditioned dynamic rendering.
\end{itemize}

\vspace{-0.5em}
\section{Related Work}
\label{sec:related}
\vspace{-0.5em}

\subsection{Generating Visual Data with Camera Effects}
\label{subsec:generating_visual_data_with_camera_effect}
\vspace{-0.5em}

Generating dynamic visual data with accurate camera-effect parameter annotations is important for training parameter-controllable video models. Existing approaches mainly rely on synthetic rendering or generative models.

\noindent {\bf Synthetic Data Approaches.}
 Traditional approaches for generating camera effect data rely heavily on synthetic rendering pipelines using graphics software such as Blender~\cite{blender_software}. These methods provide precise control over camera parameters, generating numerically labeled data. However, such methods face several fundamental limitations: (1) high labor costs for creating realistic scenes, (2) a sim-to-real gap that limits model generalization.

\noindent {\bf Generative Model Approaches.}
The emergence of powerful generative models, particularly diffusion-based methods~\cite{rombach2022high, croitoru2023diffusion, blattmann2023stable, xing2024survey, ho2022imagen, chang2025gcc, ho2022video, chu2023video, yeh2024training}, has opened new avenues for camera effect synthesis~\cite{fang2024camera}. These approaches can generate diverse visual content with various camera effects. Recent representative works include Curved Diffusion~\cite{voynov2024curved} and AKiRa~\cite{wang2025akira}, which enable control over optical distortions or camera effect parameters. However, generative methods suffer from several critical limitations for training data generation: (1) most focus on specific camera effects rather than general effect synthesis, (2) lack of precise physical constraints leading to unrealistic camera effects, (3) inability to accurately control camera parameters, and (4) difficulty in ensuring multi-view consistency or temporal consistency in dynamic scenes.
 
These limitations motivate reconstruction-based approaches that retain the appearance and motion of captured dynamic scenes while allowing explicit camera control.

\vspace{-0.5em}
\subsection{Dynamic Neural Rendering}
\label{subsec:dynamic_neural_rendering}
\vspace{-0.5em}

Dynamic neural rendering reconstructs time-varying scenes from captured observations, providing a basis for applying new camera models to real scene appearance and motion.

\noindent {\bf NeRF.} Neural Radiance Fields (NeRF)~\cite{mildenhall2021nerf} introduced a new paradigm for representing static scenes, modeling them as continuous volumetric functions with MLPs that map 3D coordinates and viewing directions to density and color for photorealistic view synthesis. Numerous extensions~\cite{barron2021mip, zhang2020nerf++, yu2021pixelnerf, muller2022instant, SunSC22, barron2022mipnerf360, su2024boostmvsnerfs, lin2025frugalnerf} address NeRF’s limitations, including adaptations for dynamic scenes~\cite{pumarola2021d, park2021nerfies, park2021hypernerf, gao2021dynamic, fang2022fast, liu2023robust, yan2023nerf, nerfplayer, gan2023v4d, cao2023hexplane, wang2023masked, cheng2024improving}. Among these, K-Planes~\cite{fridovich2023k}, HexPlane~\cite{cao2023hexplane}, and MSTH~\cite{wang2023masked} achieve state-of-the-art performance by leveraging plane- or grid-based representations to accelerate dynamic scene reconstruction while maintaining high rendering quality.

\noindent {\bf Gaussian Splatting.}
Despite NeRF’s impressive results, its high computational cost and implicit representation limit practical use. 3D Gaussian Splatting (3D-GS)~\cite{kerbl20233d} addresses these issues with an explicit 3D Gaussian~\cite{ke2025stealthattack} representation and efficient rasterization, enabling real-time, high-quality rendering~\cite{10887619, zhan2025cat}. Recent works \cite{yang2024deformable, wu20244d, liang2024gaufregaussiandeformationfields, huang2023sc, lin2024gaussian, kratimenos2024dynmf, yang2024realtimephotorealisticdynamicscene, fan2025spectromotion} extend 3D-GS to dynamic scenes. For example, Deformable 3D Gaussians~\cite{yang2024deformable} introduce deformation modeling, while 4D Gaussian Splatting (4D-GS)~\cite{wu20244d} employs a spatio-temporal encoder composed of multi-resolution voxel planes and MLPs to model the deformation of 3D Gaussians. These advances preserve the efficiency of Gaussian Splatting while enabling high-quality dynamic scene reconstruction, establishing a strong foundation for camera effect simulation.

\begin{figure*}[t]
    \centering
    \includegraphics[width=\textwidth]{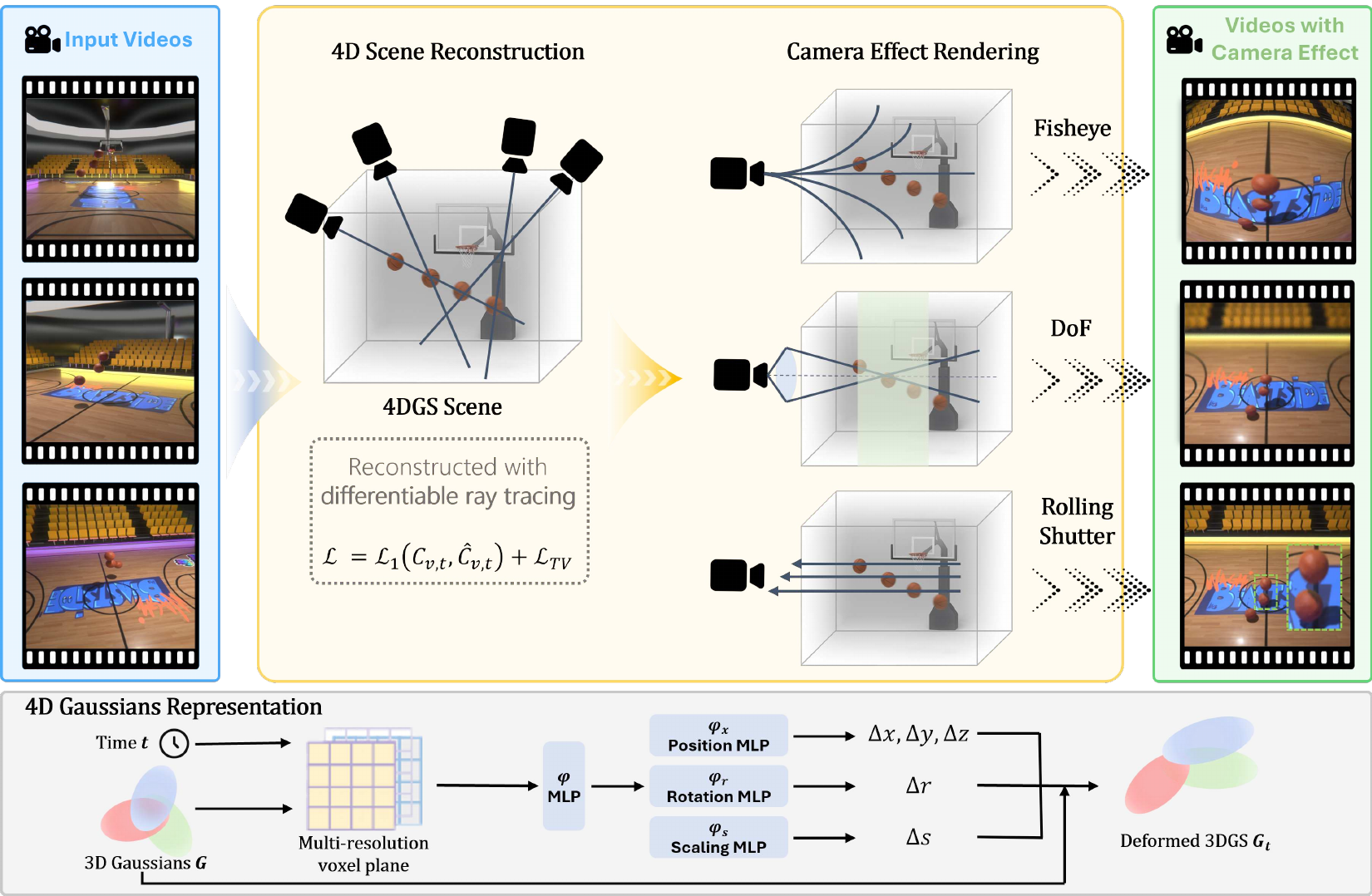}
    \caption{
    The overall pipeline of 4D-GRT. Given multi-view videos, we optimize the 4D-GS representation through differentiable ray tracing. Then, given camera-effect parameters, we can utilize ray tracing to render videos with physically grounded camera effects.
    }
    \label{fig:pipeline}
    \vspace{-1.1em}
\end{figure*}

\vspace{-0.5em}
\subsection{Neural Camera Effect Rendering}
\label{subsec:ray_tracing_and_camera_effect}
\vspace{-0.5em}

Traditional renderers simulate camera effects through ray tracing~\cite{10.1145/964965.808590}, reproducing physical optics. In neural scene representations, NeRF can achieve similar effects~\cite{Wu_2022, niu2024rsnerfneuralradiancefields} due to its ray-casting-based rendering, but NeRF-based methods are computationally expensive and lack efficient rendering. In contrast, 3D-GS offers fast rasterization but is restricted to the pinhole camera model, requiring specialized designs and complex Jacobian computations to support effects such as fisheye distortion~\cite{liao2024fisheyegslightweightextensiblegaussian}, rolling shutter with motion blur~\cite{seiskari2024gaussian, trgs_slam_2026}, and depth of field~\cite{lee2025cocogaussian, wang2025dofgsadjustabledepthoffield3dgaussian, shen2025dofgaussiancontrollabledepthoffield3d, dydeblur}. 
To overcome the limitations of the standard pinhole model in 3D-GS, recent methods such as 3DGRT~\cite{3dgrt2024}, 3DGUT~\cite{wu20253dgut}, and 3DGEER~\cite{huang20263dgeer3dgaussianrendering} have introduced advanced ray-tracing and projection techniques to enable physically based, flexible camera effect simulation. However, these methods are restricted to static scenes and cannot simulate camera effects in dynamic settings. To address this, our method integrates 4D-GS with 3DGRT, leveraging ray tracing to enable camera effect rendering in dynamic scenes.

\vspace{-0.5em}
\section{Method}
\label{sec:method}

\vspace{-0.5em}

As illustrated in Fig.~\ref{fig:pipeline}, our proposed 4D Gaussian Ray Tracing (4D-GRT) framework generates physically plausible training data in two stages: dynamic scene reconstruction (Section~\ref{subsec:dynamic_scene_reconstruction}) and camera-effect rendering  (Section~\ref{subsec:camera_effect_rendering}). In the first stage, we reconstruct a dynamic scene from synchronized multi-view videos using a 4D-GS~\cite{wu20244d} model with differentiable ray tracing~\cite{3dgrt2024}. In the second stage, we render the scene with physically based camera models, including fisheye distortion, depth of field, and rolling shutter, producing physically based videos for effect-aware vision training.

\vspace{-1.5mm}
\subsection{Preliminary}
\label{subsec:preliminary}
\vspace{-0.5em}
\noindent{\bf 3D Gaussian Splatting (3D-GS).} 3D-GS~\cite{kerbl20233d} represents a static 3D scene as a collection of 3D Gaussians, each characterized by its mean position $(x, y, z)$, a covariance matrix $\mathbf{\Sigma}$ decomposed into scaling $s$ and rotation $r$ parameters, opacity $\sigma$, and spherical harmonic coefficients $\mathcal{C}$. Rendering is done by differentiable splatting~\cite{Yifan:DSS:2019}, where $\mathbf{\Sigma}$ is projected into 2D as $\mathbf{\Sigma}' = \mathbf{J} \mathbf{W} \mathbf{\Sigma} \mathbf{W}^T \mathbf{J}^T$, with $\mathbf{J}$ and $\mathbf{W}$ denoting the Jacobian of the projective transformation and the viewing transformation, respectively. Pixel colors are then computed via alpha compositing of the projected Gaussians from front to back, 
$C = \sum_{i \in \mathcal{N}} T_i \alpha_i c_i$, 
where $T_i = \prod_{j=1}^{i-1} (1 - \alpha_j)$ is the transmittance, $\alpha_i = \sigma_i \exp(-\tfrac{1}{2} \bm{x}^T \mathbf{\Sigma}'^{-1} \bm{x})$ is the opacity, and $c_i$ is the view-dependent color derived from spherical harmonics.

\subsection{Dynamic Scene Reconstruction}
\label{subsec:dynamic_scene_reconstruction}
\vspace{-0.5em}
As shown in Fig.~\ref{fig:pipeline}, we reconstruct scene dynamics from multi-view videos $\{ \hat{C}_{v,t} \mid v \in \mathcal{V}, t \in \mathcal{T} \}$, where $\mathcal{V}$ denotes the set of camera views and $\mathcal{T}$ the set of time indices. Following 4D-GS~\cite{wu20244d}, we represent the scene with a set of 3D Gaussians $G$ and a Gaussian deformation field network, which consists of a spatio-temporal structure encoder and a multi-head Gaussian deformation decoder to model the per-frame deformations of 3D Gaussians. Specifically, to compute the deformation of 3D Gaussians $G$ at time $t$, we first employ a set of 4D voxel planes $R_l$ to extract voxel features based on each Gaussian's mean position and time step $t$. These features are first fused by a lightweight MLP $\varphi$, and then passed to separate MLPs $(\varphi_x, \varphi_r, \varphi_s)$ that predict the residuals of the Gaussian attributes $((\Delta x, \Delta y, \Delta z), \Delta r, \Delta s)$. Finally, applying the residuals to $G$ yields the deformed Gaussians $G_t$. 

For rendering, although 3D-GS~\cite{kerbl20233d} rasterization provides efficient and high-quality results, it is inherently limited in its ability to simulate physically based camera effects. In particular, phenomena that require explicit ray-based modeling, such as complex lens distortions, cannot be faithfully reproduced. To address these limitations, we adopt the differentiable ray tracing framework of 3DGRT~\cite{3dgrt2024}, which traces rays directly through 3D Gaussian primitives instead of projecting them onto the image plane. By combining a $k$-buffer hit-based marching scheme with the hardware-accelerated NVIDIA OptiX interface, this approach achieves both the accuracy required for realistic camera-effect rendering and the efficiency necessary for large-scale data generation. 

By integrating the dynamic scene model with differentiable ray tracing, we reconstruct dynamic scenes end-to-end via training with a color loss. At each training iteration, we randomly sample a view $v \in \mathcal{V}$ and a time $t \in \mathcal{T}$, obtain the deformed Gaussians $G_t$, and render them via ray tracing from view $v$ to obtain the image $C_{v,t}$. We then compute the L1 loss between $C_{v,t}$ and the ground-truth image $\hat{C}_{v,t}$, denoted by $\mathcal{L}_1$. Additionally, we apply a grid-based total variation loss $\mathcal{L}_{\mathrm{TV}}$ following~\cite{wu20244d, cao2023hexplane}. The overall loss function is defined as
\begin{align}
\mathcal{L} = \mathcal{L}_1(C_{v, t}, \hat{C}_{v, t}) + \mathcal{L}_{\mathrm{TV}}.
\label{eq:loss}
\end{align}

\subsection{Camera-Effect Rendering}
\label{subsec:camera_effect_rendering}
\vspace{-0.5em}

After reconstructing the dynamic scene, we synthesize common camera effects by integrating physical camera models into a ray-tracing renderer. Although we focus on three representative effects in this work, the ray-tracing formulation is general and can naturally support additional camera effects by incorporating their corresponding physical camera models. Specifically, we implement:

\noindent {\bf Fisheye.} To simulate wide-angle distortion, we adopt the 4th-degree polynomial radial model of Blender~\cite{blender_manual_45}.

The model is defined as:
\begin{align}
\theta &= k_0 + k_1r + k_2r^2 + k_3r^3 + k_4r^4,
\label{eq:fisheye}
\end{align}

\noindent where $r$ is the radial distance from the sensor's principal point, $\theta$ is the incoming ray’s polar angle, and $k_i$ are the lens coefficients. We convert image coordinates to physical sensor coordinates $(x, y)$, compute $r = \sqrt{x^2 + y^2}$, and evaluate the polynomial to find $\theta$ and the azimuthal angle $\phi = \operatorname{atan2}(y, x)$. These angles define the spherical ray direction for the ray tracer, ensuring a physically consistent, parameter-based distortion across multiple views.

\noindent {\bf Depth of Field (DoF).}
\label{eq:depth_of_field}
Following~\cite{10.1145/964965.808590}, we simulate DoF by computing the intersection $\mathbf{p} = \mathbf{o} + f_z \mathbf{d}$ of an ideal pinhole ray with the focal plane at focus distance $f_z$. We jitter the ray origin over a circular aperture of radius $r_a$ by uniformly sampling $\boldsymbol{\ell} = (\ell_x, \ell_y)$ and mapping it to world space via the lens-plane basis $(\mathbf{u}, \mathbf{v})$. 

The perturbed rays are then computed as
\begin{equation}
\mathbf{o}' = \mathbf{o} + \ell_x \mathbf{u} + \ell_y \mathbf{v}, 
\qquad
\mathbf{d}' = \frac{\mathbf{p} - \mathbf{o}'}{\lVert \mathbf{p} - \mathbf{o}' \rVert}.
\label{eq:dof}
\end{equation}
Averaging the traced radiance of multiple samples per pixel yields physically based defocus blur.

\noindent {\bf Rolling Shutter.}
\label{eq:rolling_shutter}
To simulate sequential row capture, we compute a specific sensing time $t_r$ for each pixel row $r$, evaluating the deformed 3D Gaussians $G_{t_r}$ to trace a single ray per pixel. Because varying sensing times across rows prevents full parallelization of the ray tracing, we adopt a chunking approximation to maintain rendering efficiency. We divide rows into chunks of size $N_c$ and approximate the sensing time for all rows in a chunk by their average. This allows us to compute the deformed Gaussians once per chunk and trace its rays in parallel, greatly improving speed while limiting blocking artifacts.

\vspace{-0.5em}
\section{Experiments}
\label{sec:experiments}
\vspace{-0.5em}
\newcommand\QualitativeCompareBoxWidth{.15\textwidth}
\newcommand\QualitativeCompareImageWidth{.15\textwidth}
\begin{figure*}[t]
    \centering
    \small
    \resizebox{1\textwidth}{!}{
    \begin{tabular}{@{}c@{\hspace{1pt}}c@{}c@{}c@{}c@{}c@{}c@{}c@{}c@{}}
    
    Pinhole &
    \parbox[c]{\QualitativeCompareBoxWidth}{
    \includegraphics[width=\QualitativeCompareImageWidth]{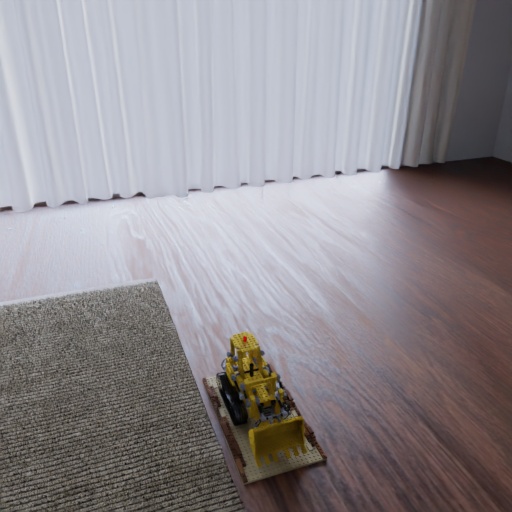} 
    } & 
    \parbox[c]{\QualitativeCompareBoxWidth}{
    \includegraphics[width=\QualitativeCompareImageWidth]{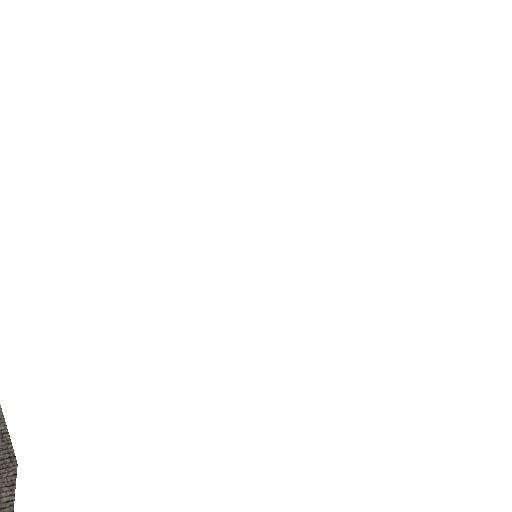}
    } & 
    \parbox[c]{\QualitativeCompareBoxWidth}{
    \includegraphics[width=\QualitativeCompareImageWidth]{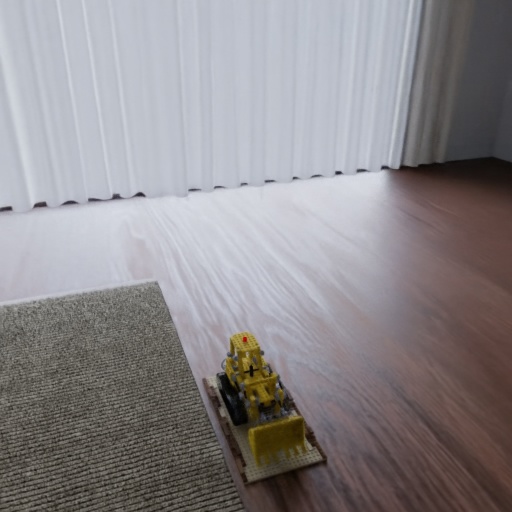} 
    } & 
    \parbox[c]{\QualitativeCompareBoxWidth}{
    \includegraphics[width=\QualitativeCompareImageWidth]{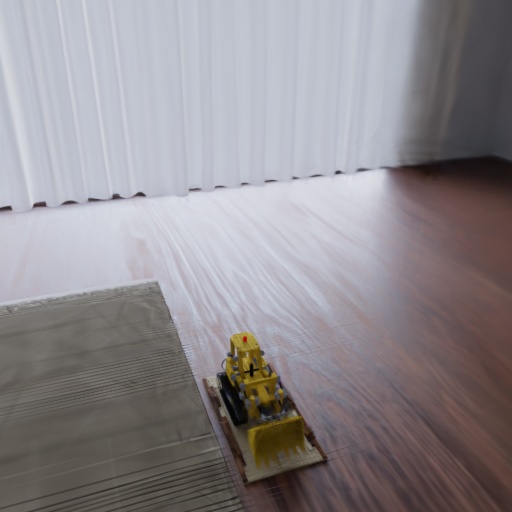} 
    } & 
    \parbox[c]{\QualitativeCompareBoxWidth}{
    \includegraphics[width=\QualitativeCompareImageWidth]{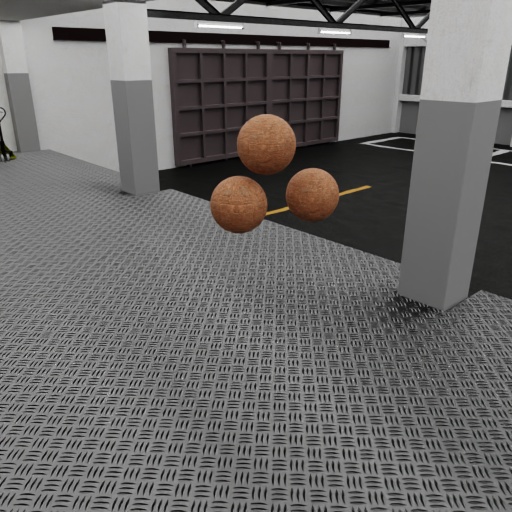} 
    } & 
    \parbox[c]{\QualitativeCompareBoxWidth}{
    \includegraphics[width=\QualitativeCompareImageWidth]{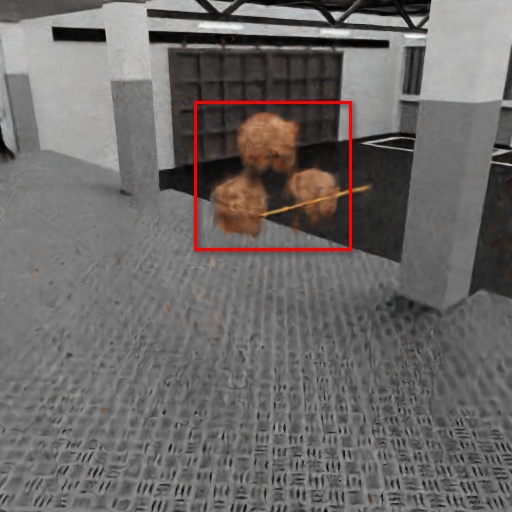} 
    } & 
    \parbox[c]{\QualitativeCompareBoxWidth}{
    \includegraphics[width=\QualitativeCompareImageWidth]{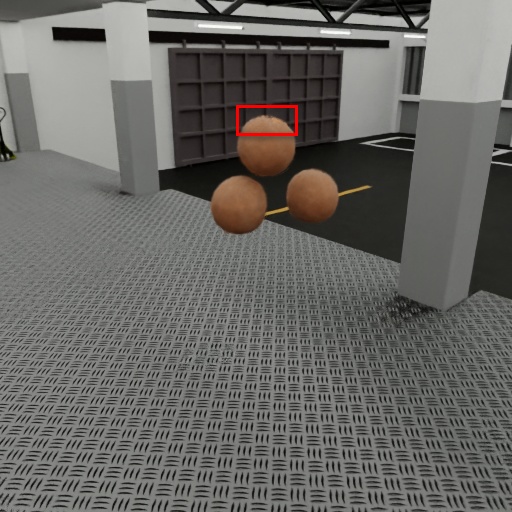} 
    } & 
    \parbox[c]{\QualitativeCompareBoxWidth}{
    \includegraphics[width=\QualitativeCompareImageWidth]{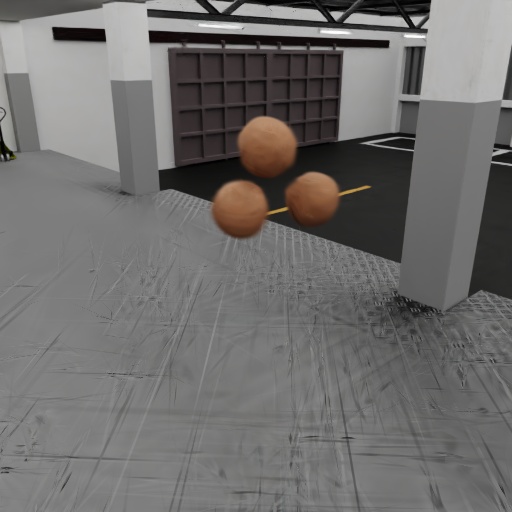} 
    } \\ 

   \begin{tabular}{c}
    Rolling \\
    Shutter
    \end{tabular}&
    \parbox[c]{\QualitativeCompareBoxWidth}{
    \includegraphics[width=\QualitativeCompareImageWidth]{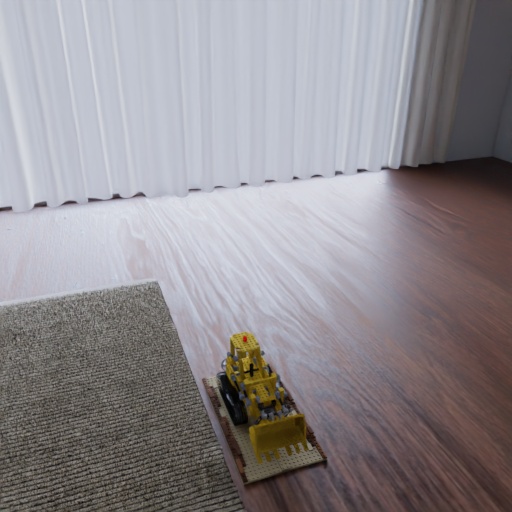} 
    } & 
    \parbox[c]{\QualitativeCompareBoxWidth}{
    \includegraphics[width=\QualitativeCompareImageWidth]{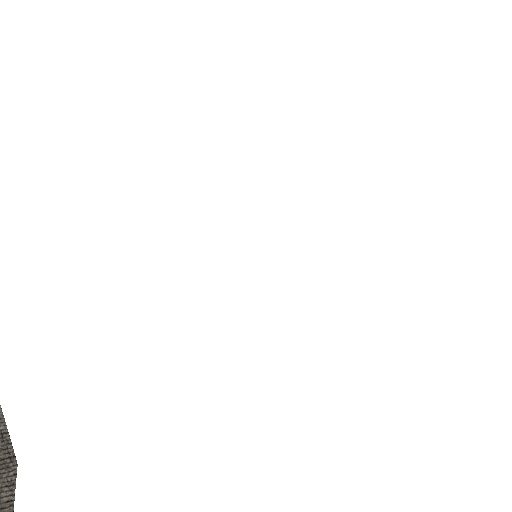}
    } & 
    \parbox[c]{\QualitativeCompareBoxWidth}{
    \includegraphics[width=\QualitativeCompareImageWidth]{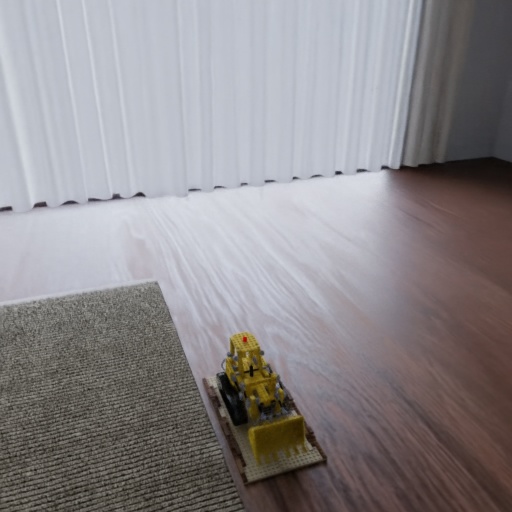} 
    } & 
    \parbox[c]{\QualitativeCompareBoxWidth}{
    \includegraphics[width=\QualitativeCompareImageWidth]{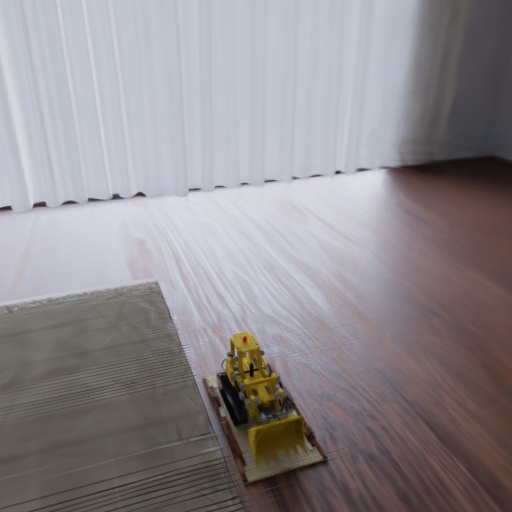} 
    } & 
    \parbox[c]{\QualitativeCompareBoxWidth}{
    \includegraphics[width=\QualitativeCompareImageWidth]{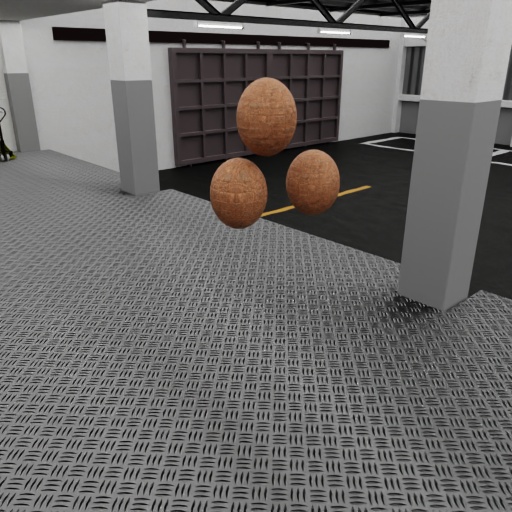} 
    } & 
    \parbox[c]{\QualitativeCompareBoxWidth}{
    \includegraphics[width=\QualitativeCompareImageWidth]{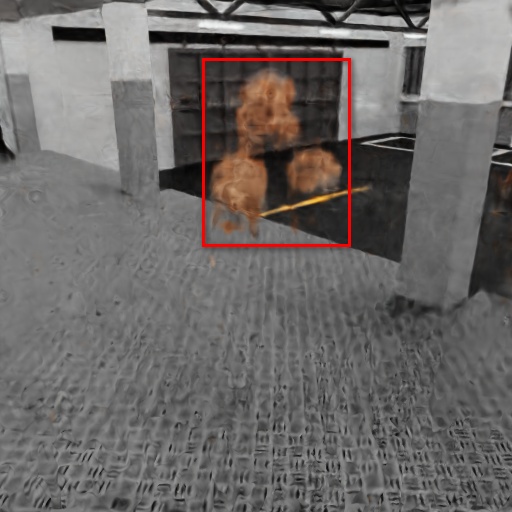} 
    } & 
    \parbox[c]{\QualitativeCompareBoxWidth}{
    \includegraphics[width=\QualitativeCompareImageWidth]{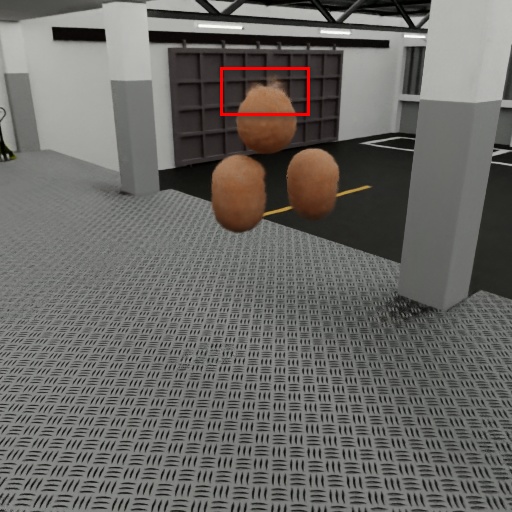} 
    } & 
    \parbox[c]{\QualitativeCompareBoxWidth}{
    \includegraphics[width=\QualitativeCompareImageWidth]{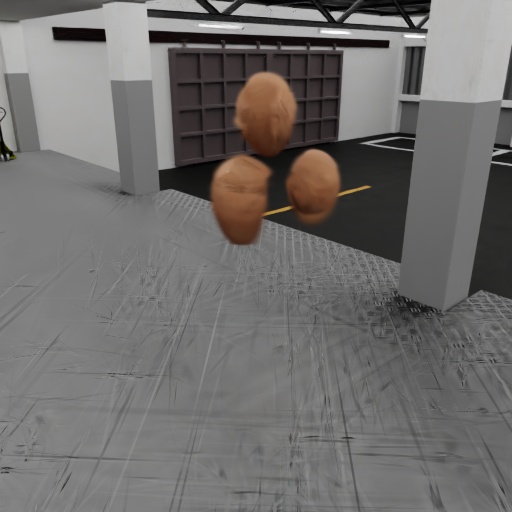} 
    } \\ 

   \begin{tabular}{c}
    Depth \\
    of \\
    Field
    \end{tabular}&
    \parbox[c]{\QualitativeCompareBoxWidth}{
    \includegraphics[width=\QualitativeCompareImageWidth]{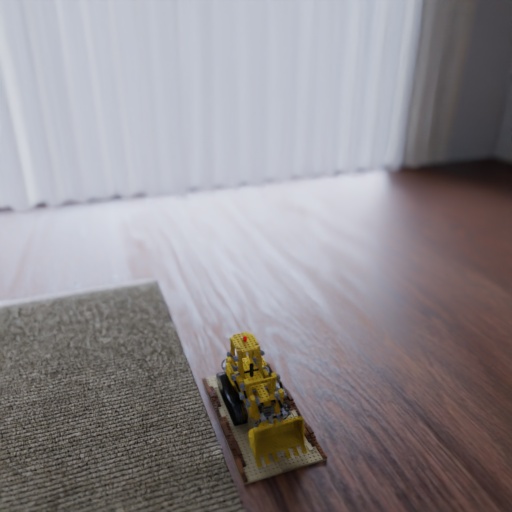} 
    } & 
    \parbox[c]{\QualitativeCompareBoxWidth}{
    \includegraphics[width=\QualitativeCompareImageWidth]{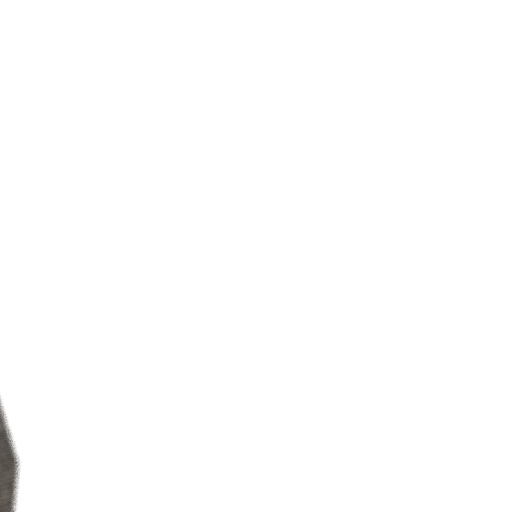}
    } & 
    \parbox[c]{\QualitativeCompareBoxWidth}{
    \includegraphics[width=\QualitativeCompareImageWidth]{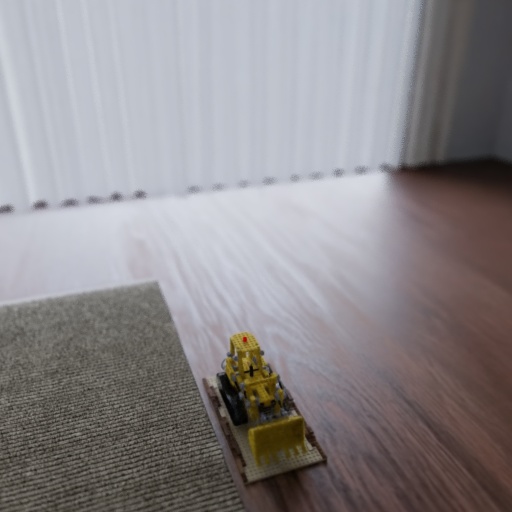} 
    } & 
    \parbox[c]{\QualitativeCompareBoxWidth}{
    \includegraphics[width=\QualitativeCompareImageWidth]{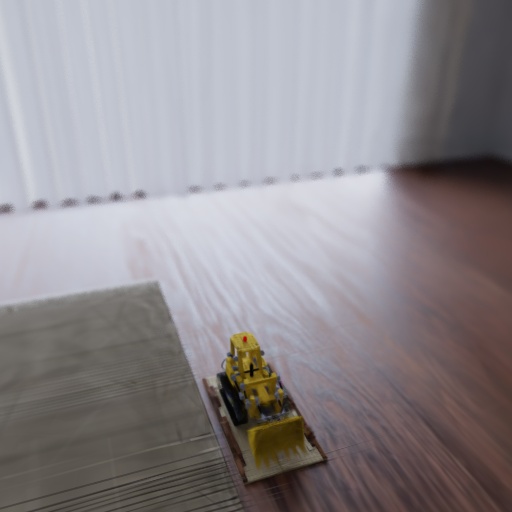} 
    } & 
    \parbox[c]{\QualitativeCompareBoxWidth}{
    \includegraphics[width=\QualitativeCompareImageWidth]{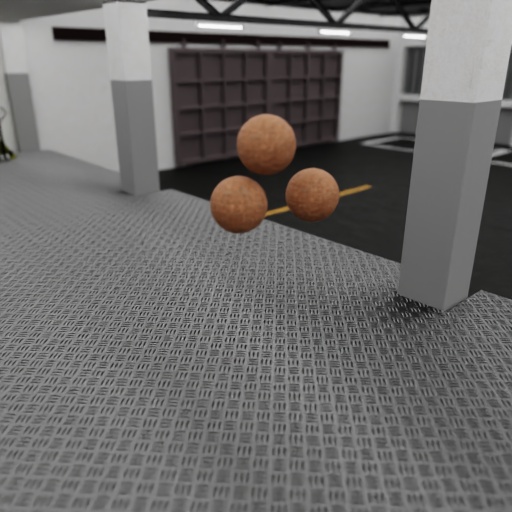} 
    } & 
    \parbox[c]{\QualitativeCompareBoxWidth}{
    \includegraphics[width=\QualitativeCompareImageWidth]{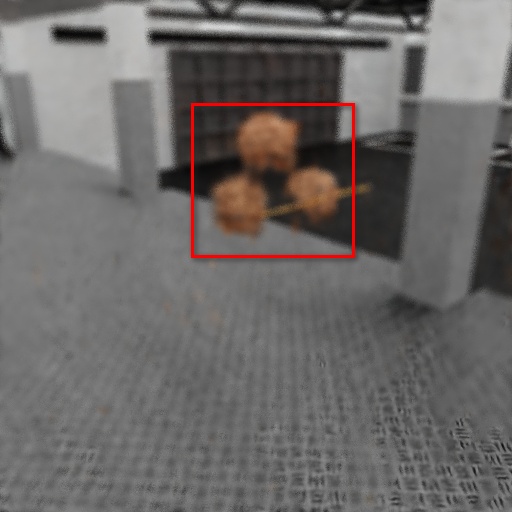} 
    } & 
    \parbox[c]{\QualitativeCompareBoxWidth}{
    \includegraphics[width=\QualitativeCompareImageWidth]{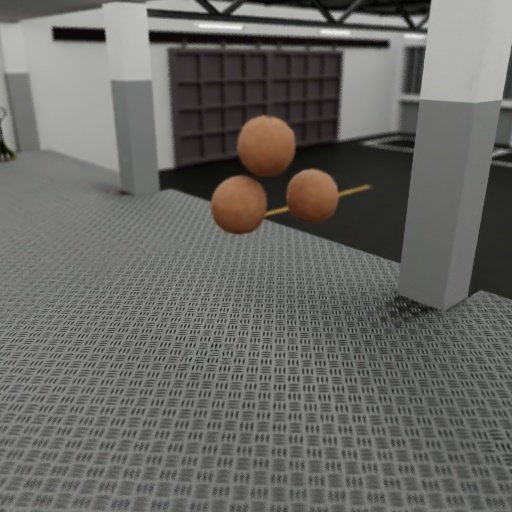} 
    } & 
    \parbox[c]{\QualitativeCompareBoxWidth}{
    \includegraphics[width=\QualitativeCompareImageWidth]{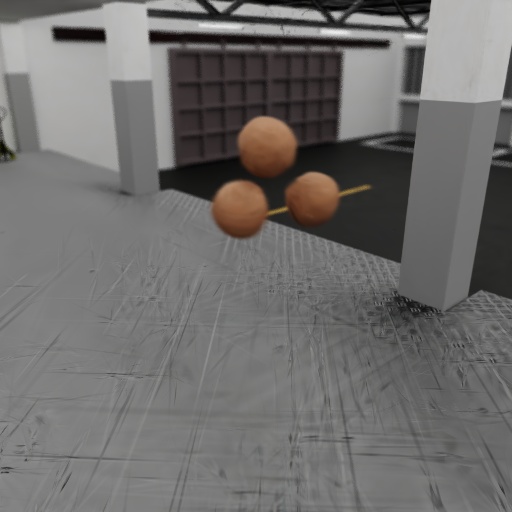} 
    } \\ 

    Fisheye &
    \parbox[c]{\QualitativeCompareBoxWidth}{
    \includegraphics[width=\QualitativeCompareImageWidth]{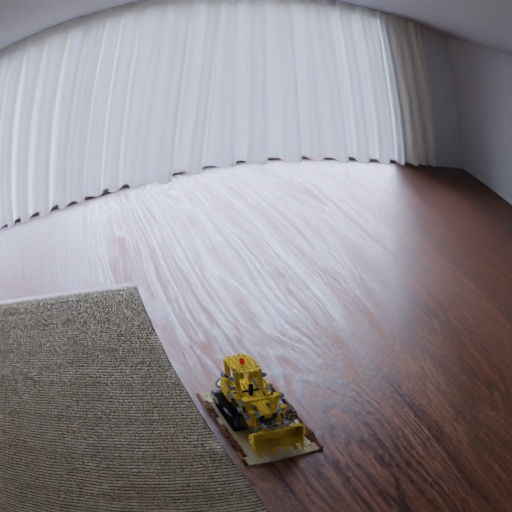} 
    } & 
    \parbox[c]{\QualitativeCompareBoxWidth}{
    \includegraphics[width=\QualitativeCompareImageWidth]{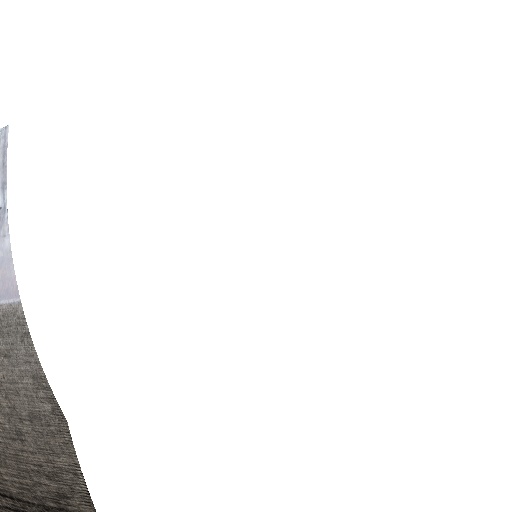}
    } & 
    \parbox[c]{\QualitativeCompareBoxWidth}{
    \includegraphics[width=\QualitativeCompareImageWidth]{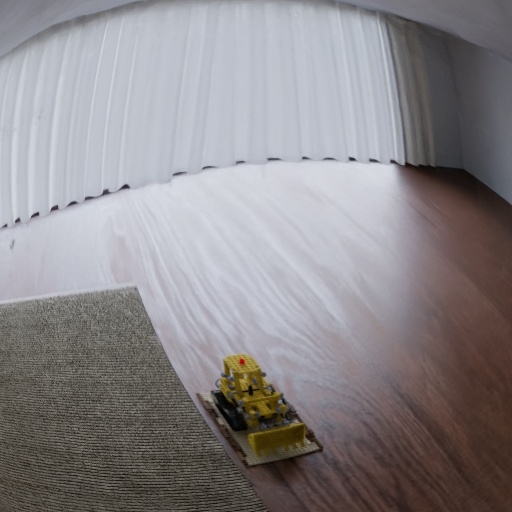} 
    } & 
    \parbox[c]{\QualitativeCompareBoxWidth}{
    \includegraphics[width=\QualitativeCompareImageWidth]{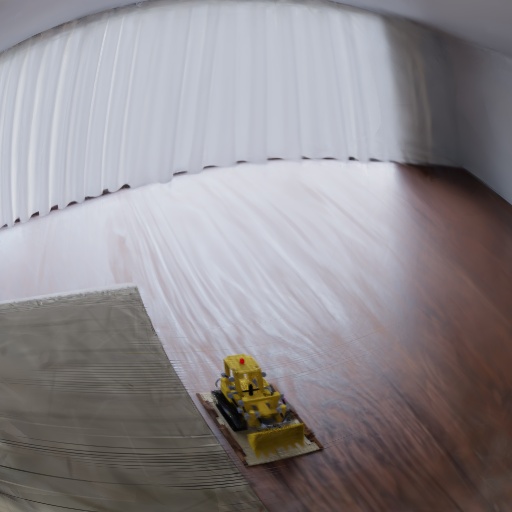} 
    } & 
    \parbox[c]{\QualitativeCompareBoxWidth}{
    \includegraphics[width=\QualitativeCompareImageWidth]{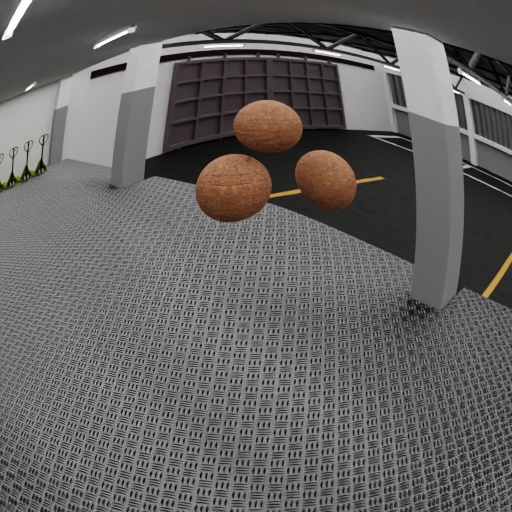} 
    } & 
    \parbox[c]{\QualitativeCompareBoxWidth}{
    \includegraphics[width=\QualitativeCompareImageWidth]{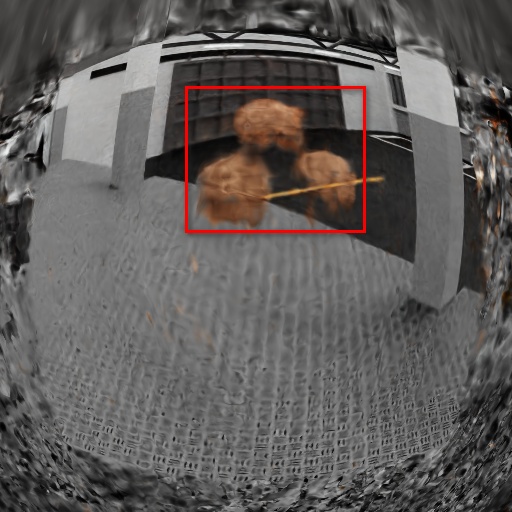} 
    } & 
    \parbox[c]{\QualitativeCompareBoxWidth}{
    \includegraphics[width=\QualitativeCompareImageWidth]{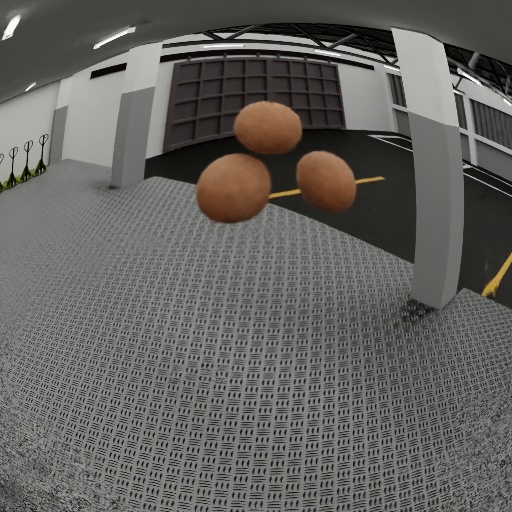} 
    } & 
    \parbox[c]{\QualitativeCompareBoxWidth}{
    \includegraphics[width=\QualitativeCompareImageWidth]{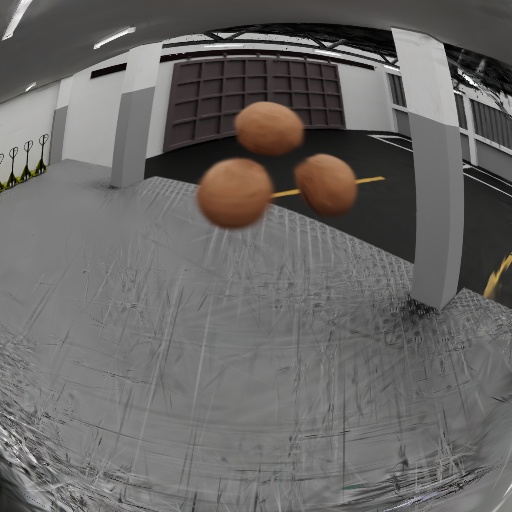} 
    } \\ 
    
     & Ground Truth & HexPlane~\cite{cao2023hexplane} & MSTH~\cite{wang2023masked} & 4D-GRT(Ours) & Ground Truth & HexPlane~\cite{cao2023hexplane} & MSTH~\cite{wang2023masked} & 4D-GRT(Ours) \\

    Pinhole &
    \parbox[c]{\QualitativeCompareBoxWidth}{
    \includegraphics[width=\QualitativeCompareImageWidth]{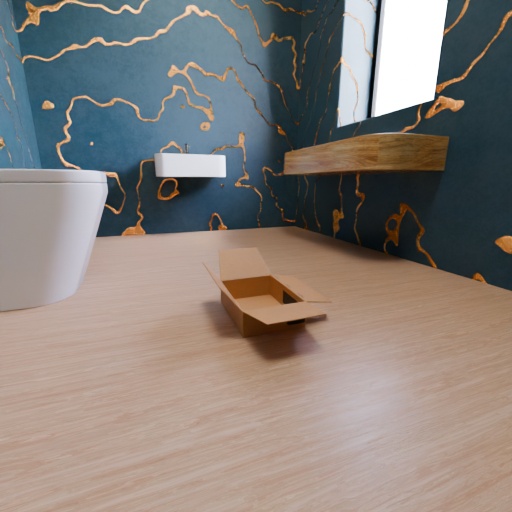} 
    } & 
    \parbox[c]{\QualitativeCompareBoxWidth}{
    \includegraphics[width=\QualitativeCompareImageWidth]{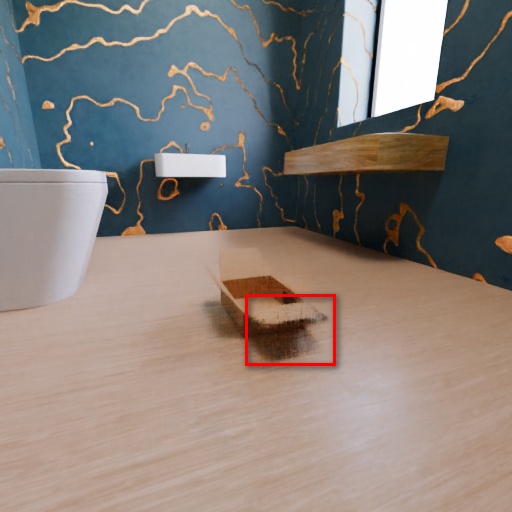}
    } & 
    \parbox[c]{\QualitativeCompareBoxWidth}{
    \includegraphics[width=\QualitativeCompareImageWidth]{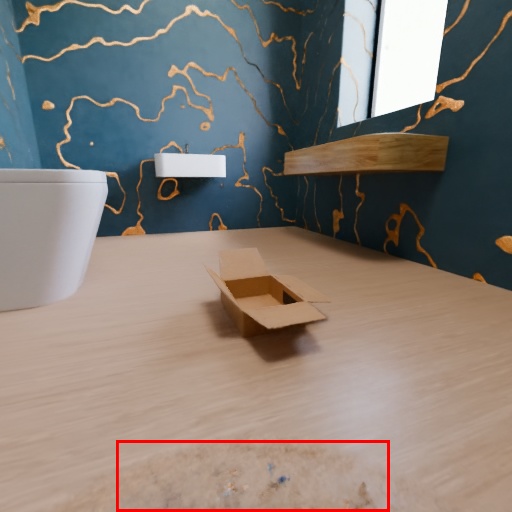} 
    } & 
    \parbox[c]{\QualitativeCompareBoxWidth}{
    \includegraphics[width=\QualitativeCompareImageWidth]{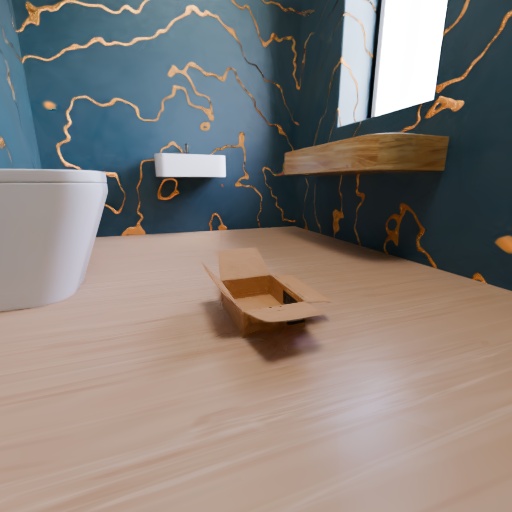} 
    } & 
    \parbox[c]{\QualitativeCompareBoxWidth}{
    \includegraphics[width=\QualitativeCompareImageWidth]{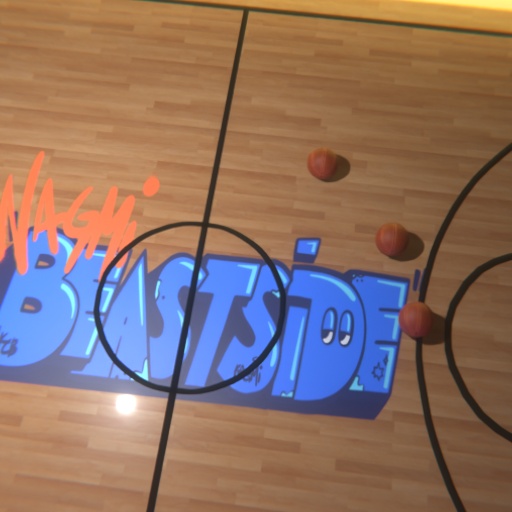} 
    } & 
    \parbox[c]{\QualitativeCompareBoxWidth}{
    \includegraphics[width=\QualitativeCompareImageWidth]{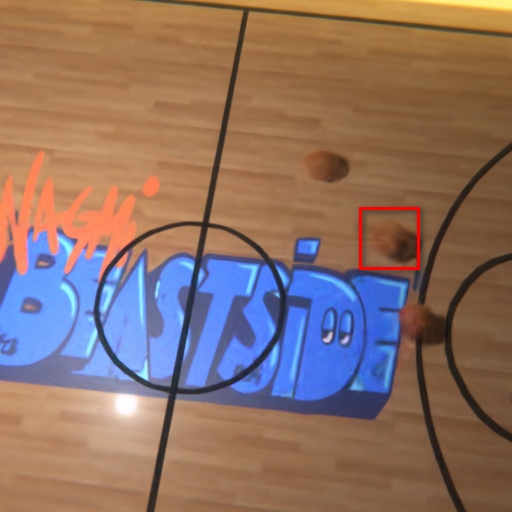} 
    } & 
    \parbox[c]{\QualitativeCompareBoxWidth}{
    \includegraphics[width=\QualitativeCompareImageWidth]{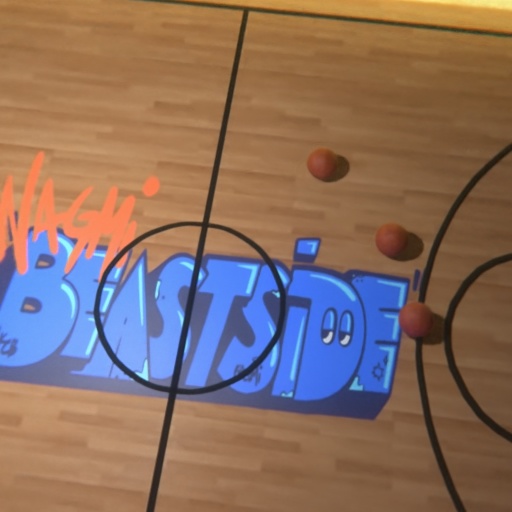} 
    } & 
    \parbox[c]{\QualitativeCompareBoxWidth}{
    \includegraphics[width=\QualitativeCompareImageWidth]{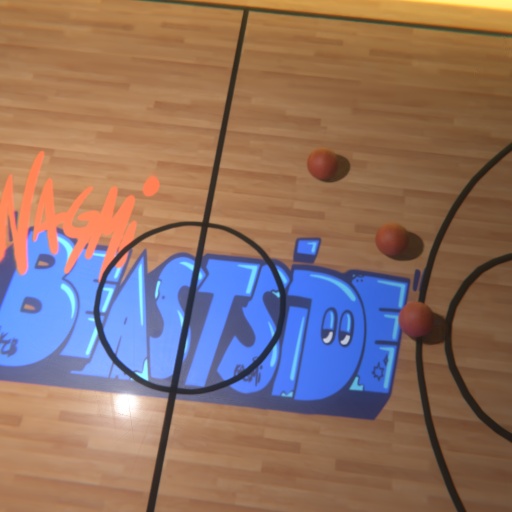} 
    } \\ 

   \begin{tabular}{c}
    Rolling \\
    Shutter
    \end{tabular}&
    \parbox[c]{\QualitativeCompareBoxWidth}{
    \includegraphics[width=\QualitativeCompareImageWidth]{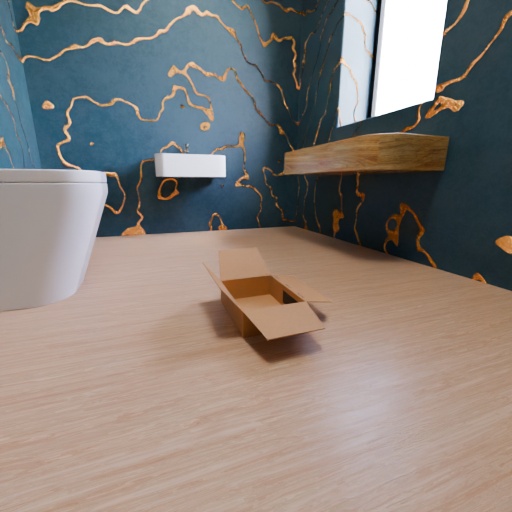} 
    } & 
    \parbox[c]{\QualitativeCompareBoxWidth}{
    \includegraphics[width=\QualitativeCompareImageWidth]{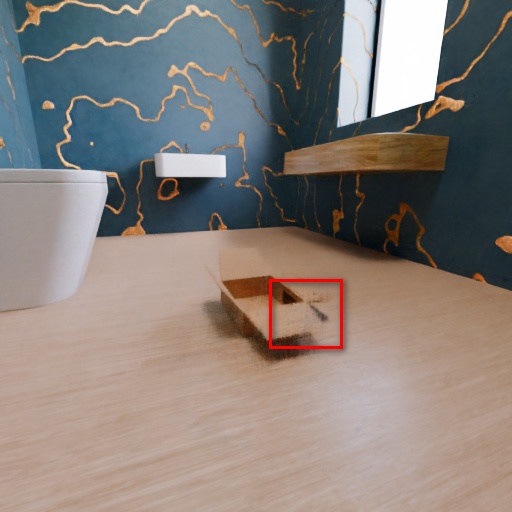}
    } & 
    \parbox[c]{\QualitativeCompareBoxWidth}{
    \includegraphics[width=\QualitativeCompareImageWidth]{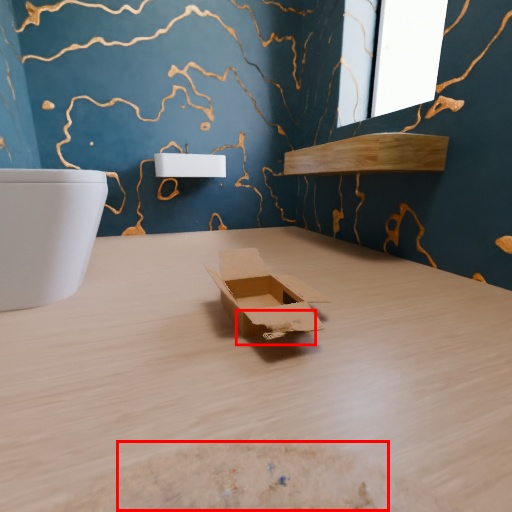} 
    } & 
    \parbox[c]{\QualitativeCompareBoxWidth}{
    \includegraphics[width=\QualitativeCompareImageWidth]{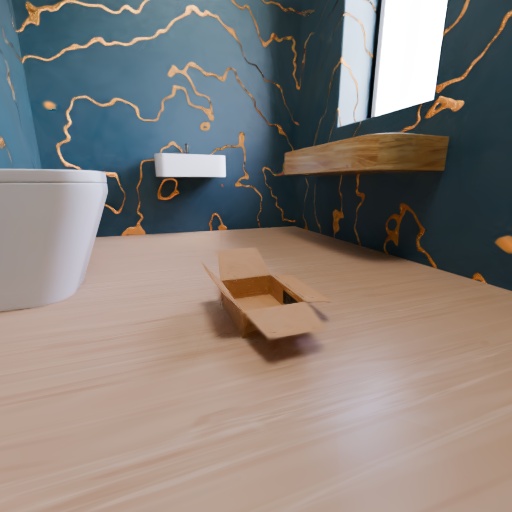} 
    } & 
    \parbox[c]{\QualitativeCompareBoxWidth}{
    \includegraphics[width=\QualitativeCompareImageWidth]{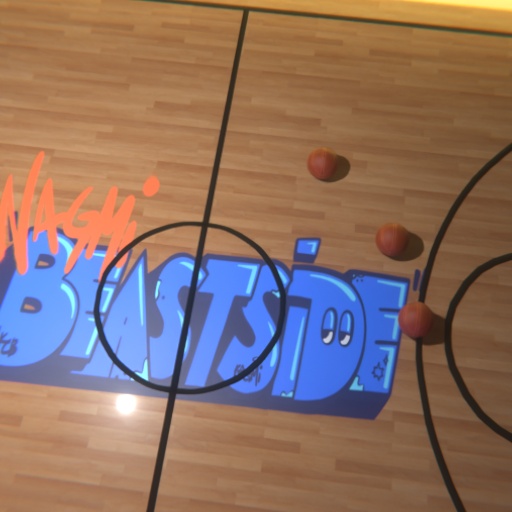} 
    } & 
    \parbox[c]{\QualitativeCompareBoxWidth}{
    \includegraphics[width=\QualitativeCompareImageWidth]{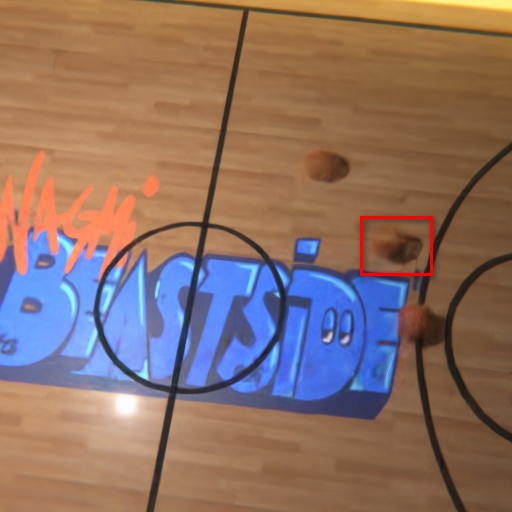} 
    } & 
    \parbox[c]{\QualitativeCompareBoxWidth}{
    \includegraphics[width=\QualitativeCompareImageWidth]{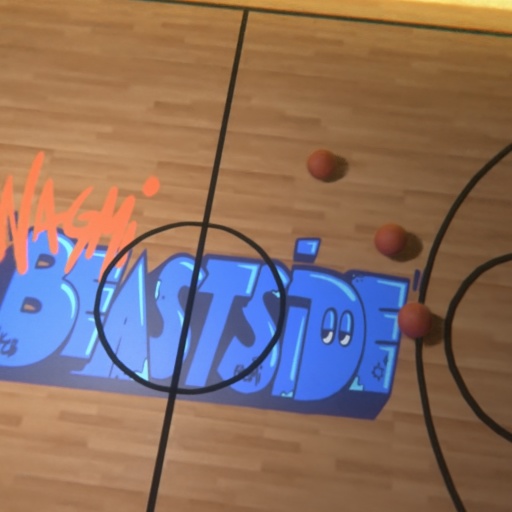} 
    } & 
    \parbox[c]{\QualitativeCompareBoxWidth}{
    \includegraphics[width=\QualitativeCompareImageWidth]{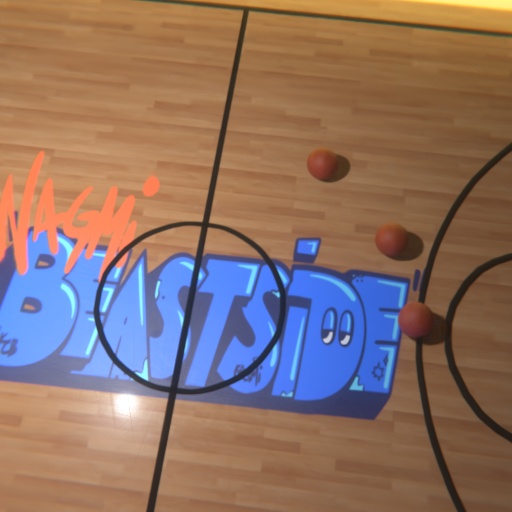} 
    } \\ 

   \begin{tabular}{c}
    Depth \\
    of \\
    Field
    \end{tabular}&
    \parbox[c]{\QualitativeCompareBoxWidth}{
    \includegraphics[width=\QualitativeCompareImageWidth]{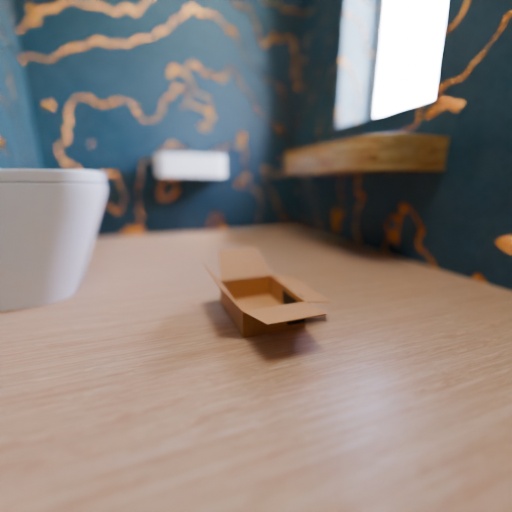} 
    } & 
    \parbox[c]{\QualitativeCompareBoxWidth}{
    \includegraphics[width=\QualitativeCompareImageWidth]{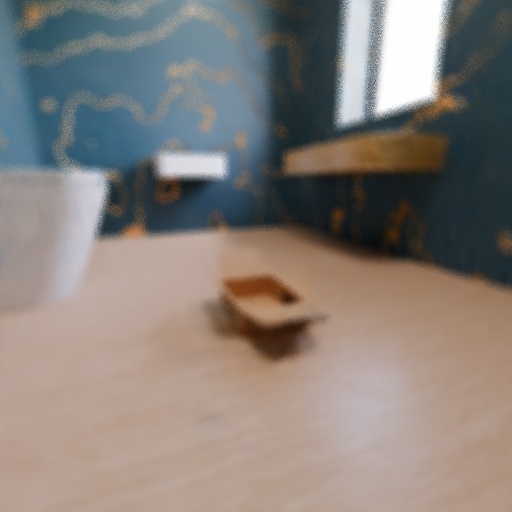}
    } & 
    \parbox[c]{\QualitativeCompareBoxWidth}{
    \includegraphics[width=\QualitativeCompareImageWidth]{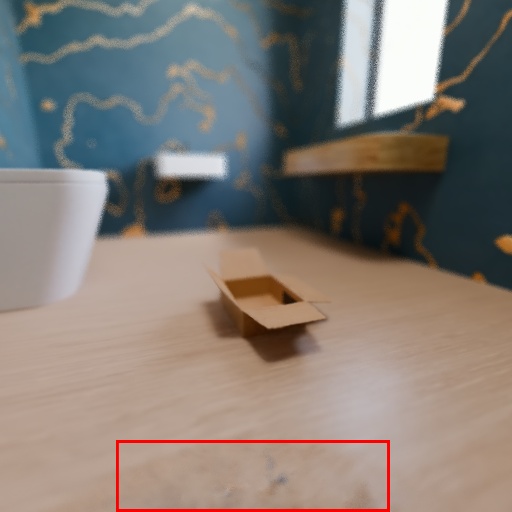} 
    } & 
    \parbox[c]{\QualitativeCompareBoxWidth}{
    \includegraphics[width=\QualitativeCompareImageWidth]{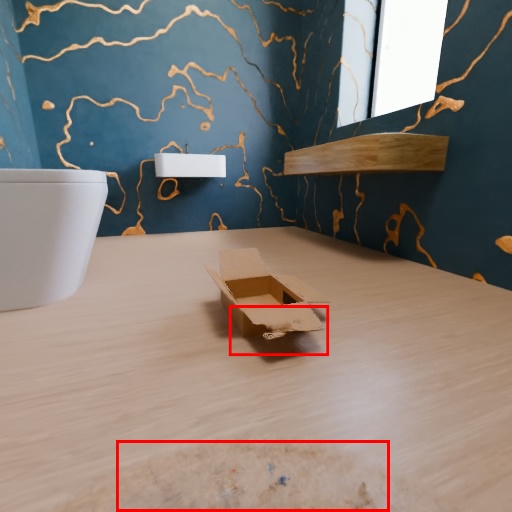} 
    } & 
    \parbox[c]{\QualitativeCompareBoxWidth}{
    \includegraphics[width=\QualitativeCompareImageWidth]{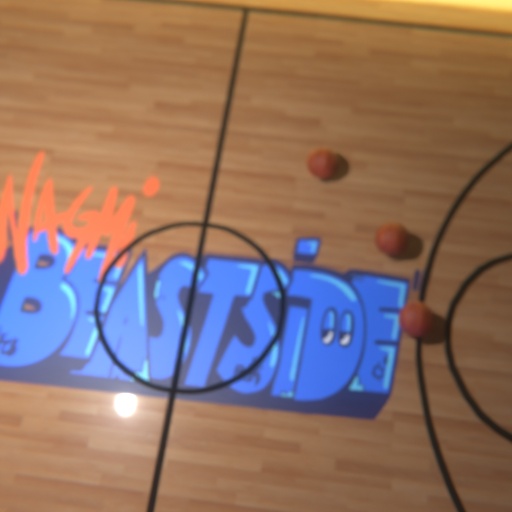} 
    } & 
    \parbox[c]{\QualitativeCompareBoxWidth}{
    \includegraphics[width=\QualitativeCompareImageWidth]{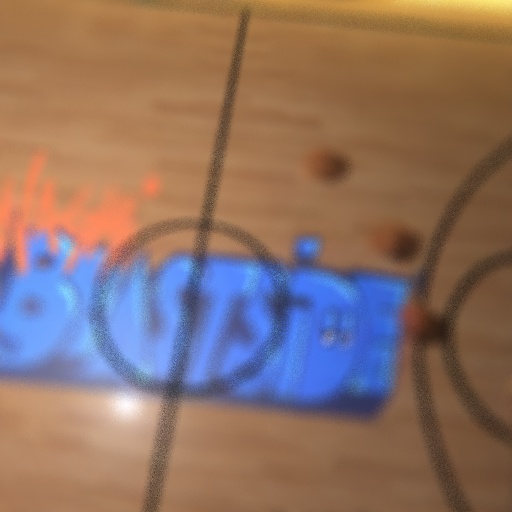} 
    } & 
    \parbox[c]{\QualitativeCompareBoxWidth}{
    \includegraphics[width=\QualitativeCompareImageWidth]{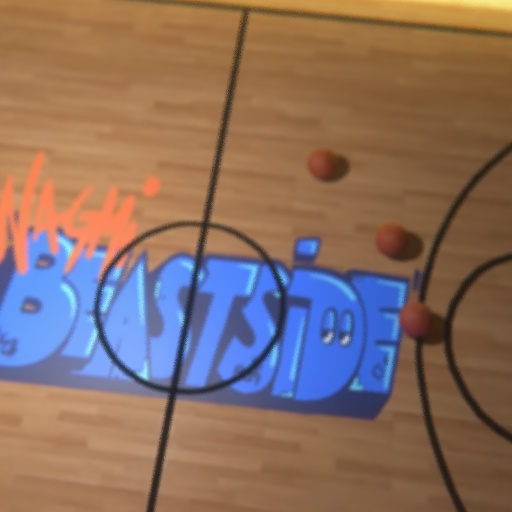} 
    } & 
    \parbox[c]{\QualitativeCompareBoxWidth}{
    \includegraphics[width=\QualitativeCompareImageWidth]{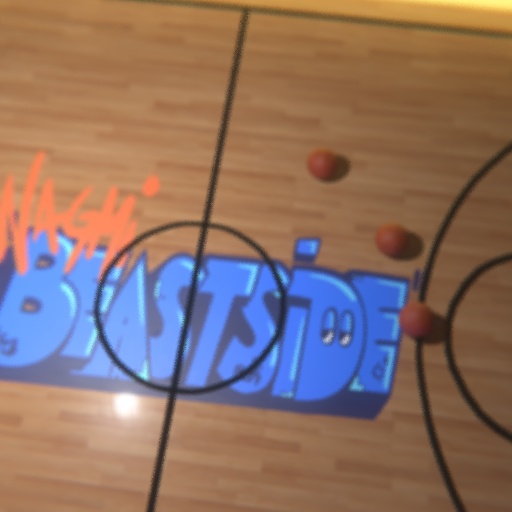} 
    } \\ 

    Fisheye &
    \parbox[c]{\QualitativeCompareBoxWidth}{
    \includegraphics[width=\QualitativeCompareImageWidth]{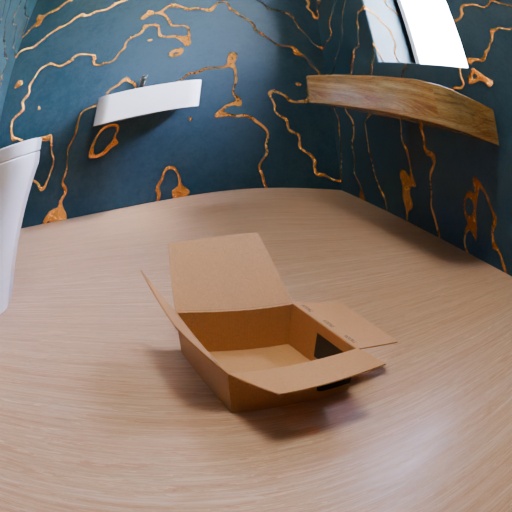} 
    } & 
    \parbox[c]{\QualitativeCompareBoxWidth}{
    \includegraphics[width=\QualitativeCompareImageWidth]{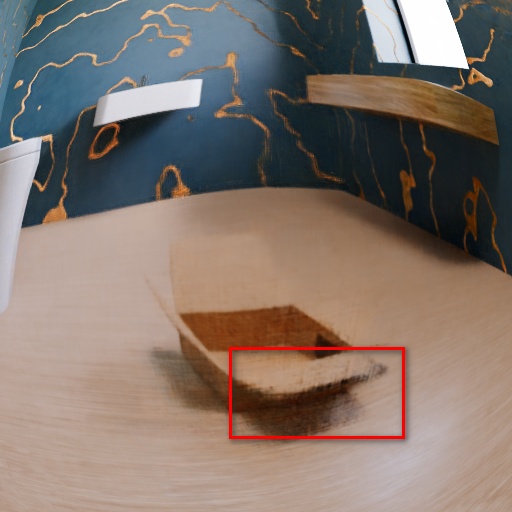}
    } & 
    \parbox[c]{\QualitativeCompareBoxWidth}{
    \includegraphics[width=\QualitativeCompareImageWidth]{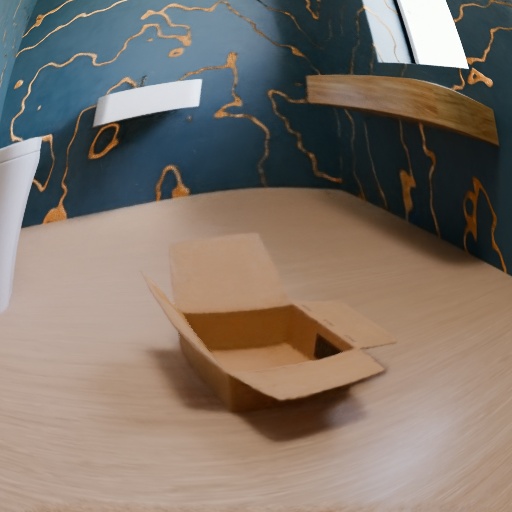} 
    } & 
    \parbox[c]{\QualitativeCompareBoxWidth}{
    \includegraphics[width=\QualitativeCompareImageWidth]{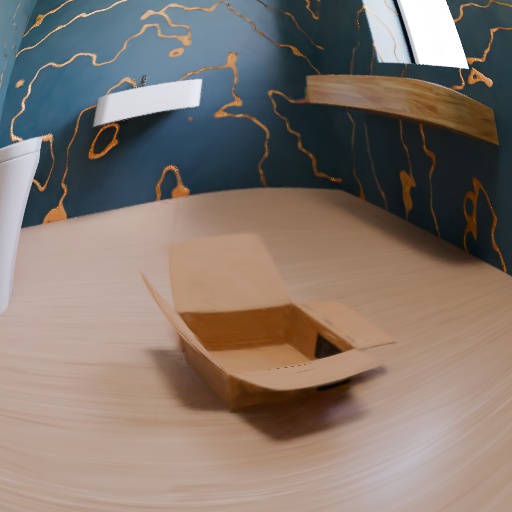} 
    } & 
    \parbox[c]{\QualitativeCompareBoxWidth}{
    \includegraphics[width=\QualitativeCompareImageWidth]{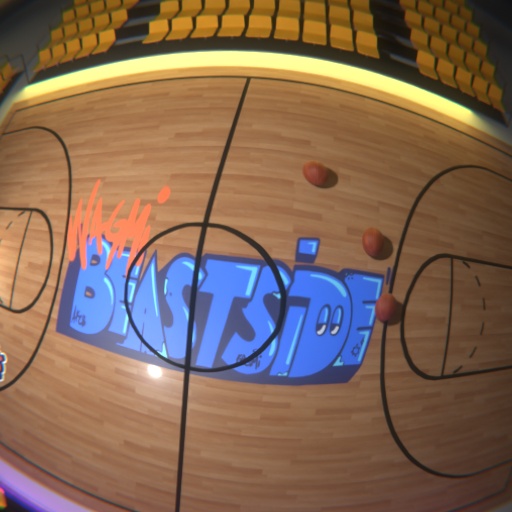} 
    } & 
    \parbox[c]{\QualitativeCompareBoxWidth}{
    \includegraphics[width=\QualitativeCompareImageWidth]{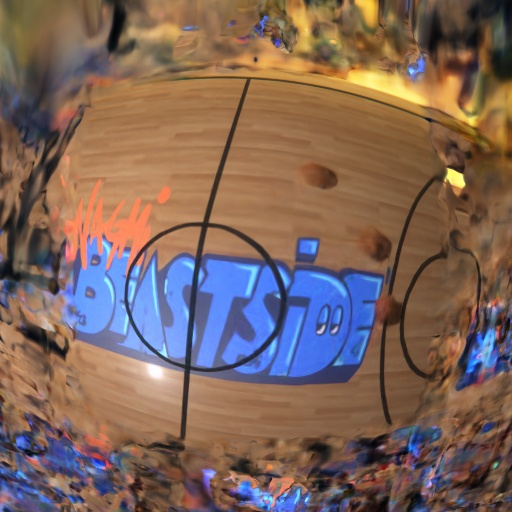} 
    } & 
    \parbox[c]{\QualitativeCompareBoxWidth}{
    \includegraphics[width=\QualitativeCompareImageWidth]{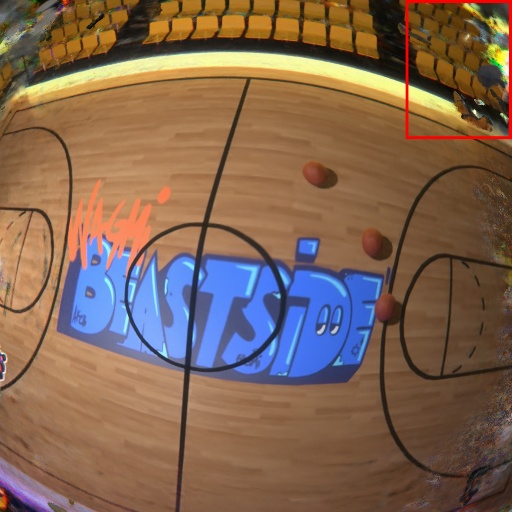} 
    } & 
    \parbox[c]{\QualitativeCompareBoxWidth}{
    \includegraphics[width=\QualitativeCompareImageWidth]{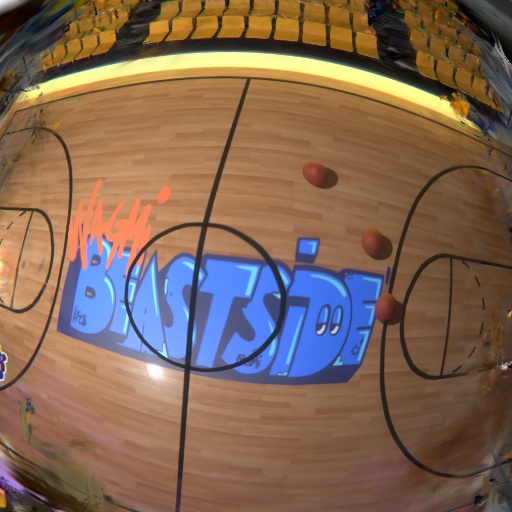} 
    } \\

    \end{tabular}
    }
    \caption{
    Qualitative comparison on CamVFX-4D. The artifacts are framed by red rectangles. \ifarxiv See the project page for the videos.\fi
    }
    \label{fig:qualitative_result}
    \vspace{-1em}
\end{figure*}

\noindent {\bf Datasets.}
Simultaneously capturing multi-view dynamic scenes under various camera settings in the real world is practically infeasible, and existing public datasets lack such configurations. To address this, we constructed \textbf{CamVFX-4D}, a synthetic multi-view dataset, using Blender 4.5~\cite{blender_software,plenoblendernerf} and BlenderKit~\cite{blenderkit} assets. CamVFX-4D contains 8 distinct dynamic indoor scenes featuring diverse material properties and physical interactions. For each scene, we rendered 50-frame, 512×512 videos from 50 viewpoints under four effects (pinhole, fisheye, rolling shutter, and depth of field).

\newcommand\QualitativeLabelWidth{.1\textwidth}
\newcommand\QualitativeBoxWidth{.19\textwidth}
\newcommand\QualitativeImageWidth{.19\textwidth}

\newcommand{\RowLabel}[1]{%
    \parbox[c]{\QualitativeLabelWidth}{\centering\scriptsize #1}%
}

\begin{figure*}[t]
    \centering
    \setlength{\tabcolsep}{1.5pt}
    \renewcommand{\arraystretch}{1}
    \resizebox{\textwidth}{!}{
    \begin{tabular}{@{}c@{}c@{}c@{}c@{}c@{}c@{}}

    \RowLabel{\shortstack{Pinhole}} &
    \parbox[c]{\QualitativeBoxWidth}{
    \includegraphics[width=\QualitativeImageWidth]{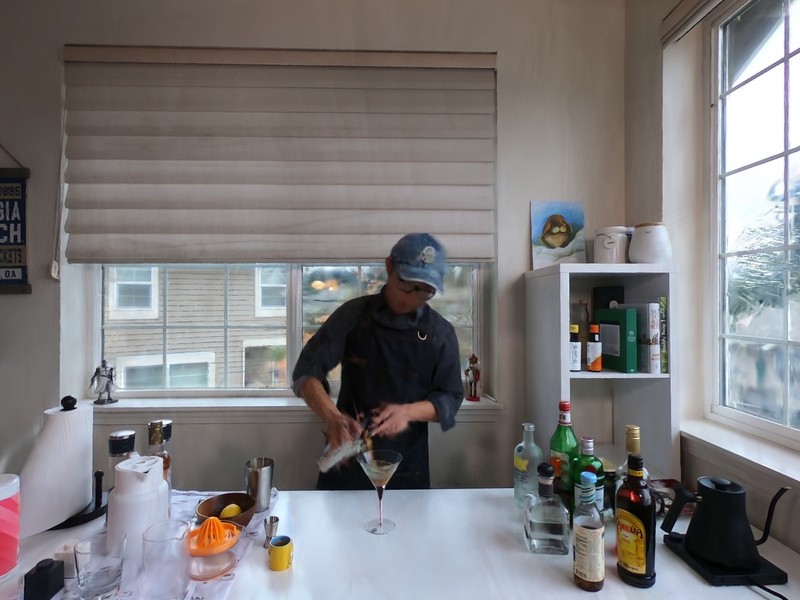}
    } &
    \parbox[c]{\QualitativeBoxWidth}{
    \includegraphics[width=\QualitativeImageWidth]{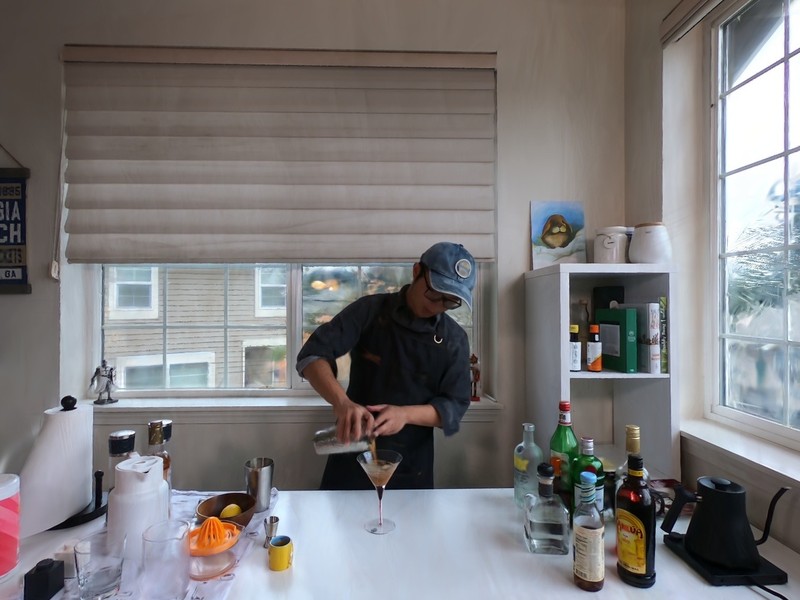}
    } &
    \parbox[c]{\QualitativeBoxWidth}{
    \includegraphics[width=\QualitativeImageWidth]{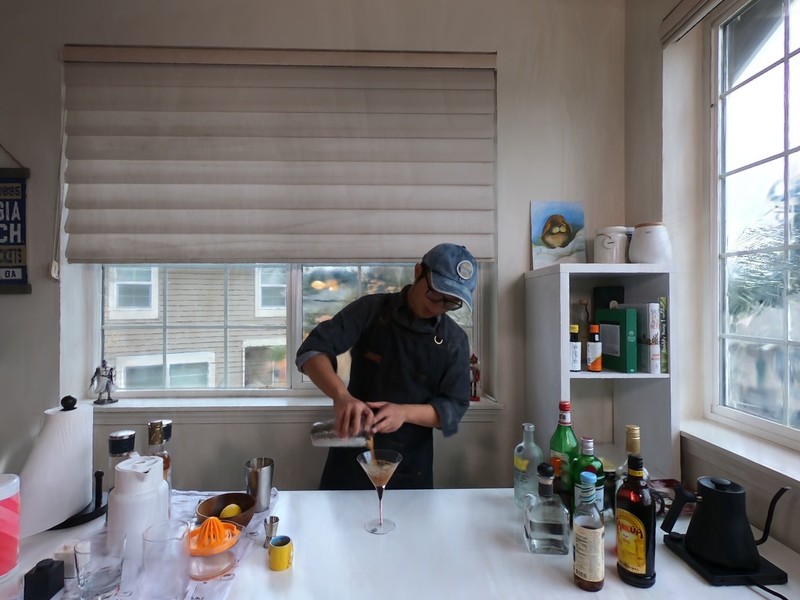}
    } &
    \parbox[c]{\QualitativeBoxWidth}{
    \includegraphics[width=\QualitativeImageWidth]{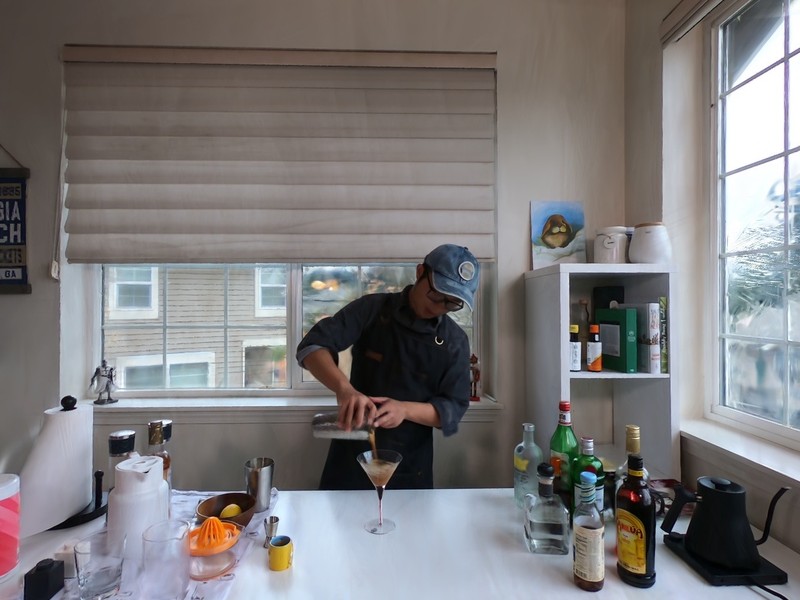}
    } &
    \parbox[c]{\QualitativeBoxWidth}{
    \includegraphics[width=\QualitativeImageWidth]{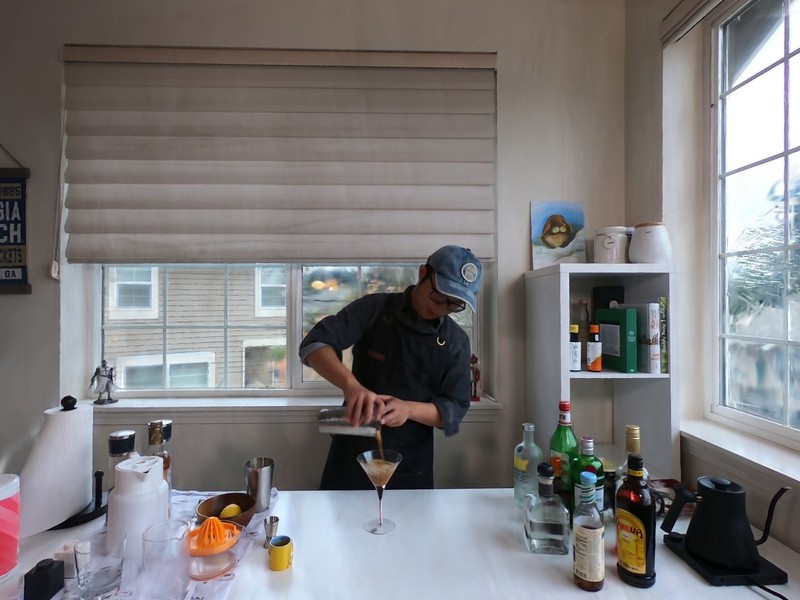}
    } \\

    \RowLabel{\shortstack{Depth\\of\\Field}} &
    \parbox[c]{\QualitativeBoxWidth}{
    \includegraphics[width=\QualitativeImageWidth]{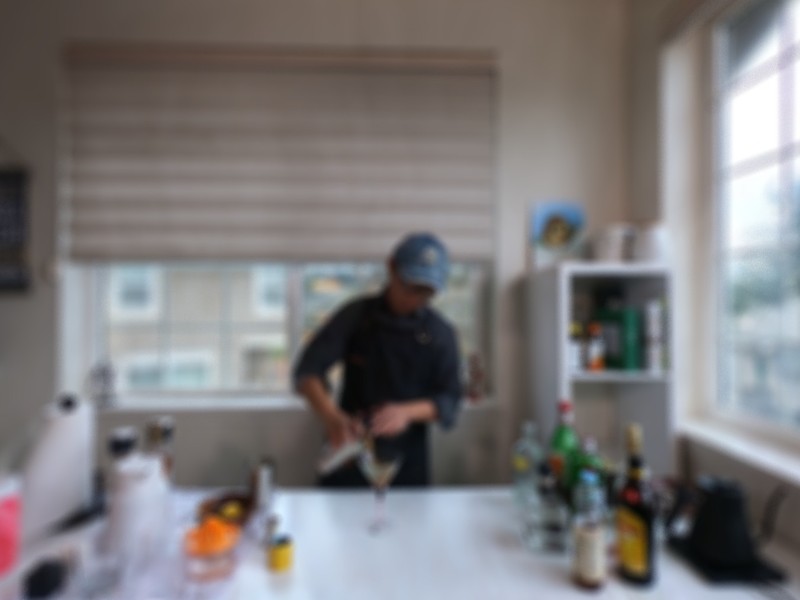}
    } &
    \parbox[c]{\QualitativeBoxWidth}{
    \includegraphics[width=\QualitativeImageWidth]{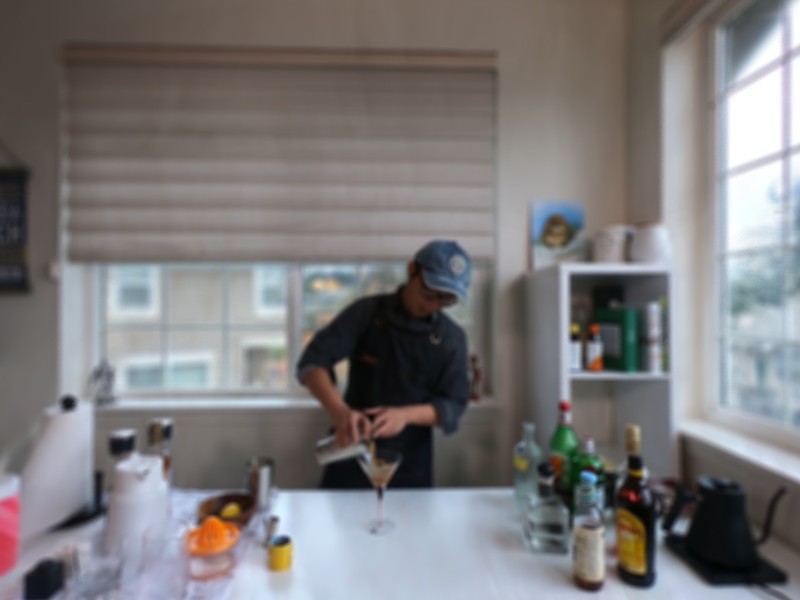}
    } &
    \parbox[c]{\QualitativeBoxWidth}{
    \includegraphics[width=\QualitativeImageWidth]{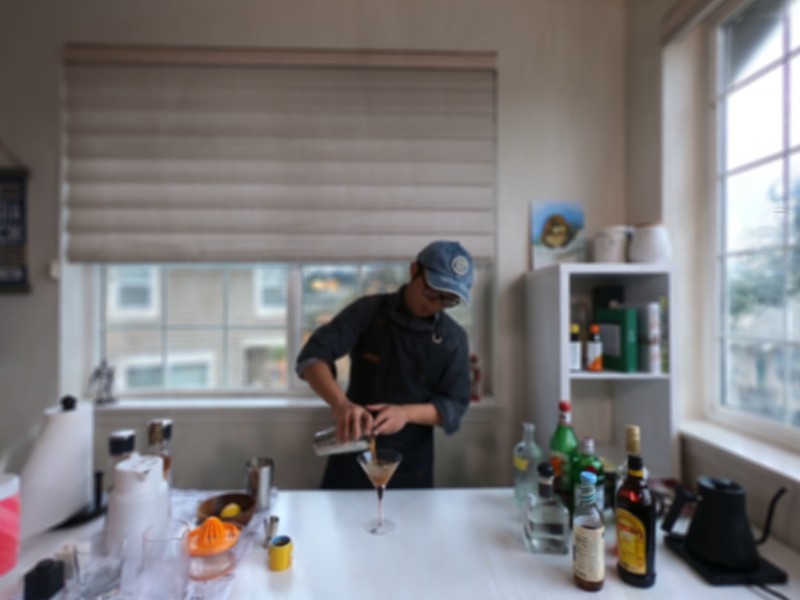}
    } &
    \parbox[c]{\QualitativeBoxWidth}{
    \includegraphics[width=\QualitativeImageWidth]{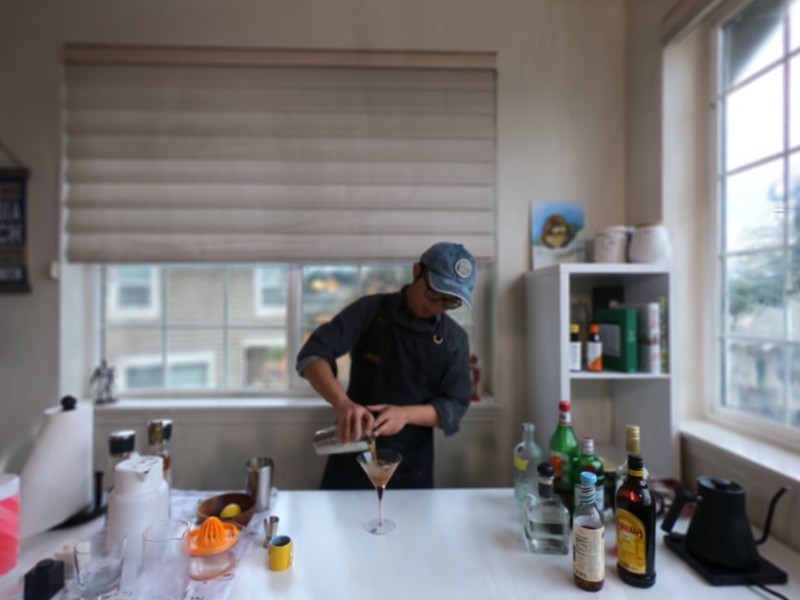}
    } &
    \parbox[c]{\QualitativeBoxWidth}{
    \includegraphics[width=\QualitativeImageWidth]{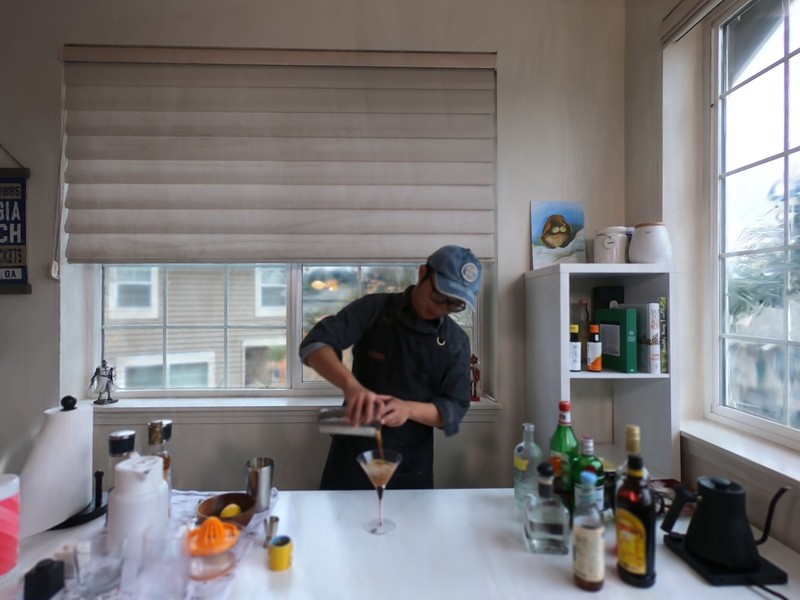}
    } \\

    \RowLabel{\shortstack{Rolling\\Shutter}} &
    \parbox[c]{\QualitativeBoxWidth}{
    \includegraphics[width=\QualitativeImageWidth]{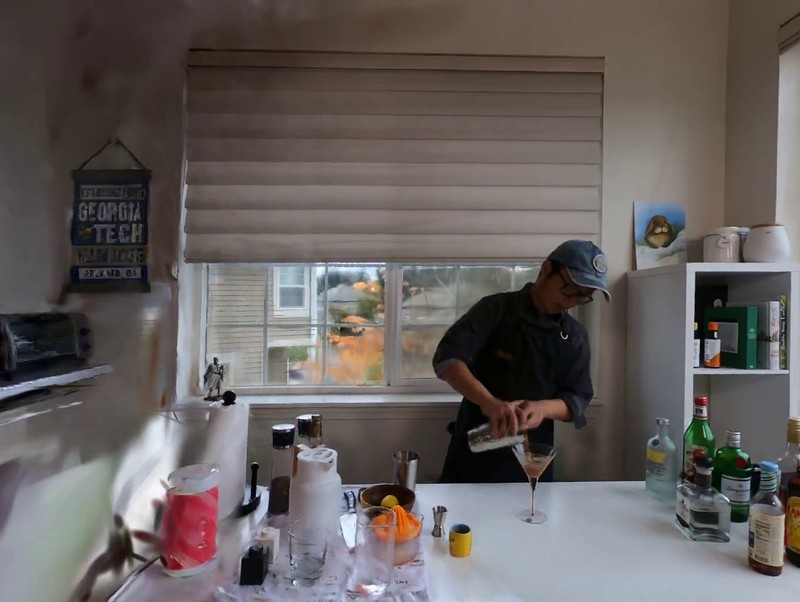}
    } &
    \parbox[c]{\QualitativeBoxWidth}{
    \includegraphics[width=\QualitativeImageWidth]{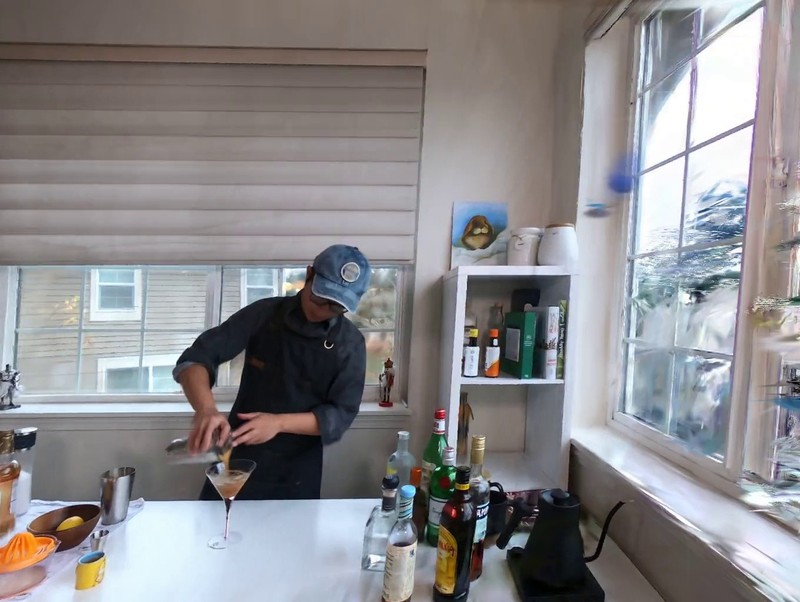}
    } &
    \parbox[c]{\QualitativeBoxWidth}{
    \includegraphics[width=\QualitativeImageWidth]{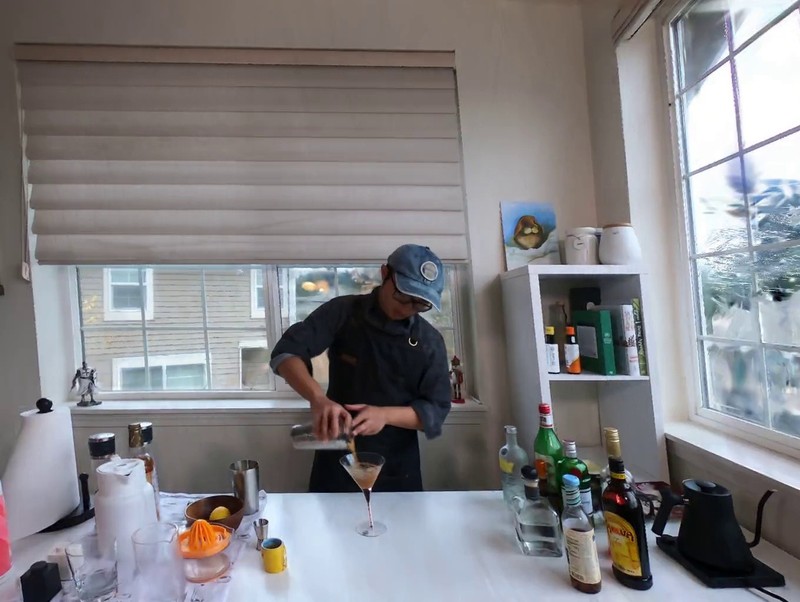}
    } &
    \parbox[c]{\QualitativeBoxWidth}{
    \includegraphics[width=\QualitativeImageWidth]{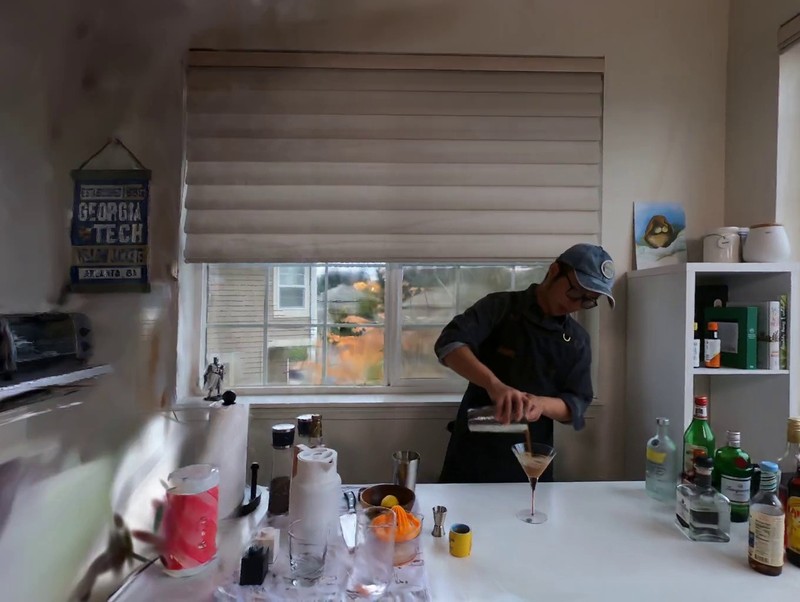}
    } &
    \parbox[c]{\QualitativeBoxWidth}{
    \includegraphics[width=\QualitativeImageWidth]{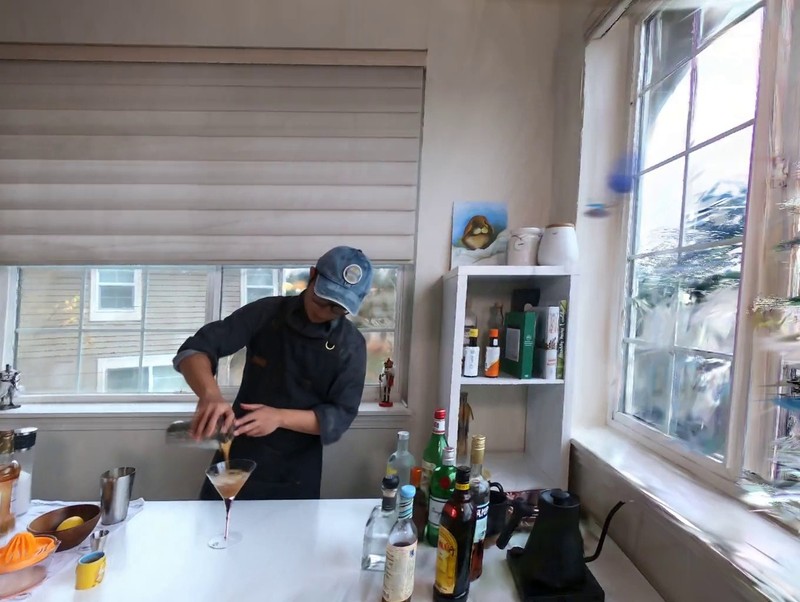}
    } \\

    \RowLabel{\shortstack{Fisheye}} &
    \parbox[c]{\QualitativeBoxWidth}{
    \includegraphics[width=\QualitativeImageWidth]{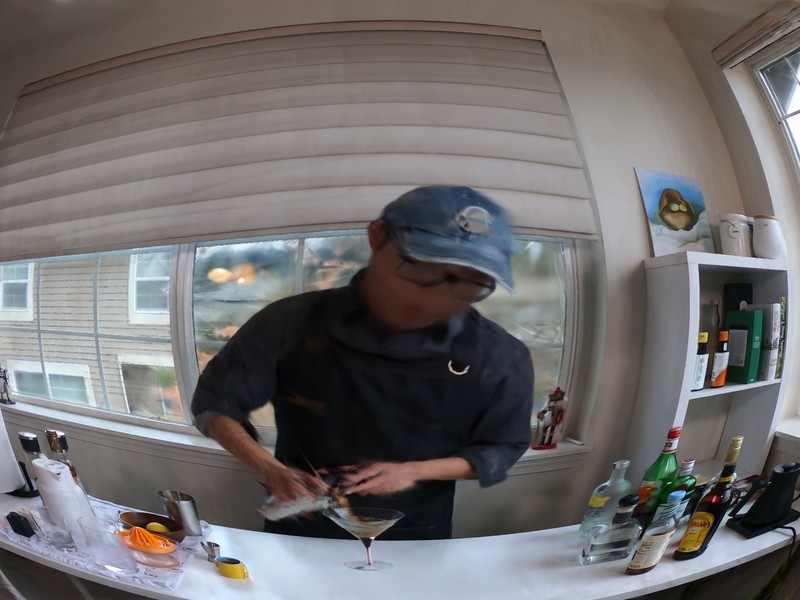}
    } &
    \parbox[c]{\QualitativeBoxWidth}{
    \includegraphics[width=\QualitativeImageWidth]{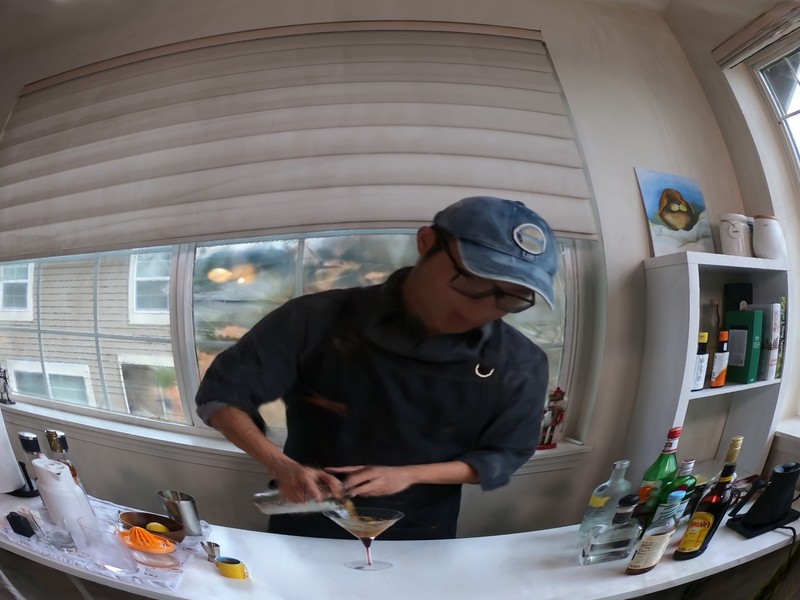}
    } &
    \parbox[c]{\QualitativeBoxWidth}{
    \includegraphics[width=\QualitativeImageWidth]{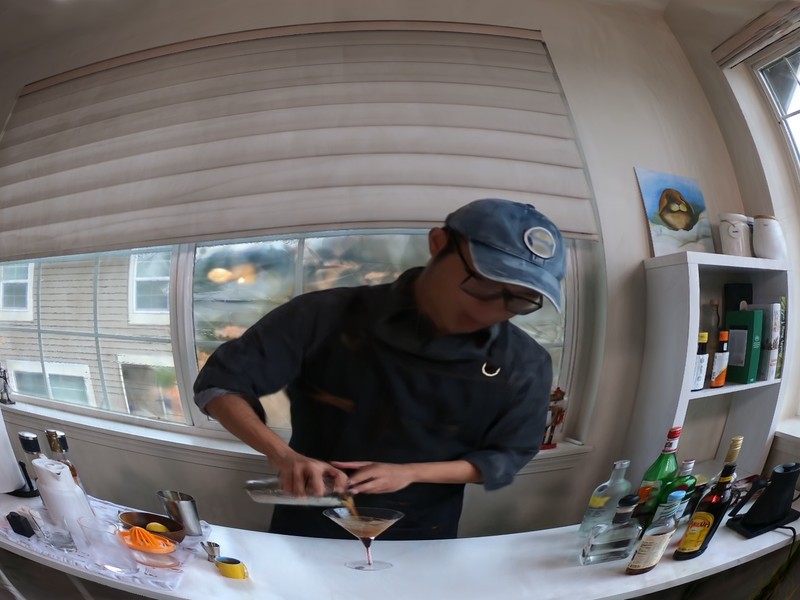}
    } &
    \parbox[c]{\QualitativeBoxWidth}{
    \includegraphics[width=\QualitativeImageWidth]{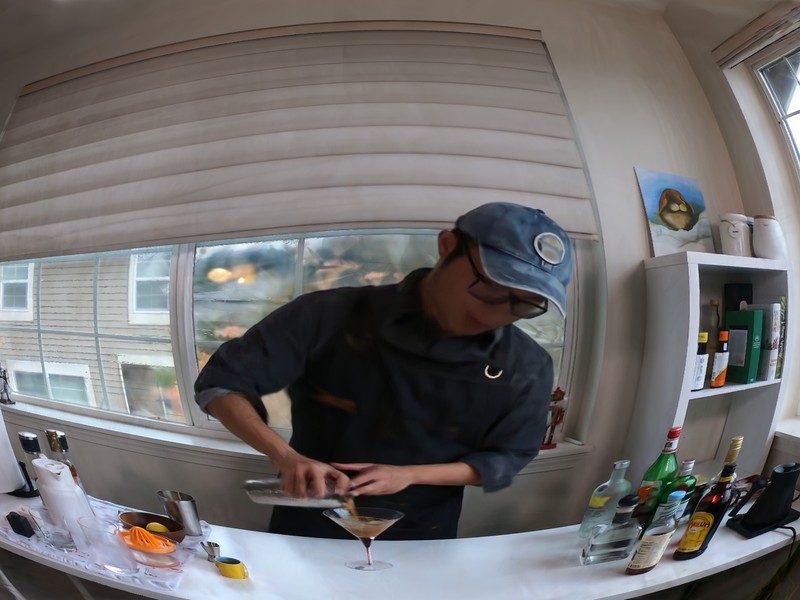}
    } &
    \parbox[c]{\QualitativeBoxWidth}{
    \includegraphics[width=\QualitativeImageWidth]{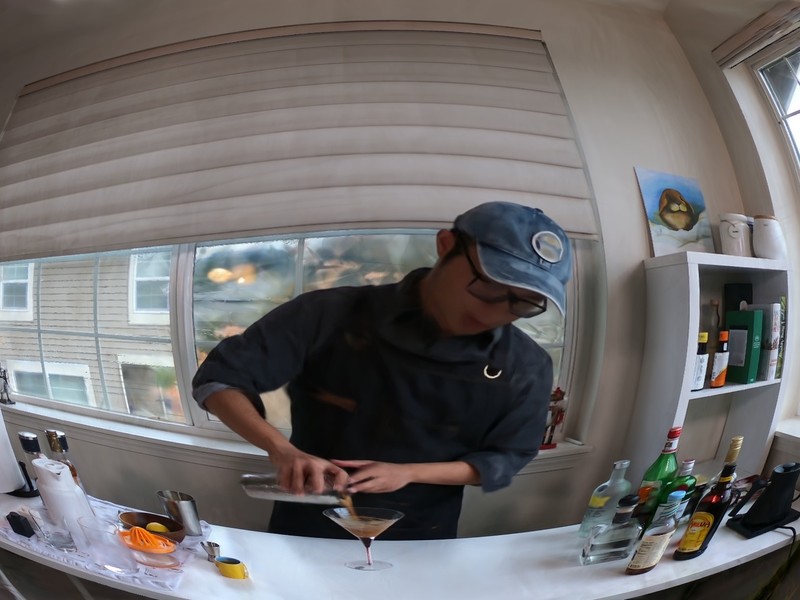}
    } \\

    \end{tabular}
    }
    \caption{
    Qualitative results of 4D-GRT on the Neural 3D Video dataset~\cite{li2022neural3dvideosynthesis}.
    Each row shows one camera effect, and each column progresses temporally from left to right. \ifarxiv See the project page for the videos.\fi 
    }
    \label{fig:qualitative_realworld}
    \vspace{-1em}
\end{figure*}

\begin{table*}[t]
  \centering
  \begin{minipage}{0.48\textwidth}
  \centering
  \caption{Quantitative comparison on pinhole camera rendering.}
  \label{tab:pinhole_quality_evaluation}
  \resizebox{\linewidth}{!}{
    \begin{tabular}{lccc|c}
      \toprule
      Methods & PSNR(dB)$\uparrow$ & SSIM$\uparrow$ & LPIPS$\downarrow$ & FPS$\uparrow$ \\
      \midrule
      HexPlane~\cite{cao2023hexplane} & 23.1124 & 0.7956 & 0.2942 & 0.20 \\
      MSTH~\cite{wang2023masked} & 29.4310 & \textbf{0.9023} & 0.1139 & 9.38 \\
      4D-GRT (Ours) & \textbf{32.8010} & 0.8898 & \textbf{0.1018} & \textbf{36.56} \\
      \bottomrule
    \end{tabular}
  }
  \vspace{-1em}
\end{minipage}%

  \hfill
  \begin{minipage}{0.48\textwidth}
  \centering
  \caption{Quantitative comparison on depth-of-field rendering.}
  \label{tab:dof_quality_evaluation}
  \resizebox{\linewidth}{!}{
    \begin{tabular}{lccc|c}
      \toprule
      Methods & PSNR(dB)$\uparrow$ & SSIM$\uparrow$ & LPIPS$\downarrow$ & FPS$\uparrow$ \\
      \midrule
      HexPlane~\cite{cao2023hexplane} & 18.3722 & 0.7343 & 0.5056 & 0.01 \\
      MSTH~\cite{wang2023masked} & 28.4692 & 0.9009 & 0.1540 & 0.57 \\
      4D-GRT (Ours) & \textbf{31.2475} & \textbf{0.9124} & \textbf{0.1210} & \textbf{3.44} \\
      \bottomrule
    \end{tabular}
  }
  \vspace{-1em}
\end{minipage}%

\end{table*}

\begin{table}[t]
  \caption{Quantitative comparison of fisheye rendering. The lower block reports metrics computed within the pre-defined mask.}
  \label{tab:fisheye_quality_evaluation}
  \centering
  \small
  \setlength{\tabcolsep}{2pt}
  \begin{tabular*}{\columnwidth}{@{\extracolsep{\fill}}lcccc@{}}
    \toprule
    Methods & PSNR(dB)$\uparrow$ & SSIM$\uparrow$ & LPIPS$\downarrow$ & FPS$\uparrow$ \\
    \midrule
    \multicolumn{5}{@{}l}{\emph{Full image}} \\
    HexPlane~\cite{cao2023hexplane} & 15.6489 & 0.6678 & 0.4142 & 0.21 \\
    MSTH~\cite{wang2023masked} & \textbf{24.4222} & \textbf{0.8163} & 0.1764 & 9.38 \\
    4D-GRT (Ours) & 24.1322 & 0.8162 & \textbf{0.1678} & \textbf{41.53} \\
    \midrule
    \multicolumn{5}{@{}l}{\emph{Within mask}} \\
    HexPlane~\cite{cao2023hexplane} & 21.6869 & 0.7480 & 0.2726 & -- \\
    MSTH~\cite{wang2023masked} & 26.7930 & \textbf{0.8571} & \textbf{0.1153} & -- \\
    4D-GRT (Ours) & \textbf{28.8927} & 0.8555 & 0.1259 & -- \\
    \bottomrule
  \end{tabular*}
  \vspace{-1em}
\end{table}

\noindent {\bf Evaluation Setup.}
Since generating camera-effect videos from dynamic scenes is a novel task, we compare against two state-of-the-art Dynamic NeRF methods, HexPlane~\cite{cao2023hexplane} and MSTH~\cite{wang2023masked}, which support 4D scene reconstruction and ray-based rendering. For a fair comparison, all methods are trained on the same multi-view pinhole videos until convergence, using all camera views and time frames. At evaluation time, we render the same viewpoints under different camera-effect settings and compare against the corresponding ground-truth videos. To isolate reconstruction quality from camera-effect implementation, we apply the same camera-effect rendering module to all baselines. We report PSNR~\cite{huynh2008scope}, SSIM~\cite{wang2004image}, and LPIPS~\cite{zhang2018unreasonable}, averaged over all frames, views, and scenes, as well as rendering speed (FPS). For rolling shutter, we additionally evaluate our method with chunk sizes $N_c = 1$, $4$, and $16$.
Existing effect-specific Gaussian methods, such as DoF-GS~\cite{wang2025dofgsadjustabledepthoffield3dgaussian} and DoF-Gaussian~\cite{shen2025dofgaussiancontrollabledepthoffield3d}, are not directly comparable because they assume different inputs and target effect-specific outputs rather than general camera-effect synthesis from multi-view pinhole videos.

\noindent {\bf Quantitative comparison.} 
Tables~\ref{tab:pinhole_quality_evaluation}--\ref{tab:rollingshutter_quality_evaluation} present the quantitative results. 4D-GRT achieves the highest rendering speed across all camera effects due to hardware-accelerated ray tracing, while its continuous 4D-GS representation yields superior or competitive visual quality. For pinhole, DoF, and rolling-shutter effects, 4D-GRT achieves higher PSNR than both baselines with SSIM and LPIPS comparable to MSTH.
For fisheye rendering, the extreme field-of-view captures regions unobserved during pinhole training, leading to unreliable standard metrics. To address this, we apply a circular mask (diameter 409.6) to isolate regions visible to the training cameras. Under this masked evaluation (Table~\ref{tab:fisheye_quality_evaluation}), our method substantially outperforms the baselines in PSNR while matching them in SSIM and LPIPS.
For rolling-shutter rendering, row-by-row evaluation ($N_c = 1$) is inherently slower than MSTH~\cite{wang2023masked}. Because 4D-GRT requires generating a separate deformed scene per row, it cannot leverage the direct spatio-temporal ray queries used by grid-based baselines. However, our chunking approximation ($N_c = 16$) overcomes this architectural difference, achieving higher FPS than both baselines with negligible quality degradation (Table~\ref{tab:rollingshutter_quality_evaluation}).

\begin{table}[!t]
  \centering
  \caption{Quantitative comparison on rolling-shutter rendering.}
  \label{tab:rollingshutter_quality_evaluation}
  \resizebox{\columnwidth}{!}{
    \begin{tabular}{lcccc|c}
      \toprule
      Methods & Chunk & PSNR(dB)$\uparrow$ & SSIM$\uparrow$ & LPIPS$\downarrow$ & FPS$\uparrow$ \\
      \midrule
      HexPlane~\cite{cao2023hexplane} & N/A &  21.3468 & 0.7723 & 0.3162 & 0.21 \\
      MSTH~\cite{wang2023masked} & N/A & 28.7003 & \textbf{0.8927} & 0.1190 & 9.35 \\
      4D-GRT (Ours) & 1 row &31.6144 & 0.8821 & \textbf{0.1056} & 0.76 \\
      4D-GRT (Ours) & 4 rows & \textbf{31.6146} & 0.8821 & \textbf{0.1056} & 4.99 \\
      4D-GRT (Ours) & 16 rows & 31.6082 & 0.8820 & \textbf{0.1056} & \textbf{13.54} \\
      \bottomrule
    \end{tabular}
  }
  \vspace{-1em}
\end{table}

\noindent {\bf Qualitative comparison.}
We present the rendering results of 4D-GRT (ours), MSTH~\cite{wang2023masked}, and HexPlane~\cite{cao2023hexplane} in Fig.~\ref{fig:qualitative_result}. Compared with the baselines, 4D-GRT preserves finer details and more consistent color. HexPlane exhibits stronger reconstruction artifacts, while MSTH shows residual artifacts and color-temperature shifts.

\newlength{\CompareLabelWidth}
\newlength{\CompareLabelImageGap}
\newlength{\CompareFrameGap}
\newlength{\CompareImageWidth}

\setlength{\CompareLabelWidth}{.075\textwidth}
\setlength{\CompareLabelImageGap}{0.4em}
\setlength{\CompareFrameGap}{0.5em}

\newcommand{\CompareRowLabel}[1]{%
    \scriptsize #1%
}

\setlength{\CompareImageWidth}{%
    (\textwidth-\CompareLabelWidth-\CompareLabelImageGap-5\CompareFrameGap)/6%
}

\newcommand{\ImageCell}[1]{%
    \includegraphics[width=\CompareImageWidth]{#1}%
}

\newcommand{\CameraTitle}[1]{%
    \parbox[c][1.6em][c]{3\CompareImageWidth+2\CompareFrameGap}{%
        \centering\textbf{#1}%
    }%
}

\begin{figure*}[t]
    \centering
    \small
    \setlength{\tabcolsep}{0pt}
    \renewcommand{\arraystretch}{1.0}

    \begin{tabular}{@{}
        >{\centering\arraybackslash}m{\CompareLabelWidth}
        m{\CompareLabelImageGap}
        >{\centering\arraybackslash}m{\CompareImageWidth}
        m{\CompareFrameGap}
        >{\centering\arraybackslash}m{\CompareImageWidth}
        m{\CompareFrameGap}
        >{\centering\arraybackslash}m{\CompareImageWidth}
        m{\CompareFrameGap}
        >{\centering\arraybackslash}m{\CompareImageWidth}
        m{\CompareFrameGap}
        >{\centering\arraybackslash}m{\CompareImageWidth}
        m{\CompareFrameGap}
        >{\centering\arraybackslash}m{\CompareImageWidth}
    @{}}

    &
    &
    \multicolumn{5}{c}{\CameraTitle{Fisheye}} &
    &
    \multicolumn{5}{c}{\CameraTitle{Depth of Field}} \\[-1pt]

    \CompareRowLabel{SVD~\cite{blattmann2023stable}} &
    &
    \ImageCell{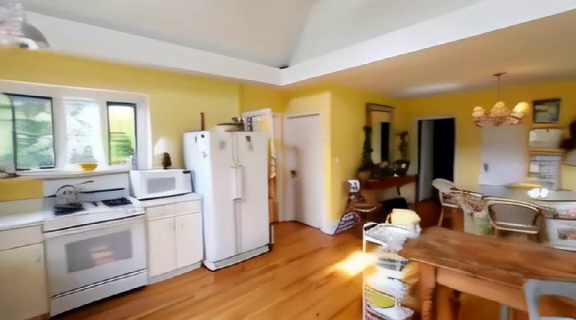} &
    &
    \ImageCell{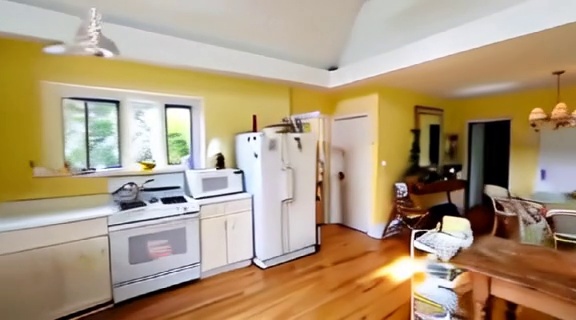} &
    &
    \ImageCell{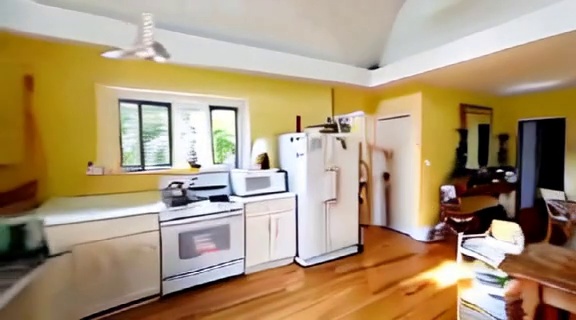} &
    &
    \ImageCell{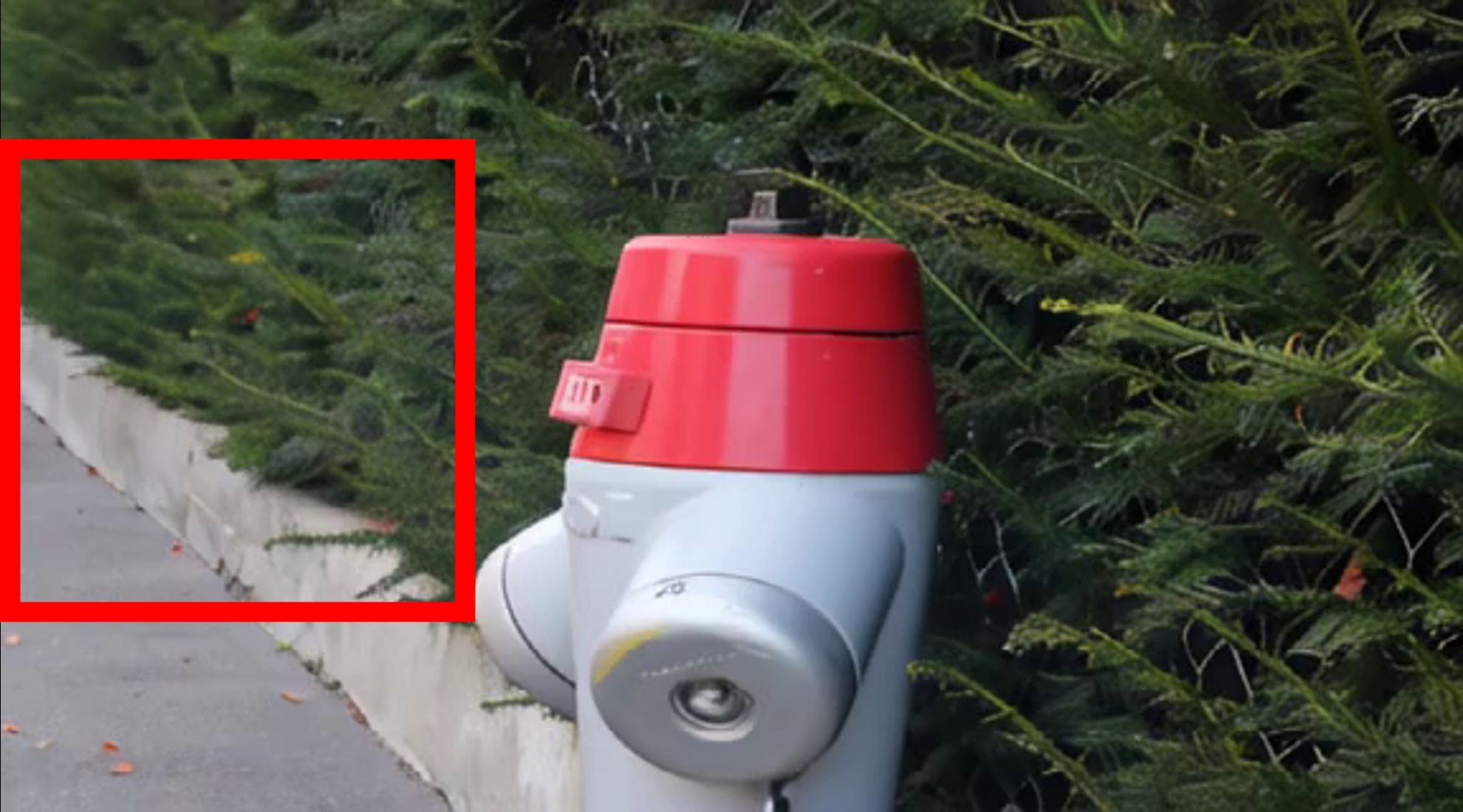} &
    &
    \ImageCell{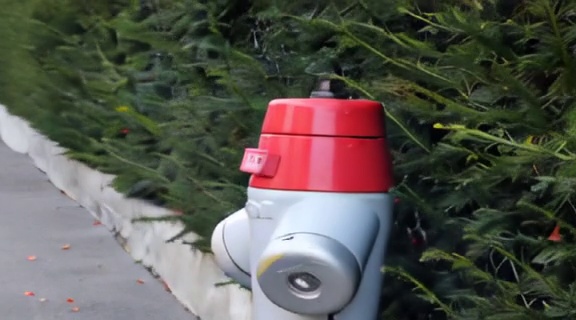} &
    &
    \ImageCell{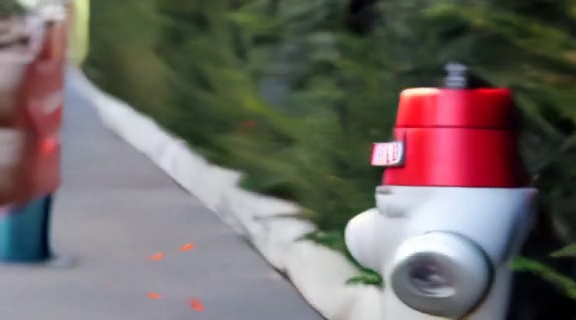} \\

    \CompareRowLabel{SVD~\cite{blattmann2023stable} (tuned)} &
    &
    \ImageCell{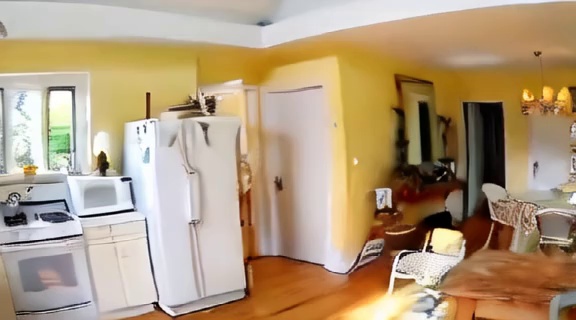} &
    &
    \ImageCell{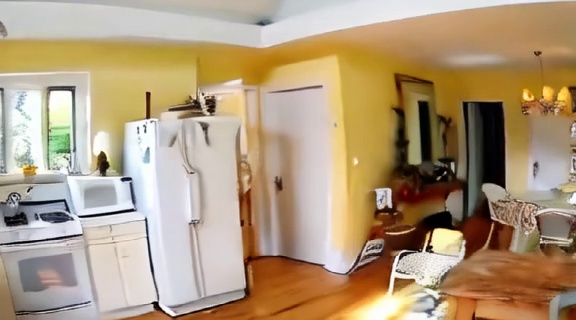} &
    &
    \ImageCell{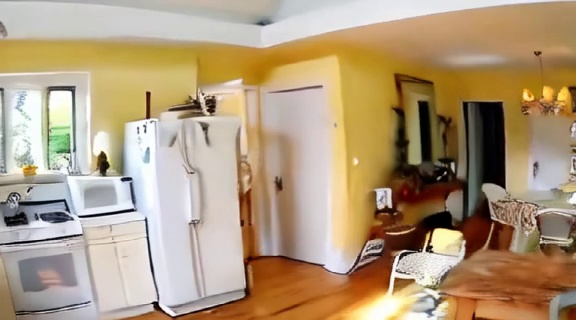} &
    &
    \ImageCell{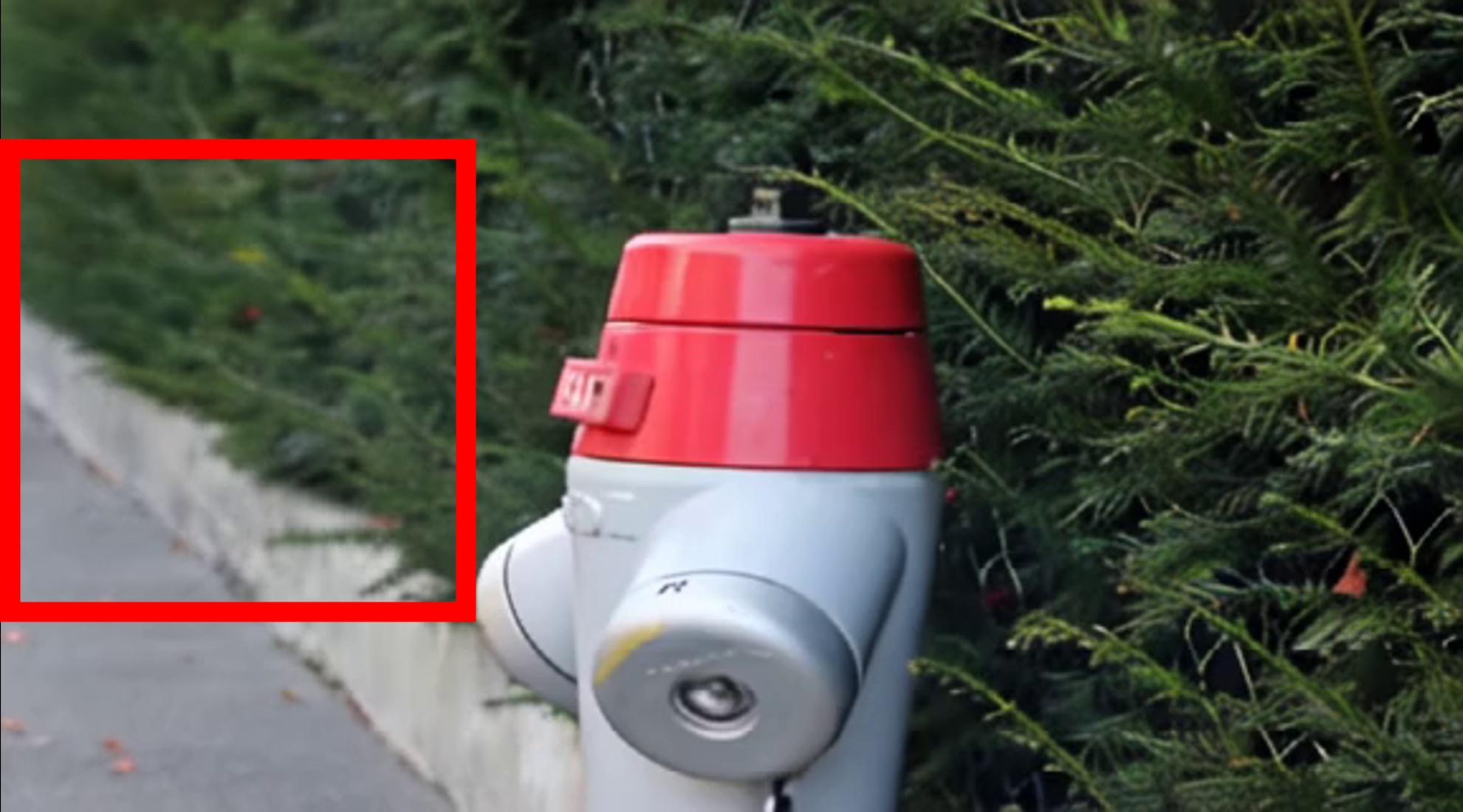} &
    &
    \ImageCell{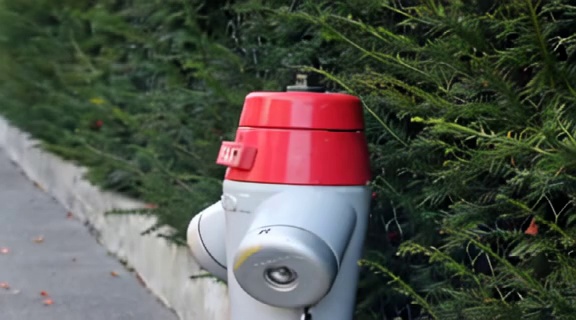} &
    &
    \ImageCell{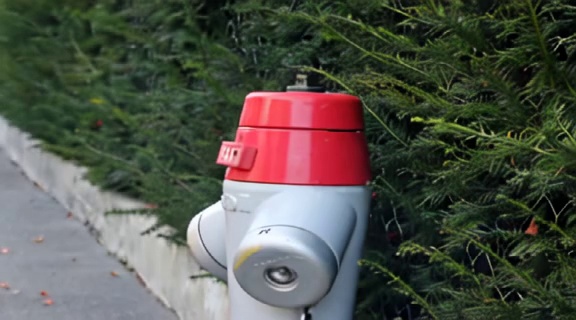} \\

    \end{tabular}

    \vspace{\CompareFrameGap}

    \begin{tabular}{@{}
        >{\centering\arraybackslash}m{\CompareLabelWidth}
        m{\CompareLabelImageGap}
        >{\centering\arraybackslash}m{\CompareImageWidth}
        m{\CompareFrameGap}
        >{\centering\arraybackslash}m{\CompareImageWidth}
        m{\CompareFrameGap}
        >{\centering\arraybackslash}m{\CompareImageWidth}
        m{\CompareFrameGap}
        >{\centering\arraybackslash}m{\CompareImageWidth}
        m{\CompareFrameGap}
        >{\centering\arraybackslash}m{\CompareImageWidth}
        m{\CompareFrameGap}
        >{\centering\arraybackslash}m{\CompareImageWidth}
    @{}}

    &
    &
    \multicolumn{5}{c}{\CameraTitle{Zoom}} &
    &
    \multicolumn{5}{c}{\CameraTitle{Dolly Zoom}} \\[-1pt]

    \CompareRowLabel{SVD~\cite{blattmann2023stable}} &
    &
    \ImageCell{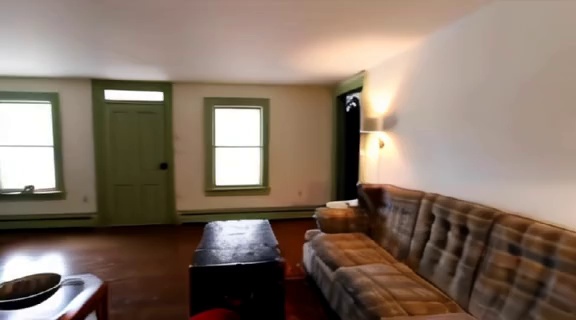} &
    &
    \ImageCell{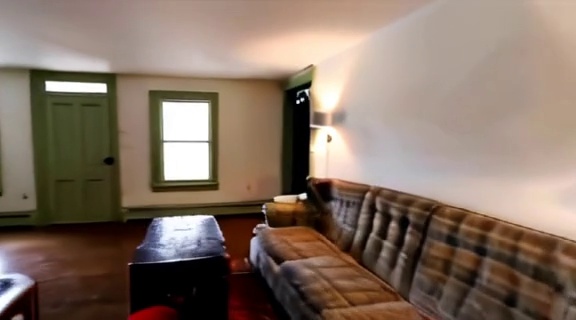} &
    &
    \ImageCell{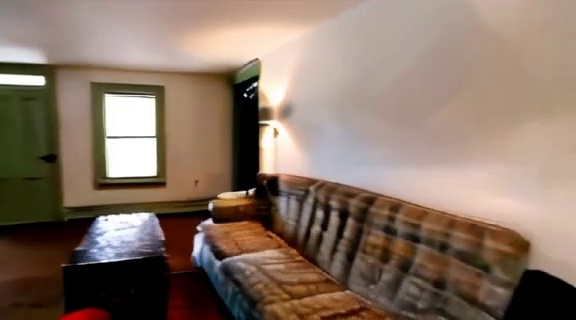} &
    &
    \ImageCell{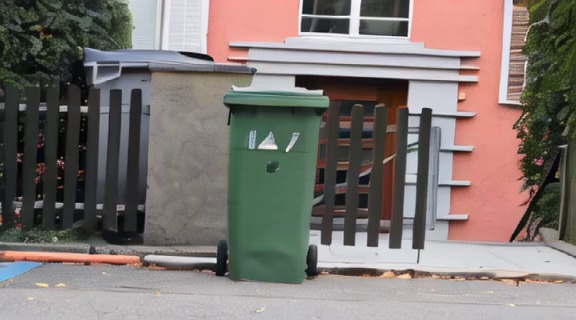} &
    &
    \ImageCell{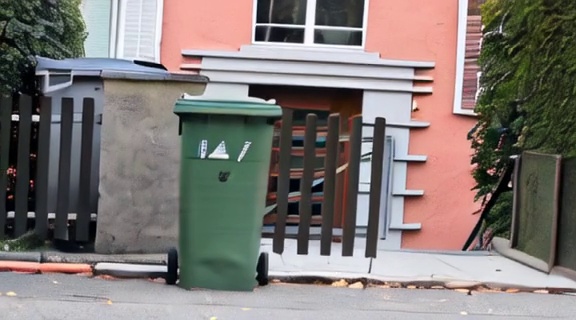} &
    &
    \ImageCell{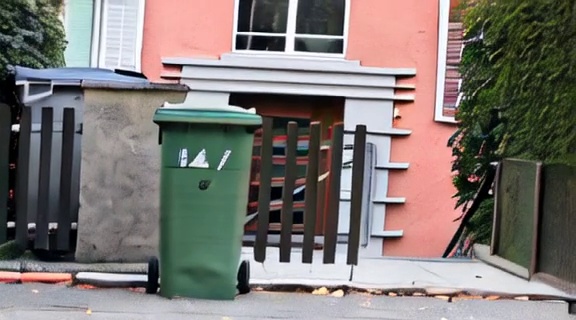} \\

    \CompareRowLabel{SVD~\cite{blattmann2023stable} (tuned)} &
    &
    \ImageCell{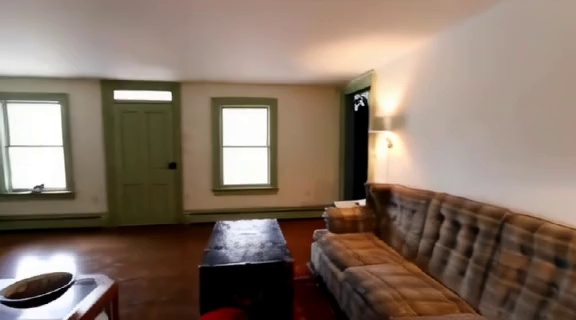} &
    &
    \ImageCell{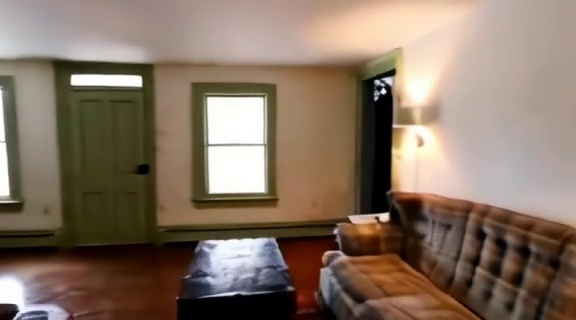} &
    &
    \ImageCell{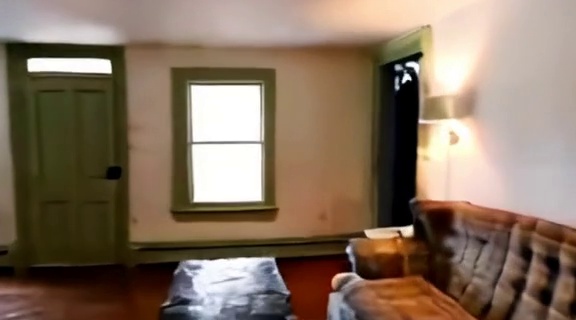} &
    &
    \ImageCell{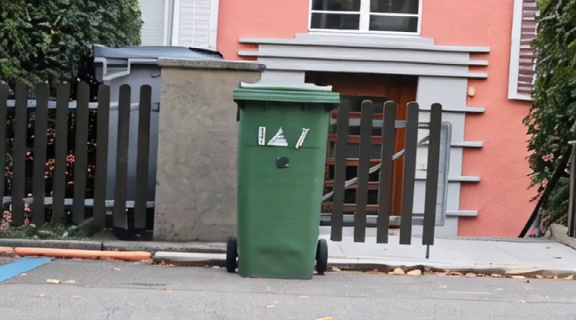} &
    &
    \ImageCell{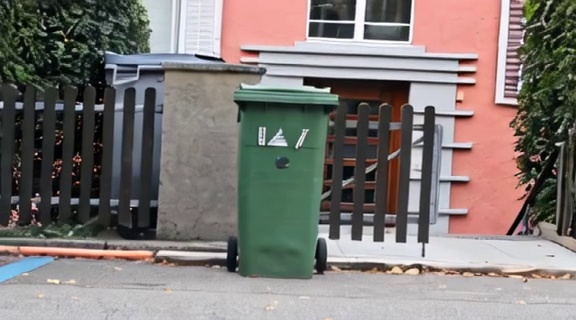} &
    &
    \ImageCell{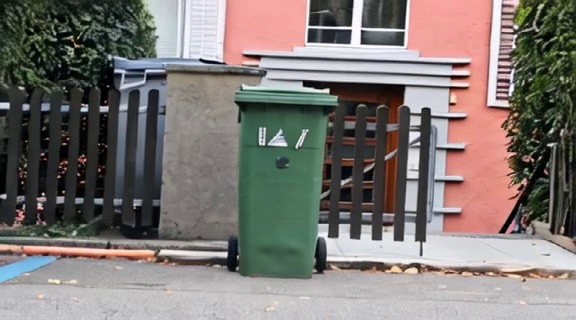} \\

    \end{tabular}

    \caption{
        Qualitative comparison of the pretrained SVD~\cite{blattmann2023stable} model versus our fine-tuned model. For each effect, the first, middle, and last frames of a generated sequence are shown from left to right. Red boxes highlight background areas that should be blurry because of the depth-of-field effect. Test images are sourced from the RealEstate10k~\cite{zhou2018stereo} and EBB!~\cite{ignatov2020rendering} datasets. \ifarxiv See the project page for the videos.\fi
    }
    \vspace{-10pt}
    \label{fig:svd_finetune}
\end{figure*}

\newcommand\MoVieSRealLabelWidth{.13\columnwidth}
\newcommand\MoVieSRealBoxWidth{.20\columnwidth}
\newcommand\MoVieSRealImageWidth{.20\columnwidth}
\newcommand{\MoVieSRealRowLabel}[1]{%
    \parbox[c]{\MoVieSRealLabelWidth}{\centering\scriptsize #1}%
}

\begin{figure}[t]
    \centering
    \setlength{\tabcolsep}{1.5pt}
    \renewcommand{\arraystretch}{0.95}

    \begin{tabular}{@{}c@{}c@{}c@{}c@{}c@{}}

    \MoVieSRealRowLabel{Pinhole} &
    \parbox[c]{\MoVieSRealBoxWidth}{
    \includegraphics[width=\MoVieSRealImageWidth]{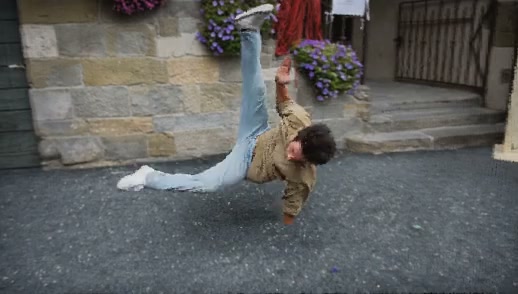}
    } &
    \parbox[c]{\MoVieSRealBoxWidth}{
    \includegraphics[width=\MoVieSRealImageWidth]{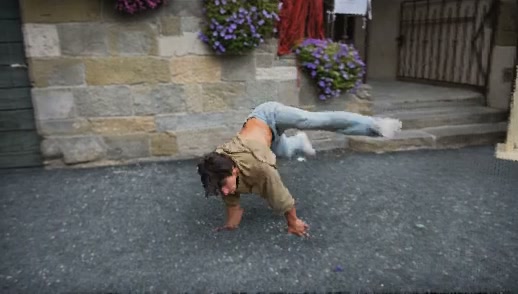}
    } &
    \parbox[c]{\MoVieSRealBoxWidth}{
    \includegraphics[width=\MoVieSRealImageWidth]{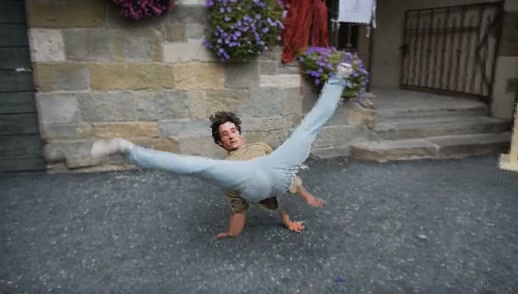}
    } &
    \parbox[c]{\MoVieSRealBoxWidth}{
    \includegraphics[width=\MoVieSRealImageWidth]{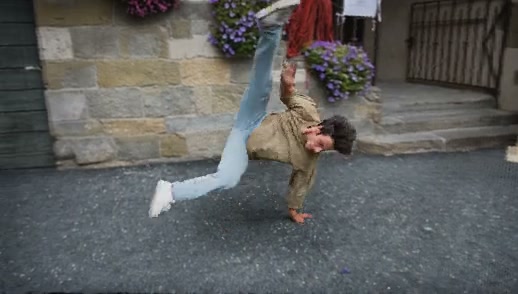}
    } \\

    \MoVieSRealRowLabel{\shortstack{Depth\\of\\Field}} &
    \parbox[c]{\MoVieSRealBoxWidth}{
    \includegraphics[width=\MoVieSRealImageWidth]{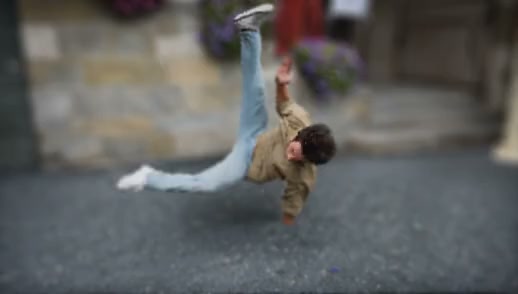}
    } &
    \parbox[c]{\MoVieSRealBoxWidth}{
    \includegraphics[width=\MoVieSRealImageWidth]{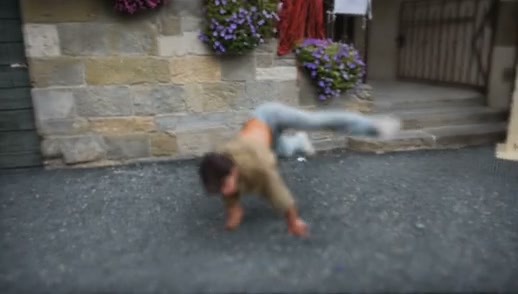}
    } &
    \parbox[c]{\MoVieSRealBoxWidth}{
    \includegraphics[width=\MoVieSRealImageWidth]{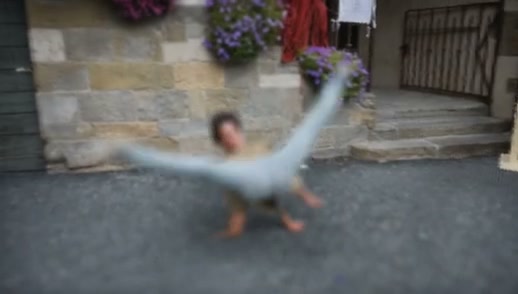}
    } &
    \parbox[c]{\MoVieSRealBoxWidth}{
    \includegraphics[width=\MoVieSRealImageWidth]{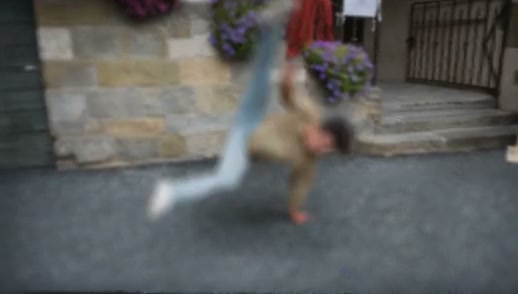}
    } \\

    \MoVieSRealRowLabel{\shortstack{Rolling\\Shutter}} &
    \parbox[c]{\MoVieSRealBoxWidth}{
    \includegraphics[width=\MoVieSRealImageWidth]{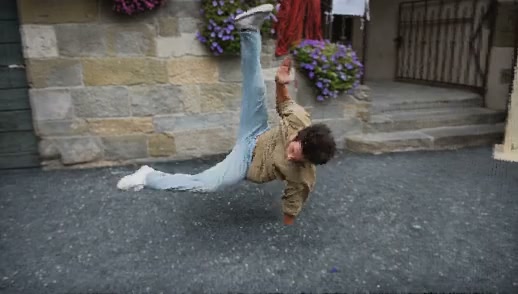}
    } &
    \parbox[c]{\MoVieSRealBoxWidth}{
    \includegraphics[width=\MoVieSRealImageWidth]{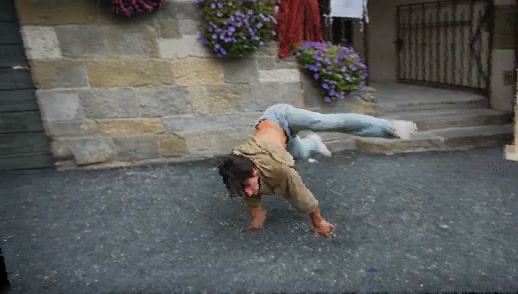}
    } &
    \parbox[c]{\MoVieSRealBoxWidth}{
    \includegraphics[width=\MoVieSRealImageWidth]{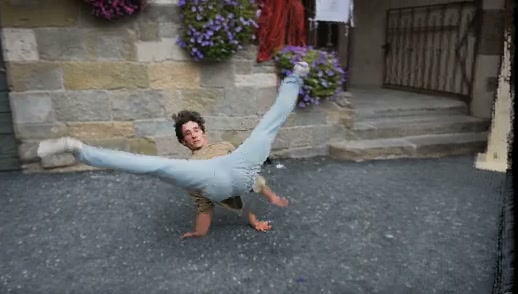}
    } &
    \parbox[c]{\MoVieSRealBoxWidth}{
    \includegraphics[width=\MoVieSRealImageWidth]{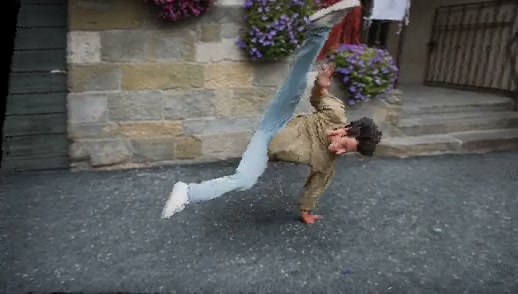}
    } \\

    \MoVieSRealRowLabel{Fisheye} &
    \parbox[c]{\MoVieSRealBoxWidth}{
    \includegraphics[width=\MoVieSRealImageWidth]{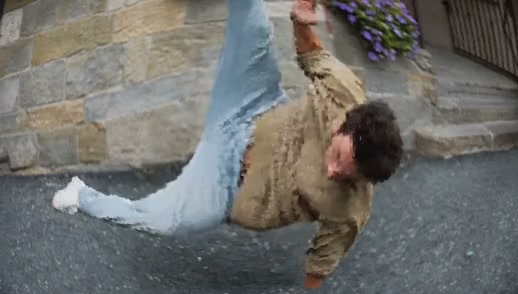}
    } &
    \parbox[c]{\MoVieSRealBoxWidth}{
    \includegraphics[width=\MoVieSRealImageWidth]{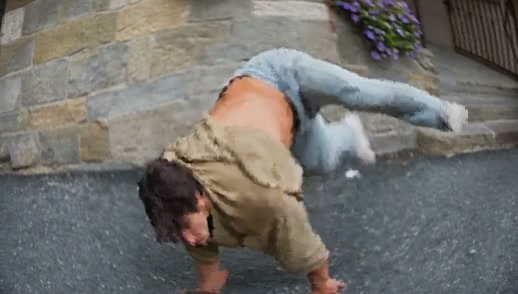}
    } &
    \parbox[c]{\MoVieSRealBoxWidth}{
    \includegraphics[width=\MoVieSRealImageWidth]{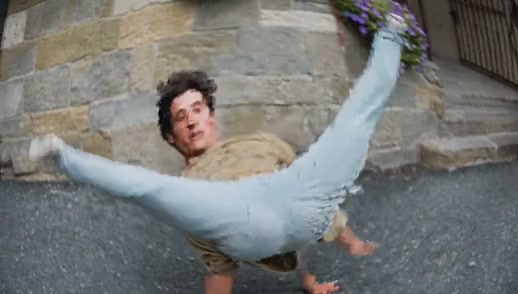}
    } &
    \parbox[c]{\MoVieSRealBoxWidth}{
    \includegraphics[width=\MoVieSRealImageWidth]{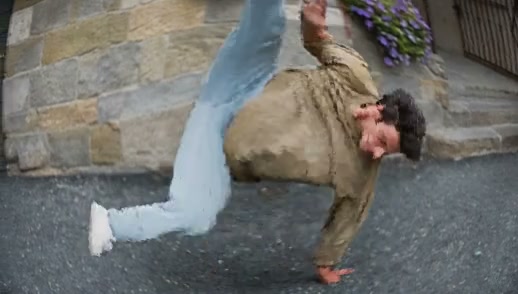}
    } \\

    \end{tabular}

    \caption{Qualitative results of applying our camera-effect renderer to a MoVieS~\cite{lin2026movies} reconstruction of a real-world monocular video from the DAVIS dataset~\cite{Perazzi_CVPR_2016}. Each row shows one camera model or camera effect. \ifarxiv See the project page for the videos.\fi}
    \label{fig:supp_movies_realworld}
    \vspace{-5mm}
\end{figure}

\noindent {\bf Qualitative results on a real-world dataset.}
We further evaluate 4D-GRT on real-world dynamic scenes by reconstructing and rendering sequences from the Neural 3D Video (Plenoptic Video) dataset~\cite{li2022neural3dvideosynthesis} with camera effects in Fig.~\ref{fig:qualitative_realworld}. These results suggest that our method generalizes beyond synthetic scenes to real-world data.

\vspace{-0.5em}
\section{Applications}
\label{sec:application}

\begin{table}[t]
    \centering
    \caption{First-frame metrics of base vs.\ fine-tuned SVD models.}
    \label{tab:svd_quantitative_evaluation}
    \setlength{\tabcolsep}{3pt}
    \begin{tabular}{llccc}
        \toprule
        Effect & Model & PSNR(dB)$\uparrow$ & SSIM$\uparrow$ & LPIPS$\downarrow$ \\
        \midrule
        Fisheye & Base SVD & 12.1991 & 0.3568 & 0.5172 \\
                 & Fine-tuned SVD & \textbf{15.4661} & \textbf{0.4660} & \textbf{0.3136} \\
        \midrule
        DoF & Base SVD & 18.6120 & 0.6511 & 0.2618 \\
            & Fine-tuned SVD & \textbf{30.5652} & \textbf{0.9231} & \textbf{0.0958} \\
        \bottomrule
    \end{tabular}
    \vspace{-1em}
\end{table}

\vspace{-0.5em}

\noindent {\bf Fine-tuning Video Generation Models.}
4D-GRT generates dynamic videos paired with precise physical camera parameters at scale, providing training data to improve camera-effect synthesis in video generation models. To demonstrate this, we follow AKiRa~\cite{wang2025akira} and fine-tune Stable Video Diffusion (SVD)~\cite{blattmann2023stable} as a proof of concept. We construct a camera-effect dataset from MultiCamVideo~\cite{bai2025recammaster}, which contains synchronized 10-view videos of dynamic scenes. Specifically, we reconstruct 4D-GS for 287 scenes and render them under different camera parameters and configurations, producing 4,018 paired camera-effect videos for training. We reserve another 31 scenes for evaluation. The dataset includes fisheye, depth-of-field, zoom, and dolly-zoom effects; zoom and dolly zoom are simulated by adjusting the focal length and camera motion.
We compare the base and fine-tuned SVD models quantitatively and qualitatively. To isolate camera-effect synthesis from differences in generated motion, we compute PSNR, SSIM, and LPIPS only on the first generated frame. Both the quantitative results in Table~\ref{tab:svd_quantitative_evaluation} and the qualitative comparisons in Fig.~\ref{fig:svd_finetune} indicate that fine-tuning with our data improves SVD's ability to synthesize correct camera effects.

\noindent {\bf Camera-Effect Synthesis from Monocular Videos.}
Although designed for multi-view video input, our pipeline can be readily extended to accept monocular videos. We demonstrate this flexibility by reconstructing a 4D-GS representation from a monocular video with the pretrained MoVieS model~\cite{lin2026movies}, then synthesizing camera effects via ray tracing. Fig.~\ref{fig:supp_movies_realworld} shows the synthesized results. This extension broadens the range of video sources our method can use for large-scale camera-effect data generation.

\vspace{-0.5em}
\section{Conclusion}
\label{sec:conclusion}
\vspace{-0.5em}
In this work, we identify a critical problem that video generation models do not reliably follow camera-effect parameters, producing physically inconsistent videos. To address this, we introduce \textbf{4D Gaussian Ray Tracing (4D-GRT)}, a two-stage pipeline coupling 4D Gaussian Splatting with differentiable ray tracing for controllable, physically based camera-effect data generation. We also release \textbf{CamVFX-4D}, a benchmark of 8 dynamic indoor scenes with synchronized multi-view videos rendered under four camera effects. 4D-GRT achieves substantially higher rendering speed than strong Dynamic NeRF baselines across four camera effects while maintaining competitive or better image quality (PSNR/SSIM/LPIPS). We further demonstrate two practical uses of 4D-GRT: its generated paired data can improve camera-effect synthesis through video-model fine-tuning, and coupling its renderer with MoVieS enables camera-effect synthesis from monocular videos. Together, 4D-GRT and CamVFX-4D provide a strong foundation for advancing camera-aware vision. \ifarxiv See Appendix~\ref{supp:limitations} for a detailed discussion of limitations.\fi

\section*{Acknowledgements}
This research was supported in part by the National Science and Technology Council (NSTC), Taiwan, under Grants NSTC 115-2628-E-001-001-MY2, 114-2634-F-002-004, 115-2634-F-001-007, 115-2218-E-001-003, 115-2628-E-A49-007-, and 115-2634-F-A49-011-, and by Academia Sinica under Grant AS-IAIA-114-M10. We thank the National Center for High-performance Computing (NCHC) of National Applied Research Laboratories (NARLabs) in Taiwan for providing computational and storage resources, and Google, NVIDIA, and MediaTek Inc. for their generous donations. Yu-Lun Liu acknowledges the Yushan Young Fellow Program by the MOE in Taiwan.

{
    \small
    \bibliographystyle{ieeenat_fullname}
    \bibliography{main}

\begin{thebibliography}{98}
\providecommand{\natexlab}[1]{#1}
\providecommand{\url}[1]{\texttt{#1}}
\expandafter\ifx\csname urlstyle\endcsname\relax
  \providecommand{\doi}[1]{doi: #1}\else
  \providecommand{\doi}{doi: \begingroup \urlstyle{rm}\Url}\fi

\bibitem[3dhaupt(2019)]{free3d_pot_plant}
3dhaupt.
\newblock Indoor pot plant 2.
\newblock \url{https://free3d.com/3d-model/indoor-pot-plant-77983.html}, 2019.
\newblock Accessed: 2026-10-30, Licensed for Personal Use Only.

\bibitem[Bai et~al.(2025)Bai, Xia, Fu, Wang, Mu, Cao, Liu, Hu, Bai, Wan, and
  Zhang]{bai2025recammaster}
Jianhong Bai, Menghan Xia, Xiao Fu, Xintao Wang, Lianrui Mu, Jinwen Cao, Zuozhu
  Liu, Haoji Hu, Xiang Bai, Pengfei Wan, and Di Zhang.
\newblock {ReCamMaster}: Camera-controlled generative rendering from a single
  video.
\newblock In \emph{Proceedings of the IEEE/CVF International Conference on
  Computer Vision (ICCV)}, pages 14834--14844, 2025.

\bibitem[Ball et~al.(2025)Ball, Bauer, Belletti, Brownfield, Ephrat, Fruchter,
  Gupta, Holsheimer, Holynski, Hron, Kaplanis, Limont, McGill, Oliveira,
  Parker-Holder, Perbet, Scully, Shar, Spencer, Tov, Villegas, Wang, Yung,
  Baetu, Berbel, Bridson, Bruce, Buttimore, Chakera, Chandra, Collins, Cullum,
  Damoc, Dasagi, Gazeau, Gbadamosi, Han, Hirst, Kachra, Kerley, Kjems,
  Knoepfel, Koriakin, Lo, Lu, Mehring, Moufarek, Nandwani, Oliveira, Pardo,
  Park, Pierson, Poole, Ran, Salimans, Sanchez, Saprykin, Shen, Sidhwani,
  Smith, Stanton, Tomlinson, Vijaykumar, Wang, Wingfield, Wong, Xu, Yew, Young,
  Zubov, Eck, Erhan, Kavukcuoglu, Hassabis, Gharamani, Hadsell, van~den Oord,
  Mosseri, Bolton, Singh, and Rockt{\"a}schel]{genie3}
Philip~J. Ball, Jakob Bauer, Frank Belletti, Bethanie Brownfield, Ariel Ephrat,
  Shlomi Fruchter, Agrim Gupta, Kristian Holsheimer, Aleksander Holynski, Jiri
  Hron, Christos Kaplanis, Marjorie Limont, Matt McGill, Yanko Oliveira, Jack
  Parker-Holder, Frank Perbet, Guy Scully, Jeremy Shar, Stephen Spencer, Omer
  Tov, Ruben Villegas, Emma Wang, Jessica Yung, Cip Baetu, Jordi Berbel, David
  Bridson, Jake Bruce, Gavin Buttimore, Sarah Chakera, Bilva Chandra, Paul
  Collins, Alex Cullum, Bogdan Damoc, Vibha Dasagi, Maxime Gazeau, Charles
  Gbadamosi, Woohyun Han, Ed Hirst, Ashyana Kachra, Lucie Kerley, Kristian
  Kjems, Eva Knoepfel, Vika Koriakin, Jessica Lo, Cong Lu, Zeb Mehring, Alex
  Moufarek, Henna Nandwani, Valeria Oliveira, Fabio Pardo, Jane Park, Andrew
  Pierson, Ben Poole, Helen Ran, Tim Salimans, Manuel Sanchez, Igor Saprykin,
  Amy Shen, Sailesh Sidhwani, Duncan Smith, Joe Stanton, Hamish Tomlinson,
  Dimple Vijaykumar, Luyu Wang, Piers Wingfield, Nat Wong, Keyang Xu,
  Christopher Yew, Nick Young, Vadim Zubov, Douglas Eck, Dumitru Erhan, Koray
  Kavukcuoglu, Demis Hassabis, Zoubin Gharamani, Raia Hadsell, A{\"a}ron
  van~den Oord, Inbar Mosseri, Adrian Bolton, Satinder Singh, and Tim
  Rockt{\"a}schel.
\newblock {Genie 3}: A new frontier for world models.
\newblock
  \url{https://deepmind.google/blog/genie-3-a-new-frontier-for-world-models/},
  2025.
\newblock Accessed: 2026-10-30.

\bibitem[Barron et~al.(2021)Barron, Mildenhall, Tancik, Hedman, Martin-Brualla,
  and Srinivasan]{barron2021mip}
Jonathan~T. Barron, Ben Mildenhall, Matthew Tancik, Peter Hedman, Ricardo
  Martin-Brualla, and Pratul~P. Srinivasan.
\newblock {Mip-NeRF}: A multiscale representation for anti-aliasing neural
  radiance fields.
\newblock In \emph{Proceedings of the IEEE/CVF International Conference on
  Computer Vision (ICCV)}, pages 5855--5864, 2021.

\bibitem[Barron et~al.(2022)Barron, Mildenhall, Verbin, Srinivasan, and
  Hedman]{barron2022mipnerf360}
Jonathan~T. Barron, Ben Mildenhall, Dor Verbin, Pratul~P. Srinivasan, and Peter
  Hedman.
\newblock {Mip-NeRF} 360: Unbounded anti-aliased neural radiance fields.
\newblock In \emph{Proceedings of the IEEE/CVF Conference on Computer Vision
  and Pattern Recognition (CVPR)}, pages 5470--5479, 2022.

\bibitem[Blattmann et~al.(2023)Blattmann, Dockhorn, Kulal, Mendelevitch,
  Kilian, Lorenz, Levi, English, Voleti, Letts, Jampani, and
  Rombach]{blattmann2023stable}
Andreas Blattmann, Tim Dockhorn, Sumith Kulal, Daniel Mendelevitch, Maciej
  Kilian, Dominik Lorenz, Yam Levi, Zion English, Vikram Voleti, Adam Letts,
  Varun Jampani, and Robin Rombach.
\newblock Stable video diffusion: Scaling latent video diffusion models to
  large datasets.
\newblock \emph{arXiv preprint arXiv:2311.15127}, 2023.

\bibitem[{Blender Documentation Team}(2025)]{blender_manual_45}
{Blender Documentation Team}.
\newblock {Blender} 4.5 {LTS} manual.
\newblock \url{https://docs.blender.org/manual/en/4.5/}, 2025.
\newblock Accessed: 2026-10-30.

\bibitem[{Blender Online Community}(2025)]{blender_software}
{Blender Online Community}.
\newblock {Blender} -- a {3D} modelling and rendering package.
\newblock \url{http://www.blender.org}, 2025.
\newblock Version 4.5, Accessed: 2026-10-30.

\bibitem[{BlenderKit Community}(2025)]{blenderkit}
{BlenderKit Community}.
\newblock {BlenderKit} -- free {3D} assets for {Blender}.
\newblock \url{https://www.blenderkit.com}, 2025.
\newblock Accessed: 2026-10-30.

\bibitem[{BlendSwap Community}()]{blendswap}
{BlendSwap Community}.
\newblock {BlendSwap}: Free {Blender} {3D} models.
\newblock \url{https://www.blendswap.com}.
\newblock Accessed: 2026-10-30.

\bibitem[{ByteDance Seed}(2026)]{seedance25}
{ByteDance Seed}.
\newblock Seedance 2.5.
\newblock \url{https://seed.bytedance.com/en/seedance2_5}, 2026.
\newblock Accessed: 2026-10-30.

\bibitem[Cao and Johnson(2023)]{cao2023hexplane}
Ang Cao and Justin Johnson.
\newblock {HexPlane}: A fast representation for dynamic scenes.
\newblock In \emph{Proceedings of the IEEE/CVF Conference on Computer Vision
  and Pattern Recognition (CVPR)}, pages 130--141, 2023.

\bibitem[Carmichael and Skinner(2026)]{trgs_slam_2026}
Spencer Carmichael and Katherine~A. Skinner.
\newblock {TRGS-SLAM}: {IMU-Aided} {Gaussian} splatting {SLAM} for blurry,
  rolling shutter, and noisy thermal images.
\newblock \emph{arXiv preprint arXiv:2603.20443}, 2026.

\bibitem[Chang et~al.(2025)Chang, Fan, Chang, Lo, Tseng, Huang, and
  Liu]{chang2025gcc}
Chen-Wei Chang, Cheng-De Fan, Chia-Che Chang, Yi-Chen Lo, Yu-Chee Tseng,
  Jiun-Long Huang, and Yu-Lun Liu.
\newblock {GCC}: Generative color constancy via diffusing a color checker.
\newblock In \emph{Proceedings of the IEEE/CVF Conference on Computer Vision
  and Pattern Recognition (CVPR)}, pages 10868--10878, 2025.

\bibitem[Cheng et~al.(2024)Cheng, Chiu, and Liu]{cheng2024improving}
Bo-Yu Cheng, Wei-Chen Chiu, and Yu-Lun Liu.
\newblock Improving robustness for joint optimization of camera poses and
  decomposed low-rank tensorial radiance fields.
\newblock In \emph{Proceedings of the AAAI Conference on Artificial
  Intelligence (AAAI)}, pages 111:1--111:11, 2024.

\bibitem[Chu et~al.(2023)Chu, Lin, and Chen]{chu2023video}
Ernie Chu, Shuo-Yen Lin, and Jun-Cheng Chen.
\newblock Video {ControlNet}: Towards temporally consistent synthetic-to-real
  video translation using conditional image diffusion models.
\newblock \emph{arXiv preprint arXiv:2305.19193}, 2023.

\bibitem[Cook et~al.(1984)Cook, Porter, and Carpenter]{10.1145/964965.808590}
Robert~L. Cook, Thomas Porter, and Loren Carpenter.
\newblock Distributed ray tracing.
\newblock In \emph{Proceedings of the 11th Annual Conference on Computer
  Graphics and Interactive Techniques}, pages 137--145, 1984.

\bibitem[Croitoru et~al.(2023)Croitoru, Hondru, Ionescu, and
  Shah]{croitoru2023diffusion}
Florinel-Alin Croitoru, Vlad Hondru, Radu~Tudor Ionescu, and Mubarak Shah.
\newblock Diffusion models in vision: A survey.
\newblock \emph{IEEE transactions on pattern analysis and machine
  intelligence}, 45\penalty0 (9):\penalty0 10850--10869, 2023.

\bibitem[dailyfree3d(2019)]{free3d_cat_running}
dailyfree3d.
\newblock Lowpoly cat rigged + run animation.
\newblock
  \url{https://free3d.com/3d-model/lowpoly-cat-rigged-run-animation-756268.html},
  2019.
\newblock Accessed: 2026-10-30, Licensed for Personal Use Only.

\bibitem[Fan et~al.(2025)Fan, Chang, Liu, Lee, Huang, Tseng, and
  Liu]{fan2025spectromotion}
Cheng-De Fan, Chen-Wei Chang, Yi-Ruei Liu, Jie-Ying Lee, Jiun-Long Huang,
  Yu-Chee Tseng, and Yu-Lun Liu.
\newblock {SpectroMotion}: Dynamic {3D} reconstruction of specular scenes.
\newblock In \emph{Proceedings of the Computer Vision and Pattern Recognition
  Conference}, pages 21328--21338, 2025.

\bibitem[Fang and Chen(2025)]{fang2025camerabench}
I-Sheng Fang and Jun-Cheng Chen.
\newblock {CameraBench}: Benchmarking visual reasoning in {MLLMs} via
  photography.
\newblock \emph{arXiv preprint arXiv:2504.10090}, 2025.

\bibitem[Fang et~al.(2024)Fang, Han, and Chen]{fang2024camera}
I-Sheng Fang, Yue-Hua Han, and Jun-Cheng Chen.
\newblock Camera settings as tokens: Modeling photography on latent diffusion
  models.
\newblock In \emph{SIGGRAPH Asia 2024 Conference Papers}, pages 62:1--62:11,
  2024.

\bibitem[Fang et~al.(2022)Fang, Yi, Wang, Xie, Zhang, Liu, Nie{\ss}ner, and
  Tian]{fang2022fast}
Jiemin Fang, Taoran Yi, Xinggang Wang, Lingxi Xie, Xiaopeng Zhang, Wenyu Liu,
  Matthias Nie{\ss}ner, and Qi Tian.
\newblock Fast dynamic radiance fields with time-aware neural voxels.
\newblock In \emph{SIGGRAPH Asia 2022 Conference Papers}, pages 11:1--11:9,
  2022.

\bibitem[{Free3D}()]{free3d}
{Free3D}.
\newblock {Free3D}: Download free {3D} models.
\newblock \url{https://www.free3d.com}.
\newblock Accessed: 2026-10-30.

\bibitem[Fridovich-Keil et~al.(2023)Fridovich-Keil, Meanti, Warburg, Recht, and
  Kanazawa]{fridovich2023k}
Sara Fridovich-Keil, Giacomo Meanti, Frederik~Rahb{\ae}k Warburg, Benjamin
  Recht, and Angjoo Kanazawa.
\newblock {K-Planes}: Explicit radiance fields in space, time, and appearance.
\newblock In \emph{Proceedings of the IEEE/CVF Conference on Computer Vision
  and Pattern Recognition (CVPR)}, pages 12479--12488, 2023.

\bibitem[Gan et~al.(2024)Gan, Xu, Huang, Chen, and Yokoya]{gan2023v4d}
Wanshui Gan, Hongbin Xu, Yi Huang, Shifeng Chen, and Naoto Yokoya.
\newblock {V4D}: Voxel for {4D} novel view synthesis.
\newblock \emph{IEEE Transactions on Visualization and Computer Graphics},
  30\penalty0 (2):\penalty0 1579--1591, 2024.

\bibitem[Gao et~al.(2021)Gao, Saraf, Kopf, and Huang]{gao2021dynamic}
Chen Gao, Ayush Saraf, Johannes Kopf, and Jia-Bin Huang.
\newblock Dynamic view synthesis from dynamic monocular video.
\newblock In \emph{Proceedings of the IEEE/CVF International Conference on
  Computer Vision (ICCV)}, pages 5712--5721, 2021.

\bibitem[Gong et~al.(2024)Gong, Hou, Zhang, and Jiang]{gong2024beyond}
Yunpeng Gong, Yongjie Hou, Chuangliang Zhang, and Min Jiang.
\newblock Beyond augmentation: Empowering model robustness under extreme
  capture environments.
\newblock In \emph{2024 International Joint Conference on Neural Networks
  (IJCNN)}, pages 1--8, 2024.

\bibitem[{Google DeepMind}(2025)]{GoogleVeo3}
{Google DeepMind}.
\newblock {Veo 3} tech report.
\newblock
  \url{https://storage.googleapis.com/deepmind-media/veo/Veo-3-Tech-Report.pdf},
  2025.
\newblock Accessed: 2026-10-30.

\bibitem[He et~al.(2026)He, Zhang, Lin, Xu, and
  Pan]{he2025pretrainedvideogenerativemodels}
Haoran He, Yang Zhang, Liang Lin, Zhongwen Xu, and Ling Pan.
\newblock Pre-trained video generative models as world simulators.
\newblock In \emph{Proceedings of the AAAI Conference on Artificial
  Intelligence (AAAI)}, pages 517:1--517:9, 2026.

\bibitem[Heinzelnisse(2014)]{blendswap_lego_856_bulldozer}
Heinzelnisse.
\newblock {LEGO} 856 bulldozer.
\newblock \url{https://www.blendswap.com/blend/11490}, 2014.
\newblock Accessed: 2026-10-30, Licensed under CC-BY-NC 3.0.

\bibitem[Heiwolt(2025)]{plenoblendernerf}
Karoline Heiwolt.
\newblock {PlenoBlenderNeRF}.
\newblock \url{https://github.com/KarolineH/PlenoBlenderNeRF}, 2025.
\newblock Accessed: 2026-10-30.

\bibitem[Ho et~al.(2022{\natexlab{a}})Ho, Chan, Saharia, Whang, Gao, Gritsenko,
  Kingma, Poole, Norouzi, Fleet, and Salimans]{ho2022imagen}
Jonathan Ho, William Chan, Chitwan Saharia, Jay Whang, Ruiqi Gao, Alexey
  Gritsenko, Diederik~P. Kingma, Ben Poole, Mohammad Norouzi, David~J. Fleet,
  and Tim Salimans.
\newblock Imagen video: High definition video generation with diffusion models.
\newblock \emph{arXiv preprint arXiv:2210.02303}, 2022{\natexlab{a}}.

\bibitem[Ho et~al.(2022{\natexlab{b}})Ho, Salimans, Gritsenko, Chan, Norouzi,
  and Fleet]{ho2022video}
Jonathan Ho, Tim Salimans, Alexey Gritsenko, William Chan, Mohammad Norouzi,
  and David Fleet.
\newblock Video diffusion models.
\newblock In \emph{Advances in neural information processing systems}, pages
  8633--8646, 2022{\natexlab{b}}.

\bibitem[Hou et~al.(2025)Hou, Hsu, Huang, Shen, Sun, Sun, Chang, Liu, and
  Lee]{10887619}
Hao-Yu Hou, Chia-Chi Hsu, Yu-Chen Huang, Mu-Yi Shen, Wei-Fang Sun, Cheng Sun,
  Chia-Che Chang, Yu-Lun Liu, and Chun-Yi Lee.
\newblock {3D} {Gaussian} splatting with grouped uncertainty for unconstrained
  images.
\newblock In \emph{ICASSP 2025 - 2025 IEEE International Conference on
  Acoustics, Speech and Signal Processing (ICASSP)}, pages 1--5, 2025.

\bibitem[Hu et~al.(2023)Hu, Russell, Yeo, Murez, Fedoseev, Kendall, Shotton,
  and Corrado]{hu2023gaia1generativeworldmodel}
Anthony Hu, Lloyd Russell, Hudson Yeo, Zak Murez, George Fedoseev, Alex
  Kendall, Jamie Shotton, and Gianluca Corrado.
\newblock {GAIA-1}: A generative world model for autonomous driving.
\newblock \emph{arXiv preprint arXiv:2309.17080}, 2023.

\bibitem[Hu et~al.(2027)Hu, Jia, Guo, Zhang, Long, and
  Yin]{hu2024drivingworldconstructingworldmodel}
Xiaotao Hu, Mingkai Jia, Xiaoyang Guo, Qian Zhang, Xiao-xiao Long, and Wei Yin.
\newblock {DrivingWorld}: Constructing world model for autonomous driving via
  video {GPT}.
\newblock In \emph{Pattern Recognition}, pages 276--291, 2027.

\bibitem[Huang et~al.(2024)Huang, Sun, Yang, Lyu, Cao, and Qi]{huang2023sc}
Yi-Hua Huang, Yang-Tian Sun, Ziyi Yang, Xiaoyang Lyu, Yan-Pei Cao, and Xiaojuan
  Qi.
\newblock {SC-GS}: Sparse-controlled {Gaussian} splatting for editable dynamic
  scenes.
\newblock In \emph{Proceedings of the IEEE/CVF Conference on Computer Vision
  and Pattern Recognition (CVPR)}, pages 4220--4230, 2024.

\bibitem[Huang et~al.(2026)Huang, Wu, Guo, Huang, and
  Ren]{huang20263dgeer3dgaussianrendering}
Zixun Huang, Cho-Ying Wu, Yuliang Guo, Xinyu Huang, and Liu Ren.
\newblock {3DGEER}: {3D} {Gaussian} rendering made exact and efficient for
  generic cameras.
\newblock In \emph{International Conference on Learning Representations
  (ICLR)}, 2026.

\bibitem[Huynh-Thu and Ghanbari(2008)]{huynh2008scope}
Q. Huynh-Thu and M. Ghanbari.
\newblock Scope of validity of {PSNR} in image/video quality assessment.
\newblock \emph{Electronics Letters}, 44:\penalty0 800--801, 2008.

\bibitem[Ignatov et~al.(2020)Ignatov, Patel, and Timofte]{ignatov2020rendering}
Andrey Ignatov, Jagruti Patel, and Radu Timofte.
\newblock Rendering natural camera bokeh effect with deep learning.
\newblock In \emph{Proceedings of the IEEE/CVF Conference on Computer Vision
  and Pattern Recognition (CVPR) Workshops}, 2020.

\bibitem[Ke et~al.(2025)Ke, Xie, Liu, and Chiu]{ke2025stealthattack}
Bo-Hsu Ke, You-Zhe Xie, Yu-Lun Liu, and Wei-Chen Chiu.
\newblock {StealthAttack}: Robust {3D} {Gaussian} splatting poisoning via
  density-guided illusions.
\newblock In \emph{Proceedings of the IEEE/CVF International Conference on
  Computer Vision}, pages 27400--27411, 2025.

\bibitem[Kerbl et~al.(2023)Kerbl, Kopanas, Leimkuehler, and
  Drettakis]{kerbl20233d}
Bernhard Kerbl, Georgios Kopanas, Thomas Leimkuehler, and George Drettakis.
\newblock {3D} {Gaussian} splatting for real-time radiance field rendering.
\newblock \emph{ACM Transactions on Graphics}, 42\penalty0 (4):\penalty0
  139:1--139:14, 2023.

\bibitem[Kratimenos et~al.(2025)Kratimenos, Lei, and
  Daniilidis]{kratimenos2024dynmf}
Agelos Kratimenos, Jiahui Lei, and Kostas Daniilidis.
\newblock {DynMF}: Neural motion factorization for real-time dynamic view
  synthesis with {3D} {Gaussian} splatting.
\newblock In \emph{Computer Vision -- ECCV 2024}, pages 252--269, 2025.

\bibitem[Lee et~al.(2025)Lee, Cho, Kim, Jang, Lee, Cha, Wee, Lee, and
  Lee]{lee2025cocogaussian}
Jungho Lee, Suhwan Cho, Taeoh Kim, Ho-Deok Jang, Minhyeok Lee, Geonho Cha,
  Dongyoon Wee, Dogyoon Lee, and Sangyoun Lee.
\newblock {CoCoGaussian}: Leveraging circle of confusion for {Gaussian}
  splatting from defocused images.
\newblock In \emph{Proceedings of the IEEE/CVF Conference on Computer Vision
  and Pattern Recognition (CVPR)}, pages 16101--16110, 2025.

\bibitem[Li et~al.(2022)Li, Slavcheva, Zollh{\"o}fer, Green, Lassner, Kim,
  Schmidt, Lovegrove, Goesele, Newcombe, and Lv]{li2022neural3dvideosynthesis}
Tianye Li, Mira Slavcheva, Michael Zollh{\"o}fer, Simon Green, Christoph
  Lassner, Changil Kim, Tanner Schmidt, Steven Lovegrove, Michael Goesele,
  Richard Newcombe, and Zhaoyang Lv.
\newblock Neural {3D} video synthesis from multi-view video.
\newblock In \emph{Proceedings of the IEEE/CVF Conference on Computer Vision
  and Pattern Recognition (CVPR)}, pages 5521--5531, 2022.

\bibitem[Liang et~al.(2025)Liang, Khan, Li, Nguyen-Phuoc, Lanman, Tompkin, and
  Xiao]{liang2024gaufregaussiandeformationfields}
Yiqing Liang, Numair Khan, Zhengqin Li, Thu~H Nguyen-Phuoc, Douglas Lanman,
  James Tompkin, and Lei Xiao.
\newblock {GauFRe}: {Gaussian} deformation fields for real-time dynamic novel
  view synthesis.
\newblock In \emph{Proceedings of the IEEE/CVF Winter Conference on
  Applications of Computer Vision (WACV)}, pages 2642--2652, 2025.

\bibitem[Liao et~al.(2024)Liao, Chen, Fu, Wang, Su, Luo, Ma, Xu, Dai, Li, Pei,
  and Zhang]{liao2024fisheyegslightweightextensiblegaussian}
Zimu Liao, Siyan Chen, Rong Fu, Yi Wang, Zhongling Su, Hao Luo, Li Ma, Linning
  Xu, Bo Dai, Hengjie Li, Zhilin Pei, and Xingcheng Zhang.
\newblock {Fisheye-GS}: Lightweight and extensible {Gaussian} splatting module
  for fisheye cameras.
\newblock \emph{arXiv preprint arXiv:2409.04751}, 2024.

\bibitem[Lin et~al.(2026)Lin, Lin, Pan, Yu, Hu, Yan, Fragkiadaki, and
  Mu]{lin2026movies}
Chenguo Lin, Yuchen Lin, Panwang Pan, Yifan Yu, Tao Hu, Honglei Yan, Katerina
  Fragkiadaki, and Yadong Mu.
\newblock {MoVieS}: Motion-aware {4D} dynamic view synthesis in one second.
\newblock In \emph{Proceedings of the IEEE/CVF Conference on Computer Vision
  and Pattern Recognition (CVPR)}, pages 295--306, 2026.

\bibitem[Lin et~al.(2025)Lin, Wu, Yeh, Yen, Sun, and Liu]{lin2025frugalnerf}
Chin-Yang Lin, Chung-Ho Wu, Chang-Han Yeh, Shih-Han Yen, Cheng Sun, and Yu-Lun
  Liu.
\newblock {FrugalNeRF}: Fast convergence for extreme few-shot novel view
  synthesis without learned priors.
\newblock In \emph{Proceedings of the IEEE/CVF Conference on Computer Vision
  and Pattern Recognition (CVPR)}, pages 11227--11238, 2025.

\bibitem[Lin et~al.(2024)Lin, Dai, Zhu, and Yao]{lin2024gaussian}
Youtian Lin, Zuozhuo Dai, Siyu Zhu, and Yao Yao.
\newblock {Gaussian-Flow}: {4D} reconstruction with dynamic {3D} {Gaussian}
  particle.
\newblock In \emph{Proceedings of the IEEE/CVF Conference on Computer Vision
  and Pattern Recognition (CVPR)}, pages 21136--21145, 2024.

\bibitem[Liu et~al.(2023)Liu, Gao, Meuleman, Tseng, Saraf, Kim, Chuang, Kopf,
  and Huang]{liu2023robust}
Yu-Lun Liu, Chen Gao, Andr{\'e}as Meuleman, Hung-Yu Tseng, Ayush Saraf, Changil
  Kim, Yung-Yu Chuang, Johannes Kopf, and Jia-Bin Huang.
\newblock Robust dynamic radiance fields.
\newblock In \emph{Proceedings of the IEEE/CVF Conference on Computer Vision
  and Pattern Recognition (CVPR)}, pages 13--23, 2023.

\bibitem[Luiten et~al.(2024)Luiten, Kopanas, Leibe, and
  Ramanan]{luiten2024dynamic}
Jonathon Luiten, Georgios Kopanas, Bastian Leibe, and Deva Ramanan.
\newblock Dynamic {3D} {Gaussians}: Tracking by persistent dynamic view
  synthesis.
\newblock In \emph{2024 International Conference on 3D Vision (3DV)}, pages
  800--809, 2024.

\bibitem[Mildenhall et~al.(2021)Mildenhall, Srinivasan, Tancik, Barron,
  Ramamoorthi, and Ng]{mildenhall2021nerf}
Ben Mildenhall, Pratul~P. Srinivasan, Matthew Tancik, Jonathan~T. Barron, Ravi
  Ramamoorthi, and Ren Ng.
\newblock {NeRF}: representing scenes as neural radiance fields for view
  synthesis.
\newblock \emph{Communications of the ACM}, 65\penalty0 (1):\penalty0 99--106,
  2021.

\bibitem[Moenne-Loccoz et~al.(2024)Moenne-Loccoz, Mirzaei, Perel, de~Lutio,
  Martinez~Esturo, State, Fidler, Sharp, and Gojcic]{3dgrt2024}
Nicolas Moenne-Loccoz, Ashkan Mirzaei, Or Perel, Riccardo de Lutio, Janick
  Martinez~Esturo, Gavriel State, Sanja Fidler, Nicholas Sharp, and Zan Gojcic.
\newblock {3D} {Gaussian} ray tracing: Fast tracing of particle scenes.
\newblock \emph{ACM Transactions on Graphics}, 43\penalty0 (6):\penalty0
  232:1--232:19, 2024.

\bibitem[M{\"u}ller et~al.(2022)M{\"u}ller, Evans, Schied, and
  Keller]{muller2022instant}
Thomas M{\"u}ller, Alex Evans, Christoph Schied, and Alexander Keller.
\newblock Instant neural graphics primitives with a multiresolution hash
  encoding.
\newblock \emph{ACM Transactions on Graphics (TOG)}, 41\penalty0 (4):\penalty0
  102:1--102:15, 2022.

\bibitem[Nikolenko(2021)]{nikolenko2021synthetic}
Sergey~I. Nikolenko.
\newblock \emph{Synthetic Data for Deep Learning}.
\newblock Springer International Publishing, 2021.

\bibitem[Niu et~al.(2025)Niu, Chen, Zhan, Li, Ji, and
  Zheng]{niu2024rsnerfneuralradiancefields}
Muyao Niu, Tong Chen, Yifan Zhan, Zhuoxiao Li, Xiang Ji, and Yinqiang Zheng.
\newblock {RS-NeRF}: Neural radiance fields from rolling shutter images.
\newblock In \emph{Computer Vision -- ECCV 2024}, pages 163--180, 2025.

\bibitem[Oh et~al.(2025)Oh, Lee, Jeon, and Park]{oh2025hybrid}
Seungjun Oh, Younggeun Lee, Hyejin Jeon, and Eunbyung Park.
\newblock Hybrid {3D-4D} {Gaussian} splatting for fast dynamic scene
  representation.
\newblock \emph{arXiv preprint arXiv:2505.13215}, 2025.

\bibitem[{OpenAI}(2025)]{OpenAISora2}
{OpenAI}.
\newblock Sora 2 is here.
\newblock \url{https://openai.com/index/sora-2/}, 2025.
\newblock Accessed: 2026-10-30.

\bibitem[Park et~al.(2021{\natexlab{a}})Park, Sinha, Barron, Bouaziz, Goldman,
  Seitz, and Martin-Brualla]{park2021nerfies}
Keunhong Park, Utkarsh Sinha, Jonathan~T. Barron, Sofien Bouaziz, Dan~B
  Goldman, Steven~M. Seitz, and Ricardo Martin-Brualla.
\newblock {Nerfies}: Deformable neural radiance fields.
\newblock In \emph{Proceedings of the IEEE/CVF International Conference on
  Computer Vision (ICCV)}, pages 5865--5874, 2021{\natexlab{a}}.

\bibitem[Park et~al.(2021{\natexlab{b}})Park, Sinha, Hedman, Barron, Bouaziz,
  Goldman, Martin-Brualla, and Seitz]{park2021hypernerf}
Keunhong Park, Utkarsh Sinha, Peter Hedman, Jonathan~T. Barron, Sofien Bouaziz,
  Dan~B Goldman, Ricardo Martin-Brualla, and Steven~M. Seitz.
\newblock {HyperNeRF}: a higher-dimensional representation for topologically
  varying neural radiance fields.
\newblock \emph{ACM Transactions on Graphics}, 40\penalty0 (6):\penalty0
  238:1--238:12, 2021{\natexlab{b}}.

\bibitem[Perazzi et~al.(2016)Perazzi, Pont-Tuset, McWilliams, Van~Gool, Gross,
  and Sorkine-Hornung]{Perazzi_CVPR_2016}
Federico Perazzi, Jordi Pont-Tuset, Brian McWilliams, Luc Van~Gool, Markus
  Gross, and Alexander Sorkine-Hornung.
\newblock A benchmark dataset and evaluation methodology for video object
  segmentation.
\newblock In \emph{Proceedings of the IEEE Conference on Computer Vision and
  Pattern Recognition (CVPR)}, 2016.

\bibitem[Pumarola et~al.(2021)Pumarola, Corona, Pons-Moll, and
  Moreno-Noguer]{pumarola2021d}
Albert Pumarola, Enric Corona, Gerard Pons-Moll, and Francesc Moreno-Noguer.
\newblock {D-NeRF}: Neural radiance fields for dynamic scenes.
\newblock In \emph{Proceedings of the IEEE/CVF Conference on Computer Vision
  and Pattern Recognition (CVPR)}, pages 10318--10327, 2021.

\bibitem[Raafat(2024)]{Raafat_BlenderNeRF_2024}
Maxime Raafat.
\newblock {BlenderNeRF}.
\newblock \url{https://doi.org/10.5281/zenodo.13250725}, 2024.
\newblock Version v6, Zenodo.

\bibitem[Rombach et~al.(2022)Rombach, Blattmann, Lorenz, Esser, and
  Ommer]{rombach2022high}
Robin Rombach, Andreas Blattmann, Dominik Lorenz, Patrick Esser, and Bj{\"o}rn
  Ommer.
\newblock High-resolution image synthesis with latent diffusion models.
\newblock In \emph{Proceedings of the IEEE/CVF Conference on Computer Vision
  and Pattern Recognition (CVPR)}, pages 10684--10695, 2022.

\bibitem[RonKo(2023)]{blendswap_box_animation_template}
RonKo.
\newblock Box animation template.
\newblock \url{https://www.blendswap.com/blend/30950}, 2023.
\newblock Accessed: 2026-10-30, Licensed under CC-BY 4.0.

\bibitem[Saurer et~al.(2013)Saurer, Koser, Bouguet, and
  Pollefeys]{saurer2013rolling}
Olivier Saurer, Kevin Koser, Jean-Yves Bouguet, and Marc Pollefeys.
\newblock Rolling shutter stereo.
\newblock In \emph{Proceedings of the IEEE International Conference on Computer
  Vision (ICCV)}, 2013.

\bibitem[Seiskari et~al.(2025)Seiskari, Ylilammi, Kaatrasalo, Rantalankila,
  Turkulainen, Kannala, Rahtu, and Solin]{seiskari2024gaussian}
Otto Seiskari, Jerry Ylilammi, Valtteri Kaatrasalo, Pekka Rantalankila, Matias
  Turkulainen, Juho Kannala, Esa Rahtu, and Arno Solin.
\newblock {Gaussian} splatting on the move: Blur and rolling shutter
  compensation for natural camera motion.
\newblock In \emph{Computer Vision -- ECCV 2024}, pages 160--177, 2025.

\bibitem[Shen et~al.(2025)Shen, Liu, Sun, Li, Cao, Li, and
  Loy]{shen2025dofgaussiancontrollabledepthoffield3d}
Liao Shen, Tianqi Liu, Huiqiang Sun, Jiaqi Li, Zhiguo Cao, Wei Li, and
  Chen~Change Loy.
\newblock {DoF-Gaussian}: Controllable depth-of-field for {3D} {Gaussian}
  splatting.
\newblock In \emph{Proceedings of the IEEE/CVF Conference on Computer Vision
  and Pattern Recognition (CVPR)}, pages 26462--26471, 2025.

\bibitem[Song et~al.(2023)Song, Chen, Li, Chen, Chen, Yuan, Xu, and
  Geiger]{nerfplayer}
Liangchen Song, Anpei Chen, Zhong Li, Zhang Chen, Lele Chen, Junsong Yuan, Yi
  Xu, and Andreas Geiger.
\newblock {NeRFPlayer}: A streamable dynamic scene representation with
  decomposed neural radiance fields.
\newblock \emph{IEEE Transactions on Visualization and Computer Graphics},
  29\penalty0 (5):\penalty0 2732--2742, 2023.

\bibitem[Su et~al.(2024)Su, Hu, Tsai, Lee, Lin, and Liu]{su2024boostmvsnerfs}
Chih-Hai Su, Chih-Yao Hu, Shr-Ruei Tsai, Jie-Ying Lee, Chin-Yang Lin, and
  Yu-Lun Liu.
\newblock {BoostMVSNeRFs}: Boosting {MVS-based} {NeRFs} to generalizable view
  synthesis in large-scale scenes.
\newblock In \emph{ACM SIGGRAPH 2024 Conference Papers}, pages 31:1--31:12,
  2024.

\bibitem[Sun et~al.(2022)Sun, Sun, and Chen]{SunSC22}
Cheng Sun, Min Sun, and Hwann-Tzong Chen.
\newblock Direct voxel grid optimization: Super-fast convergence for radiance
  fields reconstruction.
\newblock In \emph{Proceedings of the IEEE/CVF Conference on Computer Vision
  and Pattern Recognition (CVPR)}, pages 5459--5469, 2022.

\bibitem[Tremblay et~al.(2018)Tremblay, Prakash, Acuna, Brophy, Jampani, Anil,
  To, Cameracci, Boochoon, and Birchfield]{tremblay2018training}
Jonathan Tremblay, Aayush Prakash, David Acuna, Mark Brophy, Varun Jampani, Cem
  Anil, Thang To, Eric Cameracci, Shaad Boochoon, and Stan Birchfield.
\newblock Training deep networks with synthetic data: Bridging the reality gap
  by domain randomization.
\newblock In \emph{Proceedings of the IEEE Conference on Computer Vision and
  Pattern Recognition (CVPR) Workshops}, 2018.

\bibitem[Voynov et~al.(2024)Voynov, Hertz, Arar, Fruchter, and
  Cohen-Or]{voynov2024curved}
Andrey Voynov, Amir Hertz, Moab Arar, Shlomi Fruchter, and Daniel Cohen-Or.
\newblock Curved diffusion: A generative model with optical geometry control.
\newblock In \emph{Computer Vision -- ECCV 2024}, pages 149--164, 2024.

\bibitem[Wan et~al.(2025)Wan, Wang, Ai, Wen, Mao, Xie, Chen, Yu, Zhao, Yang,
  Zeng, Wang, Zhang, Zhou, Wang, Chen, Zhu, Zhao, Yan, Huang, Feng, Zhang, Li,
  Wu, Chu, Feng, Zhang, Sun, Fang, Wang, Gui, Weng, Shen, Lin, Wang, Wang,
  Zhou, Wang, Shen, Yu, Shi, Huang, Xu, Kou, Lv, Li, Liu, Wang, Zhang, Huang,
  Li, Wu, Liu, Pan, Zheng, Hong, Shi, Feng, Jiang, Han, Wu, and Liu]{wan2025}
Team Wan, Ang Wang, Baole Ai, Bin Wen, Chaojie Mao, Chen-Wei Xie, Di Chen,
  Feiwu Yu, Haiming Zhao, Jianxiao Yang, Jianyuan Zeng, Jiayu Wang, Jingfeng
  Zhang, Jingren Zhou, Jinkai Wang, Jixuan Chen, Kai Zhu, Kang Zhao, Keyu Yan,
  Lianghua Huang, Mengyang Feng, Ningyi Zhang, Pandeng Li, Pingyu Wu, Ruihang
  Chu, Ruili Feng, Shiwei Zhang, Siyang Sun, Tao Fang, Tianxing Wang, Tianyi
  Gui, Tingyu Weng, Tong Shen, Wei Lin, Wei Wang, Wei Wang, Wenmeng Zhou, Wente
  Wang, Wenting Shen, Wenyuan Yu, Xianzhong Shi, Xiaoming Huang, Xin Xu, Yan
  Kou, Yangyu Lv, Yifei Li, Yijing Liu, Yiming Wang, Yingya Zhang, Yitong
  Huang, Yong Li, You Wu, Yu Liu, Yulin Pan, Yun Zheng, Yuntao Hong, Yupeng
  Shi, Yutong Feng, Zeyinzi Jiang, Zhen Han, Zhi-Fan Wu, and Ziyu Liu.
\newblock Wan: Open and advanced large-scale video generative models.
\newblock \emph{arXiv preprint arXiv:2503.20314}, 2025.

\bibitem[Wang et~al.(2023)Wang, Chen, Wang, Song, and Liu]{wang2023masked}
Feng Wang, Zilong Chen, Guokang Wang, Yafei Song, and Huaping Liu.
\newblock Masked space-time hash encoding for efficient dynamic scene
  reconstruction.
\newblock In \emph{Advances in neural information processing systems}, pages
  70497--70510, 2023.

\bibitem[Wang et~al.(2024)Wang, Veldhuis, Brune, and
  Strisciuglio]{wang2023survey}
Shunxin Wang, Raymond Veldhuis, Christoph Brune, and Nicola Strisciuglio.
\newblock A survey on the robustness of computer vision models against common
  corruptions.
\newblock \emph{arXiv preprint arXiv:2305.06024}, 2024.

\bibitem[Wang et~al.(2025{\natexlab{a}})Wang, Courant, Christie, and
  Kalogeiton]{wang2025akira}
Xi Wang, Robin Courant, Marc Christie, and Vicky Kalogeiton.
\newblock {AKiRa}: Augmentation kit on rays for optical video generation.
\newblock In \emph{Proceedings of the IEEE/CVF Conference on Computer Vision
  and Pattern Recognition (CVPR)}, pages 2609--2619, 2025{\natexlab{a}}.

\bibitem[Wang et~al.(2025{\natexlab{b}})Wang, Chakravarthula, and
  Chen]{wang2025dofgsadjustabledepthoffield3dgaussian}
Yujie Wang, Praneeth Chakravarthula, and Baoquan Chen.
\newblock {DOF-GS}: Adjustable depth-of-field {3D} {Gaussian} splatting for
  post-capture refocusing, defocus rendering and blur removal.
\newblock In \emph{Proceedings of the IEEE/CVF Conference on Computer Vision
  and Pattern Recognition (CVPR)}, pages 21297--21306, 2025{\natexlab{b}}.

\bibitem[Wang et~al.(2004)Wang, Bovik, Sheikh, and Simoncelli]{wang2004image}
Zhou Wang, A.C. Bovik, H.R. Sheikh, and E.P. Simoncelli.
\newblock Image quality assessment: from error visibility to structural
  similarity.
\newblock \emph{IEEE Transactions on Image Processing}, 13\penalty0
  (4):\penalty0 600--612, 2004.

\bibitem[Wu et~al.(2024)Wu, Yi, Fang, Xie, Zhang, Wei, Liu, Tian, and
  Wang]{wu20244d}
Guanjun Wu, Taoran Yi, Jiemin Fang, Lingxi Xie, Xiaopeng Zhang, Wei Wei, Wenyu
  Liu, Qi Tian, and Xinggang Wang.
\newblock {4D} {Gaussian} splatting for real-time dynamic scene rendering.
\newblock In \emph{Proceedings of the IEEE/CVF Conference on Computer Vision
  and Pattern Recognition (CVPR)}, pages 20310--20320, 2024.

\bibitem[Wu et~al.(2025{\natexlab{a}})Wu, Peng, Wang, Xiao, Tang, Yan, Xiong,
  and Wang]{wu2025swiftd}
Jiahao Wu, Rui Peng, Zhiyan Wang, Lu Xiao, Luyang Tang, Jinbo Yan, Kaiqiang
  Xiong, and Ronggang Wang.
\newblock {Swift4D}: Adaptive divide-and-conquer {Gaussian} splatting for
  compact and efficient reconstruction of dynamic scene.
\newblock In \emph{The Thirteenth International Conference on Learning
  Representations}, 2025{\natexlab{a}}.

\bibitem[Wu et~al.(2025{\natexlab{b}})Wu, Esturo, Mirzaei, Mo{\"e}nne-Loccoz,
  and Gojcic]{wu20253dgut}
Qi Wu, Janick~Martinez Esturo, Ashkan Mirzaei, Nicolas Mo{\"e}nne-Loccoz, and
  Zan Gojcic.
\newblock {3DGUT}: Enabling distorted cameras and secondary rays in {Gaussian}
  splatting.
\newblock In \emph{Proceedings of the IEEE/CVF Conference on Computer Vision
  and Pattern Recognition (CVPR)}, pages 26036--26046, 2025{\natexlab{b}}.

\bibitem[Wu et~al.(2022)Wu, Li, Peng, Lu, Cao, and Zhong]{Wu_2022}
Zijin Wu, Xingyi Li, Juewen Peng, Hao Lu, Zhiguo Cao, and Weicai Zhong.
\newblock {DoF-NeRF}: Depth-of-field meets neural radiance fields.
\newblock In \emph{Proceedings of the 30th ACM International Conference on
  Multimedia}, pages 1718--1729, 2022.

\bibitem[Xing et~al.(2024)Xing, Feng, Chen, Dai, Hu, Xu, Wu, and
  Jiang]{xing2024survey}
Zhen Xing, Qijun Feng, Haoran Chen, Qi Dai, Han Hu, Hang Xu, Zuxuan Wu, and
  Yu-Gang Jiang.
\newblock A survey on video diffusion models.
\newblock \emph{ACM Computing Surveys}, 57\penalty0 (2):\penalty0 41:1--41:42,
  2024.

\bibitem[Yan et~al.(2023)Yan, Li, and Lee]{yan2023nerf}
Zhiwen Yan, Chen Li, and Gim~Hee Lee.
\newblock {NeRF-DS}: Neural radiance fields for dynamic specular objects.
\newblock In \emph{Proceedings of the IEEE/CVF Conference on Computer Vision
  and Pattern Recognition (CVPR)}, pages 8285--8295, 2023.

\bibitem[Yang et~al.(2024{\natexlab{a}})Yang, Gao, Zhou, Jiao, Zhang, and
  Jin]{yang2024deformable}
Ziyi Yang, Xinyu Gao, Wen Zhou, Shaohui Jiao, Yuqing Zhang, and Xiaogang Jin.
\newblock Deformable {3D} {Gaussians} for high-fidelity monocular dynamic scene
  reconstruction.
\newblock In \emph{Proceedings of the IEEE/CVF Conference on Computer Vision
  and Pattern Recognition (CVPR)}, pages 20331--20341, 2024{\natexlab{a}}.

\bibitem[Yang et~al.(2024{\natexlab{b}})Yang, Yang, Pan, and
  Zhang]{yang2024realtimephotorealisticdynamicscene}
Zeyu Yang, Hongye Yang, Zijie Pan, and Li Zhang.
\newblock Real-time photorealistic dynamic scene representation and rendering
  with {4D} {Gaussian} splatting.
\newblock In \emph{International Conference on Learning Representations
  (ICLR)}, 2024{\natexlab{b}}.

\bibitem[Yeh et~al.(2025)Yeh, Lee, and cheng Chen]{yeh2024training}
Po-Hung Yeh, Kuang-Huei Lee, and Jun cheng Chen.
\newblock Training-free diffusion model alignment with sampling demons.
\newblock In \emph{International Conference on Learning Representations
  (ICLR)}, 2025.

\bibitem[Yifan et~al.(2019)Yifan, Serena, Wu, {\"{O}}ztireli, and
  Sorkine{-}Hornung]{Yifan:DSS:2019}
Wang Yifan, Felice Serena, Shihao Wu, Cengiz {\"{O}}ztireli, and Olga
  Sorkine{-}Hornung.
\newblock Differentiable surface splatting for point-based geometry processing.
\newblock \emph{ACM Transactions on Graphics (proceedings of ACM SIGGRAPH
  ASIA)}, 38\penalty0 (6), 2019.

\bibitem[Yu et~al.(2021)Yu, Ye, Tancik, and Kanazawa]{yu2021pixelnerf}
Alex Yu, Vickie Ye, Matthew Tancik, and Angjoo Kanazawa.
\newblock {pixelNeRF}: Neural radiance fields from one or few images.
\newblock In \emph{Proceedings of the IEEE/CVF Conference on Computer Vision
  and Pattern Recognition (CVPR)}, pages 4578--4587, 2021.

\bibitem[Zhan et~al.(2025)Zhan, Ho, Yang, Chen, Chiang, Liu, and
  Peng]{zhan2025cat}
Yu-Ting Zhan, Cheng-Yuan Ho, Hebi Yang, Yi-Hsin Chen, Jui~Chiu Chiang, Yu-Lun
  Liu, and Wen-Hsiao Peng.
\newblock {CAT-3DGS}: A context-adaptive triplane approach to
  rate-distortion-optimized {3DGS} compression.
\newblock In \emph{International Conference on Learning Representations
  (ICLR)}, 2025.

\bibitem[Zhang et~al.(2020)Zhang, Riegler, Snavely, and
  Koltun]{zhang2020nerf++}
Kai Zhang, Gernot Riegler, Noah Snavely, and Vladlen Koltun.
\newblock {NeRF++}: Analyzing and improving neural radiance fields.
\newblock \emph{arXiv preprint arXiv:2010.07492}, 2020.

\bibitem[Zhang et~al.(2018)Zhang, Isola, Efros, Shechtman, and
  Wang]{zhang2018unreasonable}
Richard Zhang, Phillip Isola, Alexei~A. Efros, Eli Shechtman, and Oliver Wang.
\newblock The unreasonable effectiveness of deep features as a perceptual
  metric.
\newblock In \emph{Proceedings of the IEEE Conference on Computer Vision and
  Pattern Recognition (CVPR)}, 2018.

\bibitem[Zhang et~al.(2025)Zhang, Xiao, and Zhang]{dydeblur}
Xuankai Zhang, Junjin Xiao, and Qing Zhang.
\newblock Dynamic {Gaussian} splatting from defocused and motion-blurred
  monocular videos.
\newblock In \emph{Advances in Neural Information Processing Systems}, pages
  71096--71121, 2025.

\bibitem[Zhong et~al.(2021)Zhong, Zheng, and Sato]{zhong2021towards}
Zhihang Zhong, Yinqiang Zheng, and Imari Sato.
\newblock Towards rolling shutter correction and deblurring in dynamic scenes.
\newblock In \emph{Proceedings of the IEEE/CVF Conference on Computer Vision
  and Pattern Recognition (CVPR)}, pages 9219--9228, 2021.

\bibitem[Zhou et~al.(2018)Zhou, Tucker, Flynn, Fyffe, and
  Snavely]{zhou2018stereo}
Tinghui Zhou, Richard Tucker, John Flynn, Graham Fyffe, and Noah Snavely.
\newblock Stereo magnification: learning view synthesis using multiplane
  images.
\newblock \emph{ACM Transactions on Graphics}, 37\penalty0 (4):\penalty0
  65:1--65:12, 2018.

\end{thebibliography}
}

\clearpage
\appendix
\section*{Appendix Overview}
These appendices present additional experimental results and technical details. We begin by describing the construction of CamVFX-4D in Sec.~\ref{supp:dataset}. We then discuss the video generation experiment in Sec.~\ref{supp:video_gen}, followed by implementation details in Sec.~\ref{supp:implem_detail}, additional ablations in Sec.~\ref{supp:add_ablations}, additional quantitative results in Sec.~\ref{supp:add_quant}, additional qualitative results in Sec.~\ref{supp:add_qual}, camera-effect synthesis from monocular videos in Sec.~\ref{supp:movies}, and limitations in Sec.~\ref{supp:limitations}. Videos of the results are available on the project page.

\section{CamVFX-4D Dataset Construction}
\label{supp:dataset}
\vspace{-0.5em}
In practice, capturing the same dynamic scene with different camera effects is infeasible; retakes change motion timing, illumination, and other conditions. Public datasets therefore lack multi-view, same-scene recordings under multiple effects.

To fill this gap, we construct \textbf{CamVFX-4D}, a synthetic multi-view dataset, in Blender 4.5~\cite{blender_software}. CamVFX-4D contains 8 dynamic indoor scenes across 4 environments (basketball court, warehouse, living room, bathroom), each with 50 viewpoints and 50 frames at 512$\times$512, rendered under four effects (pinhole, fisheye, rolling shutter, and depth of field).

Sec.~\ref{sec:blender_construction} details the Blender~\cite{blender_software} pipeline, tools, and plugins used for multi-view dynamic rendering, and Sec.~\ref{sec:scene_descriptions} describes the scenes and their material/physics characteristics.

\subsection{Scene Construction using Blender}
\label{sec:blender_construction}
\vspace{-0.5em}
To construct CamVFX-4D for evaluating model performance under different camera effects, we selected \textbf{Blender 4.5}~\cite{blender_software} as our primary scene construction and rendering tool. Blender provides highly realistic physics simulation capabilities, enabling accurate modeling of object motion, collisions, bouncing, and other dynamic behaviors. Additionally, it ensures completely consistent scene configurations across multiple renders, guaranteeing perfect content alignment under different camera effects while avoiding the limitations of real-world filming where identical dynamics cannot be reproduced.

For scene sourcing, we selected 4 high-quality indoor scene models through the \textbf{BlenderKit}~\cite{blenderkit} asset library. In addition, certain dynamic objects, such as the basketball model used in the basketball court scene, were also obtained from BlenderKit. In the living room and bathroom scenes, the dynamic objects were obtained from other online asset platforms: the LEGO bulldozer model~\cite{blendswap_lego_856_bulldozer} and the box animation model~\cite{blendswap_box_animation_template} from \textbf{Blend Swap}~\cite{blendswap}, the running cat model~\cite{free3d_cat_running} and a potted plant model~\cite{free3d_pot_plant} from \textbf{Free3D}~\cite{free3d}. Within each indoor environment, we designed 2 different types of dynamic objects and animations, such as bouncing spheres, bouncing cubes, or other wind-influenced motion behaviors. In total, we constructed \textbf{8 distinct dynamic scenes}.

We chose 8 scenes primarily because they encompass diverse material properties and physical interaction patterns, including reflective floor surfaces and rigid body collision dynamics. This diversity adequately simulates various common lighting and motion scenarios found in the real world, providing comprehensive coverage for subsequent model generalization testing. Simultaneously, maintaining a reasonable number of scenes ensures that each scene's dynamic design and camera configuration can undergo thorough manual adjustment and verification, preventing quality degradation due to an excessively large number of scenes.

For camera configuration, we established \textbf{50 different camera viewpoints} for each scene, with each viewpoint rendered separately under four camera effects: pinhole, fisheye, rolling shutter, and depth of field. We selected 50 viewpoints to maintain manageable data volumes while covering sufficient observation angles, ensuring comprehensive capture of dynamic object shape variations and lighting responses. This design also provides adequate observational diversity for our 4D Gaussian ray tracing model, facilitating learning of complete 4D scene structure during training.

Our 50 camera viewpoint placement, corresponding camera pose data, and initial scene point cloud generation were all accomplished through the \mbox{\textbf{PlenoBlenderNeRF}}~\cite{plenoblendernerf} Blender plugin, which was originally forked from \textbf{BlenderNeRF}~\cite{Raafat_BlenderNeRF_2024}. This plugin automatically distributes multiple cameras uniformly throughout the scene and generates corresponding initial point clouds from Blender mesh vertex data before rendering, ensuring complete consistency between rendering results under different camera effects and geometric foundations, facilitating subsequent quality analysis and comparison.

For data output, we generated \textbf{50-frame videos} at \textbf{512×512} resolution for each camera under all four camera effects in each scene, ensuring complete capture of dynamic object motion trajectories, lighting changes, and camera effect characteristics for subsequent analysis.

\subsection{Scene Descriptions and Characteristics}
\label{sec:scene_descriptions}
\vspace{-0.5em}
CamVFX-4D contains 8 dynamic scenes derived from 4 environment types: \textbf{basketball court}, \textbf{warehouse}, \textbf{living room}, and \textbf{bathroom}. Each scene incorporates distinct physical interaction patterns and surface properties in modeling and material settings, ensuring coverage of diverse lighting responses and motion behaviors. The following lists describe each scene's dynamic design and attributes.

\noindent\textbf{Basketball Court:}
\begin{enumerate}[leftmargin=1.5em, labelsep=0.5em, topsep=2pt, itemsep=2pt]
    \item \textbf{Rolling Basketballs:} We simulate external forces acting on three basketballs using Blender's \textbf{wind force}, generating rolling motion on the ground. During rolling, the balls experience both friction and inertia effects, producing gradually decelerating motion that closely approximates real physical behavior.
    
    \item \textbf{Falling Basketballs:} The scene contains three basketballs positioned above the ground, falling under gravity to the ground surface. After elastic collisions with the floor, the basketballs interact through further collisions, causing directional changes and eventual rolling in different directions. This design simultaneously simulates multiple physical phenomena including multi-body collisions, bouncing, and rolling, effectively testing model capabilities in handling multi-target dynamics and interactions during rendering and geometric reconstruction.
\end{enumerate}

\noindent\textbf{Warehouse:}
\begin{enumerate}[leftmargin=1.5em, labelsep=0.5em, topsep=2pt, itemsep=2pt]
    \item \textbf{Bouncing Spheres:} Three rigid spheres positioned at a height fall under gravity to the ground surface, undergoing elastic collisions followed by mutual interactions. The collision process alters sphere motion directions and velocities, causing final rolling in different directions and creating complex multi-body interaction and scattering dynamics.
    
    \item \textbf{Bouncing Cubes:} Four cubes fall from height to the ground surface, experiencing elastic collisions and mutual interactions. During collisions, cube motion trajectories and rotation angles continuously change, simulating non-spherical rigid body motion characteristics under multiple collision scenarios.
\end{enumerate}

\noindent\textbf{Living Room:}
\begin{enumerate}[leftmargin=1.5em, labelsep=0.5em, topsep=2pt, itemsep=2pt]
    \item \textbf{LEGO Bulldozer Model:} We utilize a LEGO bulldozer model as the primary dynamic object in the living room scene. Through animation design, we simulate mechanical arm swinging motions to present realistic mechanical operation dynamics and joint motion trajectories while preserving material details and lighting variations in rendering.
    
    \item \textbf{Running Cat Model:} We also employ a cat model with added temporal displacement and running animations, creating continuous motion postures within the scene. The animation encompasses both overall positional changes and continuous limb movements with posture variations, simulating realistic animal running physics and rhythmic characteristics.
\end{enumerate}

\noindent\textbf{Bathroom:}
\begin{enumerate}[leftmargin=1.5em, labelsep=0.5em, topsep=2pt, itemsep=2pt]
    \item \textbf{Rotating Potted Plant Model:} We utilize a detailed potted plant model as an additional dynamic element in the bathroom scene. Using Blender's keyframe animation system, we create a smooth rotational motion where the potted plant rotates 180 degrees around the z-axis throughout the animation sequence. This rotation provides a controlled, predictable motion pattern that allows for systematic evaluation of how different camera effects handle circular motion and changing viewpoints of complex organic geometry.
    \item \textbf{Box Animation Model:} The second dynamic element in the bathroom scene is a self-opening box whose lid gradually reveals its interior. The motion involves coordinated hinge movement, producing geometric changes, occlusion effects, and shifting light as inner surfaces are exposed—ideal for testing camera effects on mechanical motion and surface transitions.
\end{enumerate}

\begin{figure*}[htbp]
\centering
\setlength{\tabcolsep}{2pt}
\renewcommand{\arraystretch}{1.1}

\newcommand{\imgw}{0.24\linewidth}

\begin{minipage}{\linewidth}
\footnotesize\textit{``A video of a person skateboarding in a city square, captured using a \textbf{fisheye lens} with a \textbf{120-degree field of view}.''}
\end{minipage}
\\[2pt]
\includegraphics[width=\imgw]{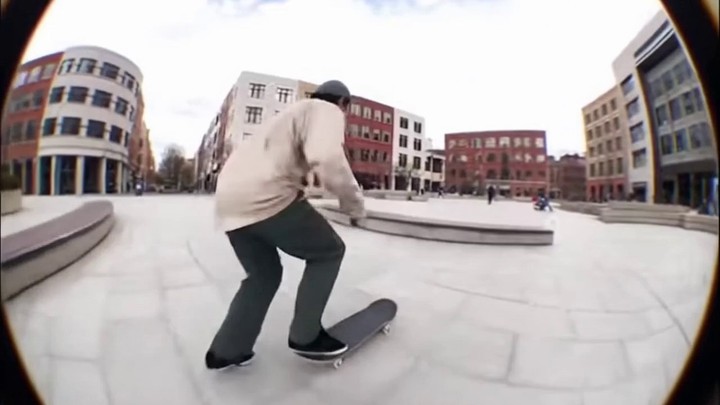}\hfill
\includegraphics[width=\imgw]{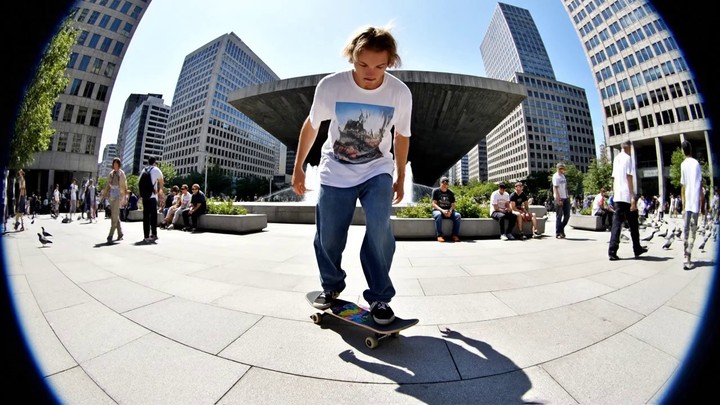}\hfill
\includegraphics[width=\imgw]{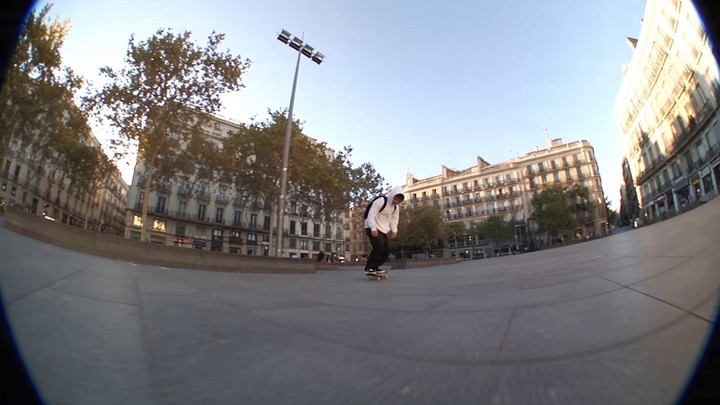}\hfill
\includegraphics[width=\imgw]{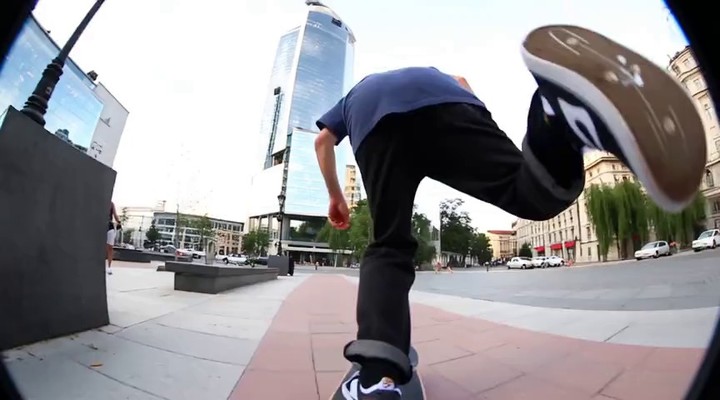}
\\[6pt]

\begin{minipage}{\linewidth}
\footnotesize\textit{``A video of a person skateboarding in a city square, captured using a \textbf{fisheye lens} with a \textbf{150-degree field of view}.''}
\end{minipage}
\\[2pt]
\includegraphics[width=\imgw]{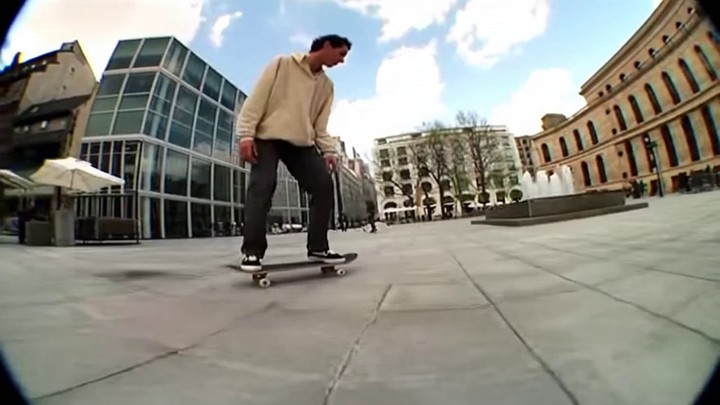}\hfill
\includegraphics[width=\imgw]{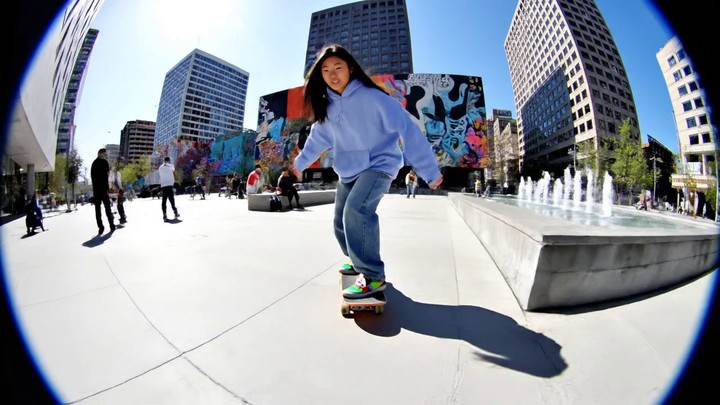}\hfill
\includegraphics[width=\imgw]{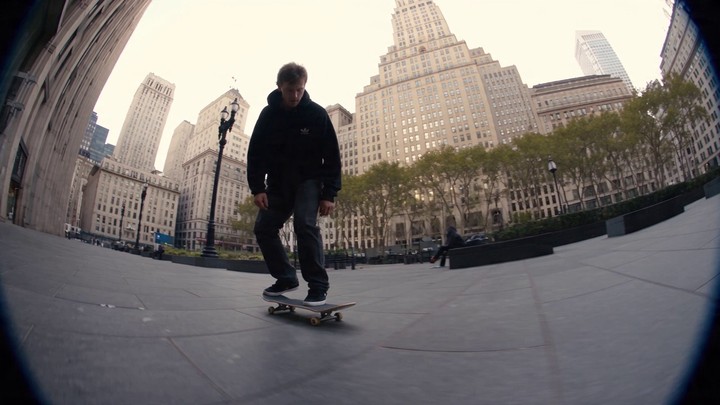}\hfill
\includegraphics[width=\imgw]{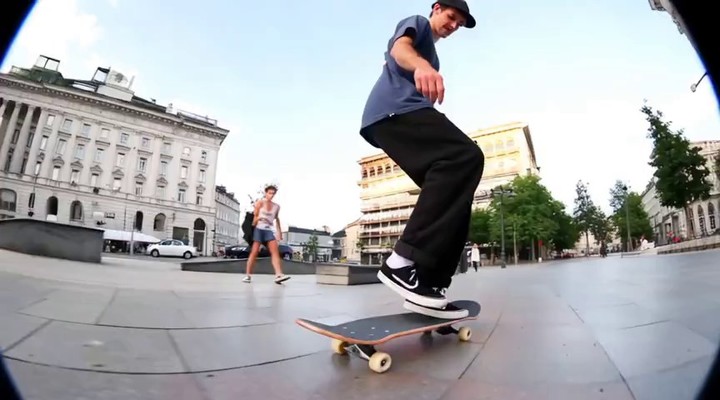}
\\[6pt]

\begin{minipage}{\linewidth}
\footnotesize\textit{``A video of a person skateboarding in a city square, captured using a \textbf{fisheye lens} with a \textbf{180-degree field of view}.''}
\end{minipage}
\\[2pt]
\includegraphics[width=\imgw]{figs/poc_result/sora_fisheye_180-degree.jpg}\hfill
\includegraphics[width=\imgw]{figs/poc_result/veo_fisheye_180-degree.jpg}\hfill
\includegraphics[width=\imgw]{figs/poc_result/seedance20_fisheye_180-degree.jpg}\hfill
\includegraphics[width=\imgw]{figs/poc_result/wan_fisheye_180.jpg}
\\[4pt]

\makebox[\imgw]{\small\textbf{Sora 2 Pro~\cite{OpenAISora2}}}\hfill
\makebox[\imgw]{\small\textbf{Veo 3.1~\cite{GoogleVeo3}}}\hfill
\makebox[\imgw]{\small\textbf{Seedance 2.5~\cite{seedance25}}}\hfill
\makebox[\imgw]{\small\textbf{Wan 2.2 T2V A14B~\cite{wan2025}}}

\caption{Fisheye video evaluation. Although the models produce recognizable fisheye imagery, changes in the requested field of view do not yield consistent, parameter-aligned projection changes.}
\label{fig:fisheye_result}
\vspace{-1em}
\end{figure*}

\section{Video Generation Experiment}
\label{supp:video_gen}
\vspace{-0.5em}
To evaluate the capabilities of state-of-the-art video generation models in synthesizing videos with realistic camera effects, we conduct comprehensive experiments using four leading video generation models: OpenAI Sora 2 Pro~\cite{OpenAISora2}, Google Veo 3.1~\cite{GoogleVeo3}, ByteDance Seedance 2.5~\cite{seedance25}, and Alibaba Wan 2.2 T2V A14B~\cite{wan2025}. Our evaluation focuses on three fundamental camera effects: fisheye distortion, rolling-shutter artifacts, and depth of field. These effects represent critical aspects of realistic video synthesis that require an accurate understanding of camera optics and sensor behavior.

\subsection{Video Generation Configuration}
\vspace{-0.5em}
To ensure fair comparison while respecting each model's individual design constraints and optimal operating conditions, we use the default output settings for each video generation model. This approach allows us to evaluate the models under their native configurations rather than forcing uniform parameters that may disadvantage certain architectures.

\begin{figure*}[htbp]
\centering
\setlength{\tabcolsep}{2pt}
\renewcommand{\arraystretch}{1.1}

\newcommand{\imgw}{0.24\linewidth}

\begin{minipage}{\linewidth}
\footnotesize\textit{``A video of a foreground chess piece on a chessboard, with several chess pieces receding into the background, shot with a \textbf{35 mm lens at aperture f/2.0}, with the focus distance set to the foreground chess piece.''}
\end{minipage}
\\[2pt]
\includegraphics[width=\imgw]{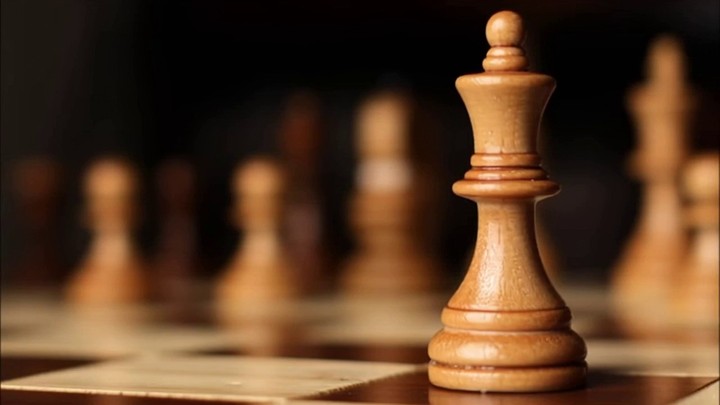}\hfill
\includegraphics[width=\imgw]{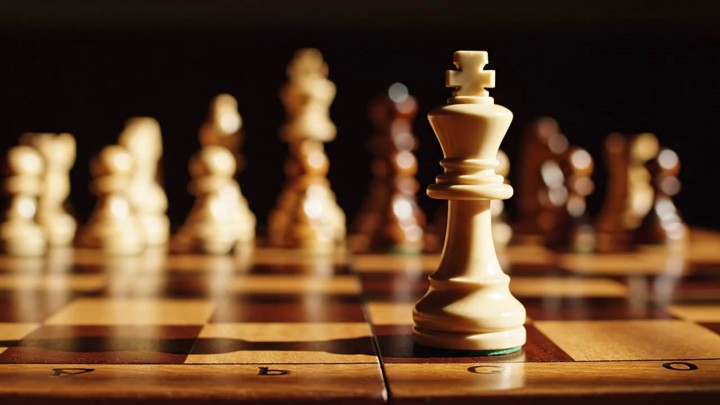}\hfill
\includegraphics[width=\imgw]{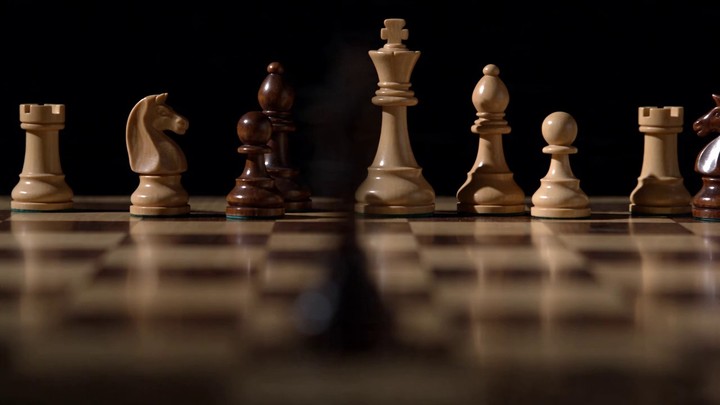}\hfill
\includegraphics[width=\imgw]{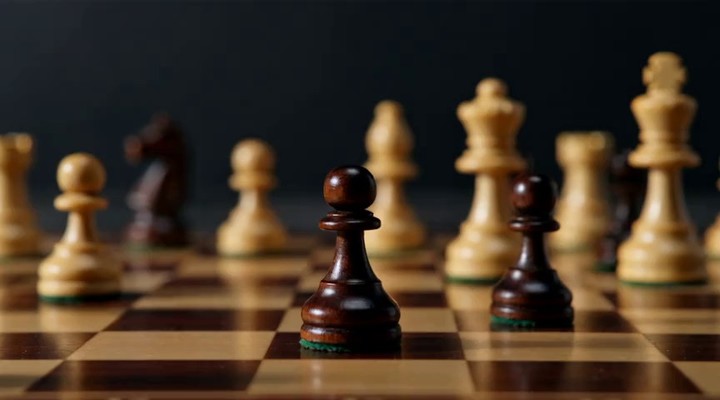}
\\[6pt]

\begin{minipage}{\linewidth}
\footnotesize\textit{``A video of a foreground chess piece on a chessboard, with several chess pieces receding into the background, shot with a \textbf{35 mm lens at aperture f/5.6}, with the focus distance set to the foreground chess piece.'' For Seedance 2.5, the corresponding prompt used \textbf{f/5.0}.}
\end{minipage}
\\[2pt]
\includegraphics[width=\imgw]{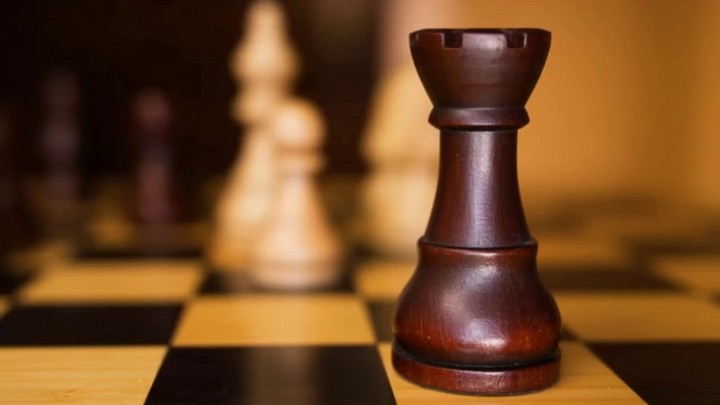}\hfill
\includegraphics[width=\imgw]{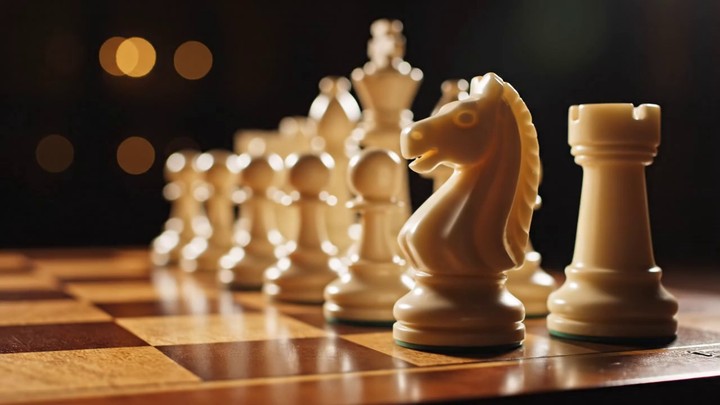}\hfill
\includegraphics[width=\imgw]{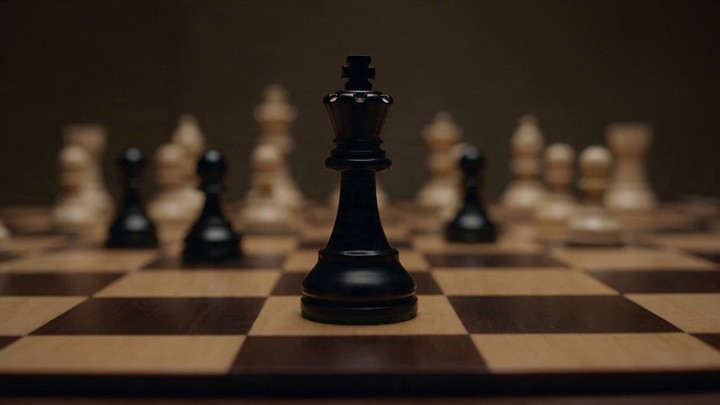}\hfill
\includegraphics[width=\imgw]{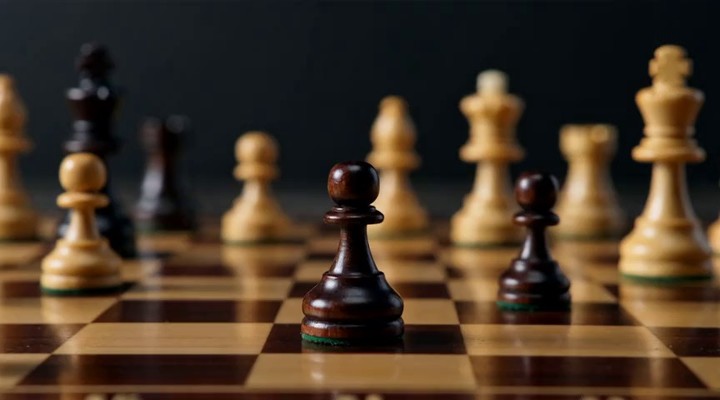}
\\[6pt]

\begin{minipage}{\linewidth}
\footnotesize\textit{``A video of a foreground chess piece on a chessboard, with several chess pieces receding into the background, shot with a \textbf{35 mm lens at aperture f/16.0}, with the focus distance set to the foreground chess piece.''}
\end{minipage}
\\[2pt]
\includegraphics[width=\imgw]{figs/poc_result/sora_dof_f-16-0.jpg}\hfill
\includegraphics[width=\imgw]{figs/poc_result/veo_dof_f-16-0.jpg}\hfill
\includegraphics[width=\imgw]{figs/poc_result/seedance20_dof_f-16-0.jpg}\hfill
\includegraphics[width=\imgw]{figs/poc_result/wan_dof_f-16-0.jpg}
\\[4pt]

\makebox[\imgw]{\small\textbf{Sora 2 Pro~\cite{OpenAISora2}}}\hfill
\makebox[\imgw]{\small\textbf{Veo 3.1~\cite{GoogleVeo3}}}\hfill
\makebox[\imgw]{\small\textbf{Seedance 2.5~\cite{seedance25}}}\hfill
\makebox[\imgw]{\small\textbf{Wan 2.2 T2V A14B~\cite{wan2025}}}

\caption{Depth-of-field video evaluation. All models produce pronounced blur regardless of the specified aperture ($f$/2.0 vs. $f$/16.0), revealing weak compliance with depth-of-field parameters and poor control of depth-of-field effects.}
\label{fig:dof_result}
\vspace{-1em}
\end{figure*}

\subsection{Fisheye Distortion Effect Generation}
\vspace{-0.5em}
We evaluate the controllability of fisheye video generation using the same prompt template while varying only the specified field of view (FOV). As shown in Fig.~\ref{fig:fisheye_result}, the tested models can produce recognizable fisheye imagery, but none reliably varies the projection according to the requested FOV. These results indicate that current video generation models lack reliable parameter-level control over fisheye camera effects.

\subsection{Depth-of-Field Effect Generation}
\vspace{-0.5em}
We evaluate depth-of-field controllability using the same prompt template with a fixed focal length and focus distance, while varying only the aperture. Ideally, a larger aperture ($f$/2.0) should produce strong background defocus, whereas a smaller aperture ($f$/16.0) should keep most of the scene in focus. However, as shown in Fig.~\ref{fig:dof_result}, all tested models generate noticeable background blur regardless of the specified aperture, demonstrating limited control over depth-of-field parameters.

\begin{figure*}[htbp]
\centering
\setlength{\tabcolsep}{2pt}
\renewcommand{\arraystretch}{1.1}

\newcommand{\imgw}{0.24\linewidth}

\begin{minipage}{\linewidth}
\footnotesize\textit{``A close-up video of a centered electric fan with clearly visible blades spinning at high speed, captured with a \textbf{rolling-shutter camera using a readout time of 5 ms}. The fan fills most of the frame, the background is simple, and the exposure is short with minimal motion blur.''}
\end{minipage}
\\[2pt]
\includegraphics[width=\imgw]{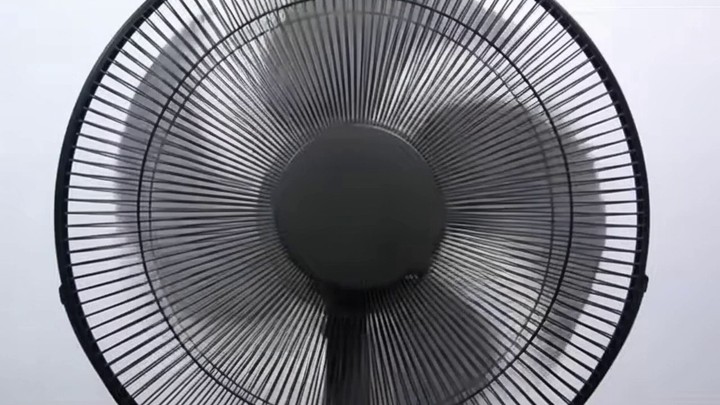}\hfill
\includegraphics[width=\imgw]{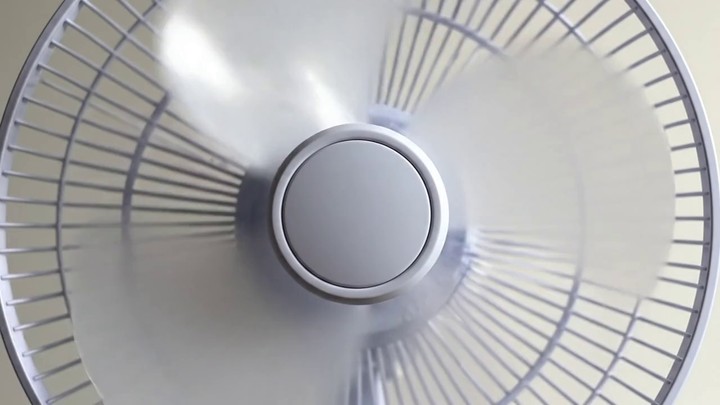}\hfill
\includegraphics[width=\imgw]{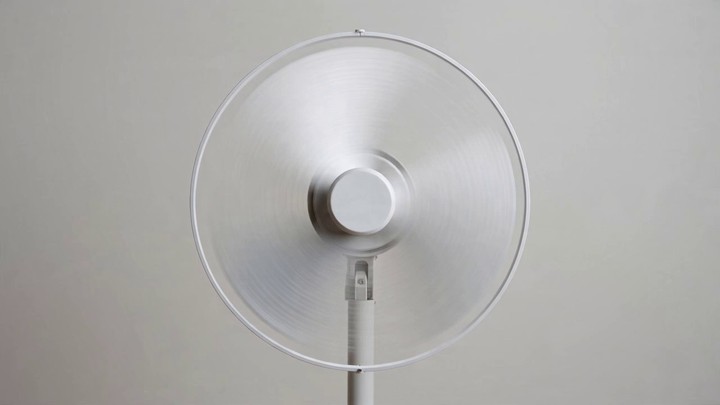}\hfill
\includegraphics[width=\imgw]{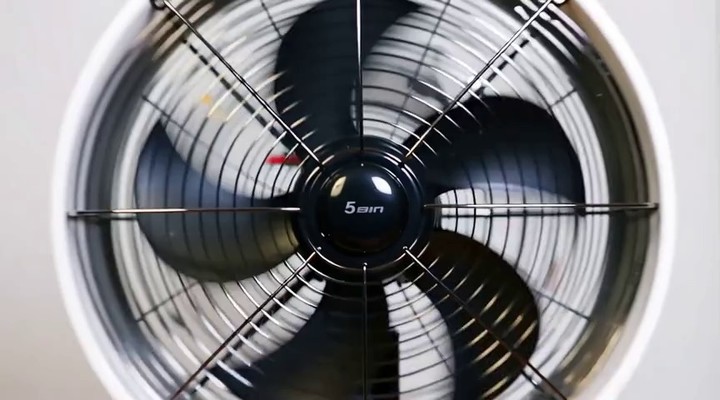}
\\[6pt]

\begin{minipage}{\linewidth}
\footnotesize\textit{``A close-up video of a centered electric fan with clearly visible blades spinning at high speed, captured with a \textbf{rolling-shutter camera using a readout time of 20 ms}. The fan fills most of the frame, the background is simple, and the exposure is short with minimal motion blur.''}
\end{minipage}
\\[2pt]
\includegraphics[width=\imgw]{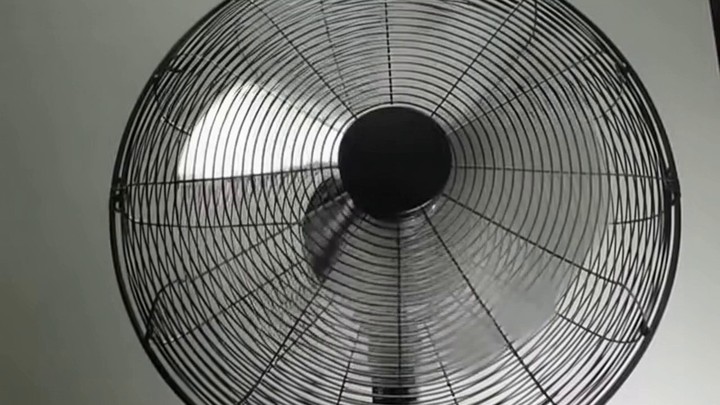}\hfill
\includegraphics[width=\imgw]{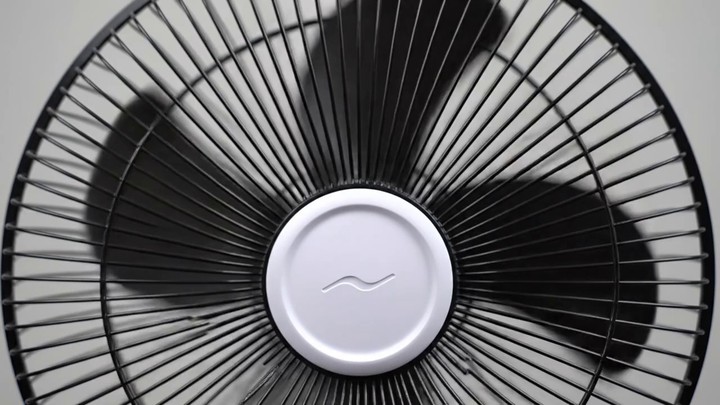}\hfill
\includegraphics[width=\imgw]{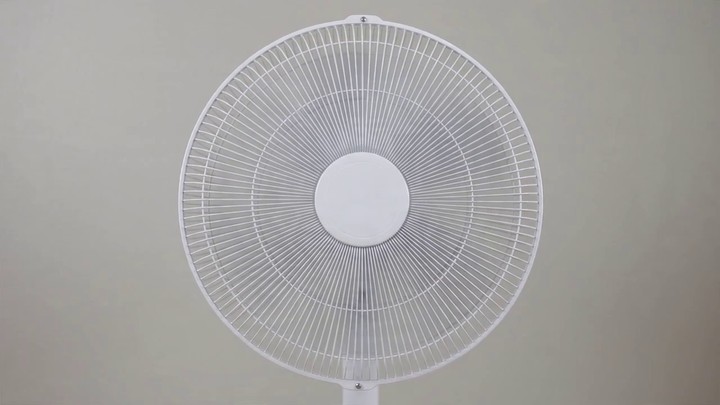}\hfill
\includegraphics[width=\imgw]{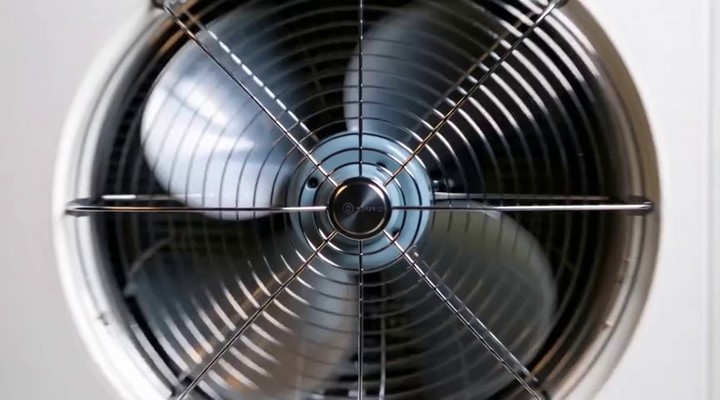}
\\[6pt]

\begin{minipage}{\linewidth}
\footnotesize\textit{``A close-up video of a centered electric fan with clearly visible blades spinning at high speed, captured with a\textbf{ rolling-shutter camera using a readout time of 40 ms}. The fan fills most of the frame, the background is simple, and the exposure is short with minimal motion blur.''}
\end{minipage}
\\[2pt]
\includegraphics[width=\imgw]{figs/poc_result/sora_rolling-shutter_40-ms.jpg}\hfill
\includegraphics[width=\imgw]{figs/poc_result/veo_rolling-shutter_40-ms.jpg}\hfill
\includegraphics[width=\imgw]{figs/poc_result/seedance25_rolling-shutter_40-ms.jpg}\hfill
\includegraphics[width=\imgw]{figs/poc_result/wan_rolling_shutter_40ms.jpg}
\\[4pt]

\makebox[\imgw]{\small\textbf{Sora 2 Pro~\cite{OpenAISora2}}}\hfill
\makebox[\imgw]{\small\textbf{Veo 3.1~\cite{GoogleVeo3}}}\hfill
\makebox[\imgw]{\small\textbf{Seedance 2.5~\cite{seedance25}}}\hfill
\makebox[\imgw]{\small\textbf{Wan 2.2 T2V A14B~\cite{wan2025}}}

\caption{Rolling-shutter video evaluation. Under a rolling-shutter camera model, the fan blades should exhibit row-wise geometric skew. However, the tested models produce either non-skewed blades or generic motion-blurred blades, even when the prompt explicitly requests minimal motion blur.}
\label{fig:rolling_shutter_result}
\vspace{-1em}
\end{figure*}

\subsection{Rolling-Shutter Effect Generation}
\vspace{-0.5em}
Similarly, we evaluate rolling-shutter controllability using the same prompt template while varying only the readout time. Ideally, a longer readout time should produce stronger row-wise skewing of the fan blades. However, in Fig.~\ref{fig:rolling_shutter_result}, the tested models generate either normal, non-skewed blades or motion-blurred blades, suggesting that the specified readout time is not faithfully reflected in the generated motion geometry.

\section{Implementation Details}
\label{supp:implem_detail}
\vspace{-0.5em}
\subsection{Gaussian Deformation Network Architecture}
\vspace{-0.5em}
We adopt the 4D-GS~\cite{wu20244d} formulation of the Gaussian deformation network to predict the deformation of 3D Gaussians at a specific time step. Specifically, the network consists of a spatio-temporal structure encoder and a multi-head Gaussian deformation decoder. 

The spatio-temporal encoder is constructed using six multi-resolution plane modules
\[
R_l(i, j) \in \mathbb{R}^{h \times lN_i \times lN_j},
\]
and an MLP $\varphi$, where $l$ is the upsampling scale, $h$ is the dimension of the voxel feature, and $N_i$ and $N_j$ are the base resolutions of the plane. Given the mean value $(x, y, z)$ of a 3D Gaussian
\[
G = \{(x, y, z), r, s, \sigma, \mathcal{C}\},
\]
we first compute the voxel feature
\begin{align}
f_h &= \bigcup_{l} \prod \mathbf{interp}(R_l(i, j)) \in \mathbb{R}^{h\times l}, \\
(i, j) &\in \{(x, y), (x, z), (y, z), (x, t), (y, t), (z, t)\},
\end{align}
where $\mathbf{interp}$ denotes bilinear interpolation. The feature $f_h$ is then merged by $\varphi$ to produce
\[
f = \varphi(f_h).
\]

Next, the multi-head Gaussian deformation decoder predicts the deformation of 3D Gaussians. The decoder consists of separate MLPs $\{\varphi_x, \varphi_r, \varphi_s\}$, which predict the deformation residuals of each Gaussian attribute:
\[
(\Delta x, \Delta y, \Delta z) = \varphi_x(f), \quad 
\Delta r = \varphi_r(f), \quad 
\Delta s = \varphi_s(f).
\] 

These residuals are then applied to deform a 3D Gaussian $G$ into
\[
G_t = \{(x + \Delta x, y + \Delta y, z + \Delta z), \, r + \Delta r, \, s + \Delta s, \, \sigma, \, \mathcal{C}\}.
\]

\subsection{Training Details}
\label{supp:train_detail}
\vspace{-0.5em}
The hyperparameters used in our experiments are based primarily on those reported in 4D-GS~\cite{wu20244d}. We observe that the original 4D-GS configuration performs poorly on scenes with rapidly moving objects. We therefore increase the learning rate, deformation-field capacity, base resolution of the plane modules, and number of training iterations. Specifically, for scenes with rapid motion, we set the deformation MLP width and the base resolution of the plane modules to $128$, with upsampling factors of $2$, $4$, and $8$. We increase the densification threshold from $2.5 \times 10^{-5}$ to $1 \times 10^{-4}$ and train for $5 \times 10^4$ iterations in the coarse stage and $8 \times 10^4$ iterations in the fine stage.

We adopt a two-stage optimization strategy similar to 4D-GS~\cite{wu20244d}. In the first stage, we initialize the positions and colors of the 3D Gaussians using point clouds generated from Blender’s synthetic scenes. We then train the 3D Gaussian scene without the deformation network, using only the first frame. This step reconstructs the static components of the scene and provides a strong initialization for dynamic objects. In the second stage, we introduce the deformation network and jointly optimize both the 3D Gaussians and the network across all time frames.

In 3D-GS~\cite{kerbl20233d}, densification is guided by the magnitude of the 2D positional gradients of Gaussians. In contrast, since our rendering relies on ray tracing rather than rasterization, we follow the approach of~\cite{3dgrt2024} and use the magnitude of the 3D gradient of the Gaussian means to decide when to densify a Gaussian. Additionally, as in~\cite{3dgrt2024}, we scale the gradient magnitude by the distance of the deformed Gaussian from the camera, encouraging densification in the background.

\subsection{Fine-tuning Details}
\label{supp:fine_tune_detail}
\vspace{-0.5em}
Following AKiRa~\cite{wang2025akira}, we introduce camera-effect control into Stable Video Diffusion (SVD)~\cite{blattmann2023stable}. The conditioning signal is formed by concatenating Pl\"ucker rays and a defocus map, resulting in a 7-channel input. The Pl\"ucker rays are constructed using the specified camera parameters, and the defocus map is computed as $D_{\text{px}}(u,v)=\frac{2\alpha f\left|z(u,v)-z_f\right|}{z(u,v)\left(z_f-f\right)}$, where $\alpha$ is the aperture size, $f$ is the focal length, $z(u,v)$ is the depth at pixel $(u,v)$, and $z_f$ is the focus depth. 

To prepare the training data, we use the \mbox{MultiCamVideo}~\cite{bai2025recammaster} dataset, which contains dynamic scenes constructed by compositing synthetic backgrounds with dynamic foreground objects. We render each reconstructed scene under multiple camera-effect configurations, including 3 zooming videos, 3 fisheye videos, 5 depth-of-field videos, and 3 dolly-zoom videos per scene. For each effect, we randomize the camera parameters to encourage model generalization.

\section{Additional Ablations}
\label{supp:add_ablations}
\vspace{-0.5em}
\subsection{Dynamic Gaussian Representations}
\vspace{-0.5em}
To examine whether our framework depends on a particular dynamic Gaussian representation, we evaluate two recent methods, Swift4D~\cite{wu2025swiftd} and Hybrid 3D-4DGS~\cite{oh2025hybrid}, in place of 4D-GS~\cite{wu20244d}, while keeping 3DGRT~\cite{3dgrt2024} as the rendering backend. Table~\ref{tab:supp_dynamic_gs} reports the quantitative results. All three representations are compatible with the same camera-effect rendering pipeline, although reconstruction quality varies with the underlying representation. These results demonstrate the flexibility of our paradigm: it can switch among dynamic Gaussian representations while maintaining competitive rendering quality across camera effects.

\FloatBarrier
\begin{table}[t]
    \centering
    \caption{Comparison of dynamic Gaussian representations using 3DGRT as the rendering backend. Fisheye metrics are computed within the circular image mask, and rolling shutter uses one row per chunk.}
    \label{tab:supp_dynamic_gs}
    \footnotesize
    \setlength{\tabcolsep}{2pt}
    \begin{tabular*}{\columnwidth}{@{\extracolsep{\fill}}lrrrr@{}}
        \toprule
        Representation & PSNR$\uparrow$ & SSIM$\uparrow$ & LPIPS$\downarrow$ & FPS$\uparrow$ \\
        \midrule
        \multicolumn{5}{@{}l}{\emph{Pinhole}} \\
        4D-GS~\cite{wu20244d}        & 32.8010          & 0.8898          & \textbf{0.1018} & 36.56 \\
        Swift4D~\cite{wu2025swiftd}  & \textbf{33.1394} & \textbf{0.8925} & 0.1697          & 48.75 \\
        Hybrid 3D-4DGS~\cite{oh2025hybrid} & 25.7181 & 0.8092 & 0.3576 & \textbf{61.60} \\
        \midrule
        \multicolumn{5}{@{}l}{\emph{Fisheye}} \\
        4D-GS~\cite{wu20244d}        & \textbf{28.8927} & 0.8555          & \textbf{0.1259} & 41.53 \\
        Swift4D~\cite{wu2025swiftd}  & 26.3598          & \textbf{0.8748} & 0.1499          & 58.55 \\
        Hybrid 3D-4DGS~\cite{oh2025hybrid} & 23.4842 & 0.8205 & 0.2587 & \textbf{62.18} \\
        \midrule
        \multicolumn{5}{@{}l}{\emph{Depth of field}} \\
        4D-GS~\cite{wu20244d}        & \textbf{31.2475} & \textbf{0.9124} & \textbf{0.1210} & 3.44 \\
        Swift4D~\cite{wu2025swiftd}  & 30.7638          & 0.9109          & 0.2015          & 8.99 \\
        Hybrid 3D-4DGS~\cite{oh2025hybrid} & 26.8416 & 0.8812 & 0.2956 & \textbf{14.27} \\
        \midrule
        \multicolumn{5}{@{}l}{\emph{Rolling shutter}} \\
        4D-GS~\cite{wu20244d}        & \textbf{31.6144} & \textbf{0.8821} & \textbf{0.1056} & \textbf{0.76} \\
        Swift4D~\cite{wu2025swiftd}  & 28.8920          & 0.8761          & 0.1849          & 0.24 \\
        Hybrid 3D-4DGS~\cite{oh2025hybrid} & 24.9709 & 0.8049 & 0.3614 & 0.14 \\
        \bottomrule
    \end{tabular*}
\end{table}

\subsection{Rendering Backends}
\vspace{-0.5em}
We further fix the dynamic representation to 4D-GS and compare three rendering backends in Table~\ref{tab:supp_rendering_backends}. Our current 3DGUT~\cite{wu20253dgut} integration reports only pinhole and fisheye results because this backend does not directly support the effect-aware or row-dependent projection and tile assignment required for depth of field and rolling-shutter rendering. Despite differences in their rendering mechanisms, the backends maintain competitive rendering quality across supported camera effects, showing that our paradigm is not tied to a specific backend.

\section{Additional Quantitative Results}
\label{supp:add_quant}
\vspace{-0.5em}
Table~\ref{tab:training_time_storage} provides a comparison of training time and model storage. HexPlane requires the least storage, while MSTH has the shortest training time. 4D-GRT strikes a balance between the two, requiring substantially less storage than MSTH while training considerably faster than HexPlane.

\section{Additional Qualitative Results}
\label{supp:add_qual}
\begin{figure*}[t]
    \centering
    \includegraphics[width=\textwidth]{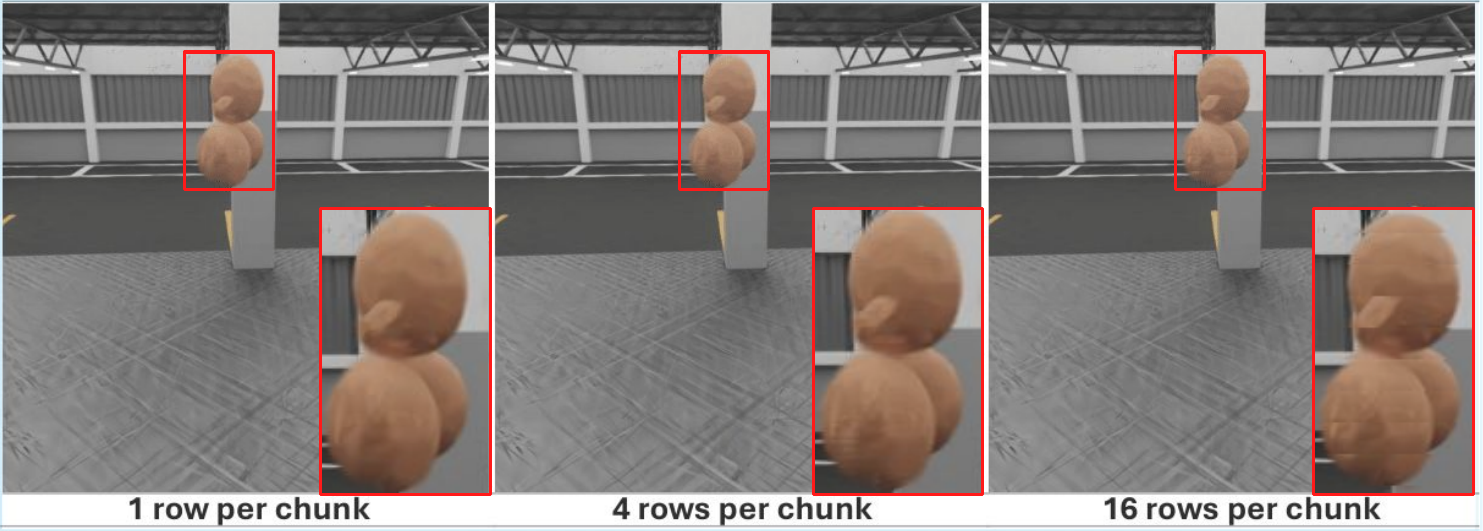}
    \caption{ 
    Qualitative comparison of the rolling-shutter effect with different numbers of rows per chunk.
    }
    \label{fig:rs_w_diff_chunk}
\end{figure*}

We present a qualitative comparison of rolling-shutter rendering with different chunk sizes. As shown in Fig.~\ref{fig:rs_w_diff_chunk}, smaller chunk sizes produce more accurate renderings, while larger chunk sizes lead to noticeable blocking artifacts.

\begin{table}[t]
    \centering
    \caption{Comparison of rendering backends using 4D-GS as the dynamic representation. Fisheye metrics are computed within the circular image mask, and rolling shutter uses one row per chunk. A dash denotes an effect not directly supported by the current backend integration.}
    \label{tab:supp_rendering_backends}
    \footnotesize
    \setlength{\tabcolsep}{2pt}
    \begin{tabular*}{\columnwidth}{@{\extracolsep{\fill}}lrrrr@{}}
        \toprule
        Backend & PSNR$\uparrow$ & SSIM$\uparrow$ & LPIPS$\downarrow$ & FPS$\uparrow$ \\
        \midrule
        \multicolumn{5}{@{}l}{\emph{Pinhole}} \\
        3DGRT~\cite{3dgrt2024} & \textbf{32.8010} & \textbf{0.8898} & \textbf{0.1018} & 36.56 \\
        3DGUT~\cite{wu20253dgut} & 32.2641 & 0.8756 & 0.1433 & \textbf{74.37} \\
        3DGEER~\cite{huang20263dgeer3dgaussianrendering} & 31.9906 & 0.8660 & 0.1400 & 53.36 \\
        \midrule
        \multicolumn{5}{@{}l}{\emph{Fisheye}} \\
        3DGRT~\cite{3dgrt2024} & \textbf{28.8927} & \textbf{0.8555} & \textbf{0.1259} & 41.53 \\
        3DGUT~\cite{wu20253dgut} & 24.6197 & 0.8001 & 0.2309 & \textbf{64.51} \\
        3DGEER~\cite{huang20263dgeer3dgaussianrendering} & 26.6393 & 0.8175 & 0.1904 & 48.84 \\
        \midrule
        \multicolumn{5}{@{}l}{\emph{Depth of field}} \\
        3DGRT~\cite{3dgrt2024} & \textbf{31.2475} & \textbf{0.9124} & \textbf{0.1210} & 3.44 \\
        3DGUT~\cite{wu20253dgut} & -- & -- & -- & -- \\
        3DGEER~\cite{huang20263dgeer3dgaussianrendering} & 30.6853 & 0.9070 & 0.1237 & \textbf{15.34} \\
        \midrule
        \multicolumn{5}{@{}l}{\emph{Rolling shutter}} \\
        3DGRT~\cite{3dgrt2024} & \textbf{31.6144} & \textbf{0.8821} & \textbf{0.1056} & \textbf{0.76} \\
        3DGUT~\cite{wu20253dgut} & -- & -- & -- & -- \\
        3DGEER~\cite{huang20263dgeer3dgaussianrendering} & 30.3761 & 0.8594 & 0.1436 & 0.10 \\
        \bottomrule
    \end{tabular*}
\end{table}

\begin{table}[t]
  \centering
  \caption{Comparison of training time and storage.}
  \label{tab:training_time_storage}
  \small
  \setlength{\tabcolsep}{4pt}
  \begin{tabular}{lcc}
    \toprule
    Methods & Time$\downarrow$ & Storage (MB)$\downarrow$ \\
    \midrule
    HexPlane~\cite{cao2023hexplane}     & 12 hours & \textbf{238}  \\
    MSTH~\cite{wang2023masked}         & \textbf{8 mins} & 889  \\
    4D-GRT (Ours) & 3 hours & 364.875  \\
    \bottomrule
  \end{tabular}
  \vspace{-1.0em}
\end{table}

\begin{table}[htbp]
    \centering
    \caption{Quantitative results of applying our camera-effect renderer to monocular 4D-GS reconstructions produced by MoVieS~\cite{lin2026movies}.}
    \label{tab:supp_movies}
    \small
    \setlength{\tabcolsep}{7pt}
    \begin{tabular}{lrrr}
        \toprule
        Effect & PSNR $\uparrow$ & SSIM $\uparrow$ & LPIPS $\downarrow$ \\
        \midrule
        Pinhole         & 24.7302 & 0.8842 & 0.0723 \\
        Fisheye         & 22.1009 & 0.8057 & 0.1609 \\
        Depth of field  & 23.4781 & 0.7760 & 0.3093 \\
        Rolling shutter & 24.5100 & 0.8723 & 0.0765 \\
        \bottomrule
    \end{tabular}
\end{table}

\section{Camera-Effect Synthesis from Monocular Videos}
\label{supp:movies}
\vspace{-0.5em}
We evaluate the monocular extension of our pipeline using the pretrained MoVieS model~\cite{lin2026movies} on all eight CamVFX-4D scenes. We treat each of the 50 camera views in a scene as an independent monocular input, producing 400 reconstructions in total. Each reconstruction is passed to our ray-tracing renderer to synthesize pinhole, fisheye, depth-of-field, and rolling-shutter videos. Table~\ref{tab:supp_movies} reports the pinhole reconstruction quality and camera-effect synthesis quality. The results indicate that the monocular pipeline achieves reasonable visual quality across all four settings.

\section{Limitations}
\label{supp:limitations}
\vspace{-0.5em}
As with other reconstruction-based rendering methods, the output quality of 4D-GRT depends on the coverage of the input observations and the fidelity of the underlying dynamic-scene reconstruction. Errors caused by sparse observations, occlusions, or challenging motion may propagate to the rendered camera effects. Future work could combine the explicit physical renderer with diffusion priors to improve weakly observed regions and reduce residual reconstruction artifacts while preserving explicit control over camera parameters.

Our camera models are also limited to those used by Blender, which renders our ground truth. In particular, the Blender lens-polynomial fisheye model may not represent every real lens. Other models, such as Kannala--Brandt fisheye or polygonal apertures, can be supported by modifying ray generation, but we have not evaluated them. For rolling shutter, the chunking approximation shares one sensing time across all rows in a chunk, which can introduce artifacts under fast object or camera motion. Smaller chunks reduce this error at a higher rendering cost.

\end{document}